\documentclass[runningheads]{llncs}

\usepackage{eccv}

\usepackage{eccvabbrv}

\usepackage{graphicx}
\usepackage{booktabs}
\usepackage[table]{xcolor}
\usepackage{colortbl}
\usepackage{wrapfig}
\usepackage{caption}
\usepackage[accsupp]{axessibility}  
\usepackage{multirow}

\usepackage{xr-hyper}
\usepackage{hyperref}

\usepackage{orcidlink}

\definecolor{firstblue}{RGB}{140,180,255}
\definecolor{secondblue}{RGB}{200,220,255}
\definecolor{thirdblue}{RGB}{229,238,255}
\definecolor{bestcell}{RGB}{140,180,255}
\definecolor{secondcell}{RGB}{200,220,255}
\definecolor{thirdcell}{RGB}{229,238,255}

\newcommand{\JC}[1]{{\color{red}{[Jorge: #1]}}}
\newcommand{\QW}[1]{{\color{magenta}{[Qi: #1]}}}
\newcommand{\MND}[1]{{\color{blue}{[Merlin: #1]}}}
\newcommand{\ZG}[1]{{\color{cyan}{[Zan: #1]}}}
\newcommand{\PD}[1]{{\color{Purple}{[Piotr: #1]}}}
\newcommand{\NML}[1]{{\color{Green}{[Nico: #1]}}}

\renewcommand{\JC}[1]{}
\renewcommand{\QW}[1]{}
\renewcommand{\MND}[1]{}
\renewcommand{\ZG}[1]{}
\renewcommand{\PD}[1]{}
\renewcommand{\NML}[1]{}

\begin{document}

\title{Neural Harmonic Textures for High-Quality Primitive Based Neural Reconstruction}

\titlerunning{Neural Harmonic Textures}

\author{Jorge Condor\inst{1,2}\orcidlink{0000-0002-9958-0118} \and
Nicolas Mo\"enne-Loccoz\inst{1}\orcidlink{0000-0002-2312-9275}\and
Merlin Nimier-David\inst{1}\orcidlink{0000-0002-6234-3143} \and
Piotr Didyk\inst{2}\orcidlink{0000-0003-0768-8939} \and
Zan Gojcic\inst{1}\orcidlink{0000-0001-6392-2158} \and
Qi Wu\inst{1}\orcidlink{0000-0003-0342-9366}}

\authorrunning{J.~Condor et al.}

\institute{NVIDIA\\
\email{\{nicolasm,mnimierdavid,zgojcic,qiwu\}@nvidia.com}
 \and
 Universit\`a della Svizzera italiana, Lugano, Switzerland\\
\email{\{jorge.condor,piotr.didyk\}@usi.ch}}

\maketitle

\begin{figure*}
    \centering
    \includegraphics[width=\textwidth]{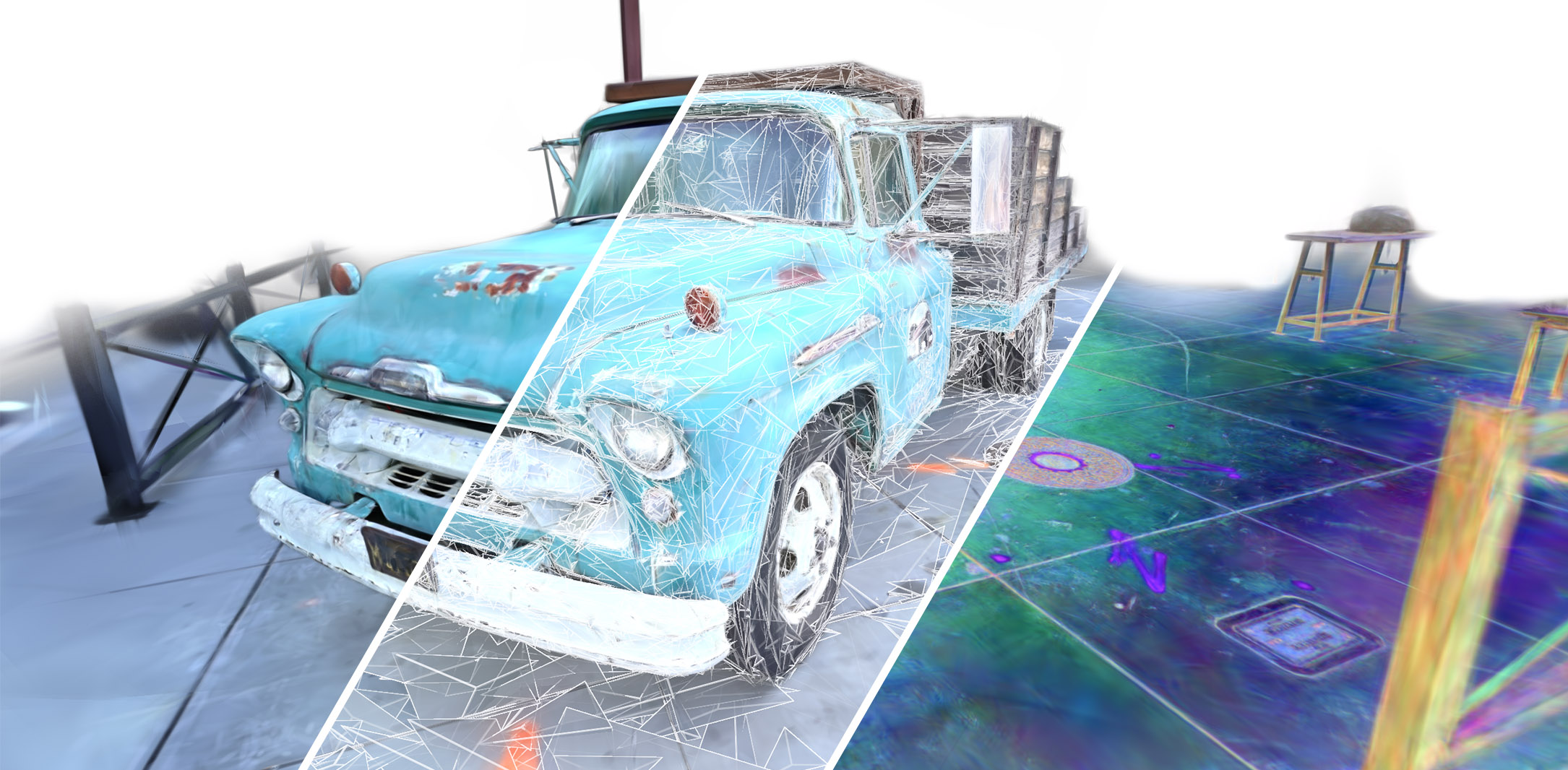}
    \caption{
    \textbf{Neural Harmonic Textures for novel view synthesis.}
We attach learnable feature vectors (right) to the virtual vertices of bounding tetrahedra encapsulating each primitive (center). After harmonic encoding and accumulation along the ray, a small neural network decodes the resulting signal into RGB color in a deferred manner (left).
Source code and further results are available at \url{https://research.nvidia.com/labs/sil/projects/neural-harmonic-textures/}.
}
    \label{fig:teaser}
\end{figure*}

\begin{abstract}
  \label{sec_abstract}

Primitive-based methods such as 3D Gaussian Splatting have recently become the state-of-the-art for novel-view synthesis and related reconstruction tasks. Compared to neural fields, these representations are more flexible, adaptive, and scale better to large scenes. However, the limited expressivity of individual primitives makes modeling high-frequency detail challenging.
We introduce \emph{Neural Harmonic Textures}, a neural representation approach that anchors latent feature vectors on a virtual scaffold surrounding each primitive. These features are interpolated within the primitive at ray intersection points. Inspired by Fourier analysis, we apply periodic activations to the interpolated features, turning alpha blending into a weighted sum of harmonic components. The resulting signal is then decoded in a single deferred pass using a small neural network, significantly reducing computational cost.
Neural Harmonic Textures yield state-of-the-art results in real-time novel view synthesis while bridging the gap between primitive- and neural-field-based reconstruction.
Our method integrates seamlessly into existing primitive-based pipelines such as 3DGUT, Triangle Splatting, and 2DGS. We further demonstrate its generality with applications to 2D image fitting and semantic reconstruction. 

\end{abstract}

\section{Introduction}
\label{sec_introduction}

Since the introduction of 3D Gaussian Splatting~\cite{kerbl20233Dgaussians}, (Lagrangian) primitive-based approaches have largely displaced (Eulerian) continuous neural radiance fields~\cite{sitzmann20,mueller2022instant,takikawa2023compact} for novel view synthesis~\cite{kerbl20233Dgaussians,trianglesplatting}.
This shift is driven not only by significantly improved rendering speed, but also by the structural advantages of explicit representations. Primitive-based methods naturally adapt to scene detail, scale more gracefully, and readily support motion, deformation, and editing~\cite{4dgaussians,chen2024gaussianeditor}. Furthermore, they align well with feed-forward reconstruction pipelines~\cite{gslrm2024, liang2024btimer} and closely resemble widely adopted point-map representations~\cite{VGGT, wang2025pi}.
Despite these advantages, the limited expressive power of individual primitives remains a fundamental bottleneck. Geometry and appearance are tightly coupled within each primitive, forcing high-frequency spatial detail to be represented by increasing the number of primitives, which directly increases memory consumption and hampers rendering speed. Directional appearance modeling is constrained in a similar manner. View-dependent effects are typically represented using low-order Spherical Harmonics, which are spectrally band-limited and poorly suited for high-frequency phenomena such as sharp specular highlights. Alternative angular parametric functions, such as spherical Gaussians, spherical Beta kernels~\cite{betasplatting}, and spherical Voronoi~\cite{sphericalVoronoi} increase directional bandwidth, but leave the fundamental coupling between spatial support and appearance modeling unchanged.
In contrast, neural radiance fields achieve high local representational capacity by combining positional encodings such as Fourier features~\cite{tancik2020fourier, mildenhall2021nerf} or multi-resolution hash grids~\cite{mueller2022instant} with a learned multi-layer perceptron (MLP) 
decoding. Recent works~\cite{neuralshells,radiancemesh} attempted to bridge both paradigms by employing primitives as acceleration structures for querying a global neural field. However, these hybrid approaches inherit key limitations of global neural representations, particularly with respect to motion, deformation, editability and scalability.

In this work, we introduce \emph{Neural Harmonic Textures}, a novel primitive-based neural representation that increases the expressive power of individual primitives while retaining the advantages of Lagrangian representations. To this end, we reinterpret primitives as both geometric carriers and local positional encodings. Specifically, we attach learnable feature vectors to a virtual scaffold encapsulating each primitive (e.g., Gaussians or triangles) and interpolate them locally at ray–primitive intersection points. Compared to globally anchored feature representations, our formulation moves with the primitives and therefore naturally supports operations such as motion, deformation, and scene editing. 
In prior work, such interpolated features are typically processed by a lightweight MLP to decode RGB values, followed by alpha compositing across intersected primitives to produce the final color, requiring numerous neural evaluations. Instead, we draw inspiration from the Fourier transform, where a complex signal is expressed as a sum of harmonic components (i.e., periodic functions with different amplitudes). We therefore apply periodic functions (sine and cosine) to the interpolated feature vectors before compositing them along the ray. Under this formulation, the activated features act as frequency components, while the primitive opacity modulates their amplitude (\cref{fig:method}). The resulting implicit signal in image space is sufficiently rich to be directly decoded using a lightweight MLP in a single evaluation per pixel, in a deferred-shading manner.

We validate \emph{Neural Harmonic Textures} on standard novel view synthesis benchmarks, achieving state-of-the-art performance among both real-time and offline methods (\cref{sec:results-nvs}). Our proposed formulation is agnostic to the underlying primitive type: triangles, 2D or 3D Gaussians, tetrahedra, and other elements can be used interchangeably, enabling seamless integration into existing primitive-based pipelines while significantly increasing their representational capacity (\cref{sec:other-applications}).
Finally, by abstracting the encoded signal, our approach naturally extends to higher-dimensional signals. Joint modeling of color and semantic features becomes possible within a unified local representation, allowing us to exploit both spatial and cross-modal correlations for improved efficiency and compactness (\cref{sec:app-semantic-field-reconstruction}).
\section{Related Work}
\label{sec_relatedwork}
\subsection{Neural Fields}
Neural fields have become a foundational model for representing complex, multi-scale signals across a wide range of domains~\cite{xie2022neural}, including audio~\cite{luo2022learning}, images~\cite{acorn,saragadam2023wire}, 3D geometry~\cite{park2019deepsdf,wang2021neus,takikawa2021nglod,li2023neuralangelo}, radiance fields~\cite{mildenhall2021nerf}, scientific data~\cite{wu2022instant,wu2024distributed,gadirov2025hyperflint}, and volumetric video\cite{pumarola2021d,fridovich2023k}. These methods primarily differ in how they encode their inputs and parameterize their outputs. In most formulations, latent features are stored in a compressed representation and subsequently decoded by a neural network. This storage can be implicit within the network weights~\cite{mildenhall2021nerf,gaussianscalespaces,Lombardi21,neuralVolumes,kilonerf,sitzmann2020implicit,tancik2020fourier}, or explicit through dedicated spatial data structures. Explicit representations typically offer improved scalability and performance, though often at increased memory cost. Common examples include voxel grids~\cite{acorn}, triplanes~\cite{chen2022tensorf,fridovich2023k}, point clouds~\cite{govindarajan2024lagrangian}, hash grids~\cite{mueller2022instant}, and meshes~\cite{thies2019}. In addition, explicit spatial structures are frequently used as acceleration mechanisms to skip empty regions and efficiently identify relevant query locations~\cite{mueller2022instant, neuralshells,neuralsparsevoxelfields}. More recently, neural fields have been combined with primitive-based methods by turning the primitives into tiny neural fields with closed-form antiderivatives, enhancing their individual modelling power~\cite{zhou2025splat}.

Beyond the choice of spatial representation, prior work has shown that positional encodings play a crucial role in enabling neural networks to represent high-frequency signals in low-dimensional spatial domains~\cite{tancik2020fourier,mildenhall2021nerf,mueller2022instant,sitzmann2020implicit}. 
In contrast to globally anchored spatial structures such as hierarchical voxel grids (hashgrid encodings), which follow an Eulerian formulation, our approach is Lagrangian. We anchor latent feature vectors relative to explicit particles, enabling adaptive spatial resolution and naturally supporting explicit motion, deformation, and scene editing.

\subsection{Primitive-based Representations for Neural Reconstruction}

Primitive-based methods have recently gained significant traction for 3D reconstruction and novel view synthesis, largely driven by the success of 3D Gaussian Splatting~\cite{kerbl20233Dgaussians}. 
Numerous extensions have since explored variants of Gaussian primitives~\cite{4dgaussians,betasplatting,3dgut,3dgrt,condor2025,condor2026, kulhanek2025lodge,2dgs} as well as alternative representations, including Voronoi cells~\cite{radiantfoam}, triangle soups~\cite{trianglesplatting}, and connected meshes~\cite{radiancemesh}. 
While offering high rendering performance, editability, and interpretability, these approaches typically exhibit poor memory scaling when fitting high-frequency detail due to their joint modeling of geometry and appearance.

Neural field methods, in contrast, aggregate information across the entire scene, enabling compact representations. 
Concurrent work~\cite{neuralshells,radiancevolumetricmeshes} attempts to bridge this gap by using geometric primitives as acceleration structures for neural field evaluation. 
However, these approaches lose some of the advantages of purely primitive-based representations, as the learned features remain globally anchored (e.g. through hashgrids) and do not follow the primitives under motion or deformation. Furthermore, they inherit the limitations of Eulerian encodings, namely poor scaling in large scenes and high-frequency detail modelling. In contrast, we propose that geometric primitives themselves, when appropriately parameterized, can serve as an effective positional encoding mechanism, and enable more efficient signal decoding (i.e. deferred shading).

\subsection{Deferred Shading and Texturing in Radiance fields}

Alternatives to direct color evaluation in novel view synthesis have been explored to improve expressivity, rendering quality, and efficiency. Early work introduced image-based neural deferred shading conditioned on 3D meshes~\cite{thies2019}. Later approaches extended this idea to neural radiance fields by baking appearance features into extracted geometry for real-time rendering~\cite{bakedsdf,chen2022mobilenerf,hedman2021snerg}, removing the necessity to run neural networks during runtime. However, these methods rely on complex multi-stage pipelines that first optimize a global field, then extract geometry and bake appearance features. In contrast, our method jointly optimizes primitives and neural appearance from scratch, eliminating the need for a separate baking stage while retaining high quality and speed. Deferred shading has also been explored in the context of radiance field rendering using neural networks as deferred shaders. In Han et al.~\cite{han2023}, global feature grids are queried, with features accumulated and decoded once in a deferred manner. Similarly, point-based neural renderers have recently mixed local neural decoding with deferred shading. Hahlbohm et al.~\cite{hahlbohm25} uses point-clouds as query points for a global hash-grid encoded feature field, decoded locally using a shallow neural network into implicit features. These are splatted to the camera plane and finally rendered through a convolutional neural network. 
However, the necessity of large-bandwidth neural networks to uplift the poor scalability and detail afforded by the spatial encoding made them slow and difficult to train; the former requiring guiding networks with regular NeRF-style rendering to steer the optimization, while the latter required expensive perceptual-loss supervision.
Earlier work on neural point-based graphics~\cite{aliev2020neural} and trilinear point splatting~\cite{franke2024trips} also attach local features to discrete primitives and decode them with neural networks. 
Kopanas et al.~\cite{kopanas2022catacaustics} combine per-point features with deferred shading for reflective content, while closer to our work Feature Splatting~\cite{martins2024featuresplat} also introduced per-primitive higher dimensional features to be decoded deferredly. 
Hybrid mesh and surface representations such as NeuMesh~\cite{yang2022neumesh} and SurRF~\cite{zhang2021surrf} similarly pair local learnable features with neural decoding for appearance modeling. Despite the large body of work combining explicit geometry, implicit fields or features and neural decoding, leveraging them efficiently without exploding memory or computational costs remains a challenge.

To improve expressivity without global feature grids, several primitive-based methods introduce per-primitive textures~\cite{bbsplat,xu2024texture,texturedgaussians,huang2024textured,wurster2024gabor,zhou20253dgabsplat}. 
In particular, BBSplat~\cite{bbsplat} and Textured Gaussians~\cite{texturedgaussians} further model opacity as a spatially varying texture on each primitive. Generally, however, highly expressive primitives complicate optimization and are prone to overfitting, apart from the substantial storage cost, which translates as well to slower rendering.
Instead, we decouple geometry and appearance by pairing local per-particle features with a shared lightweight neural decoder, in turn enabling the reduction of queries to the neural field to a single deferred pass. This allows particle density to more closely follow geometric complexity rather than being artificially increased to reproduce high-frequency appearance.
\section{Preliminaries}
\label{sec_preliminaries}

We review primitive-based radiance fields, primitive-based volume rendering, and positional encodings, focusing on the aspects most relevant to our method.
\paragraph{Primitive-based Radiance Fields.}

Primitive-based radiance fields represent complex scenes using an unstructured collection of 2D or 3D geometric primitives, such as anisotropic Gaussians~\cite{kerbl20233Dgaussians}, oriented disks~\cite{2dgs}, or triangles~\cite{trianglesplatting}. In~\cref{sec_method}, we present our formulation on the example of 3D Gaussian primitives~\cite{kerbl20233Dgaussians,3dgut} whose spatial response function 

\begin{equation}
\rho(\mathbf{x}) = \exp\!\left(-\frac{1}{2}(\mathbf{x}-\boldsymbol{\mu})^{T}\boldsymbol{\Sigma}^{-1}(\mathbf{x}-\boldsymbol{\mu})\right),
\label{eq:gaussian_kernel}
\end{equation}
is defined by the particle center $\boldsymbol{\mu}\in\mathbb{R}^3$ and covariance matrix $\boldsymbol{\Sigma}\in\mathbb{R}^{3\times3}$. 
To ensure positive semi-definiteness during optimization, the covariance is parameterized as $\boldsymbol{\Sigma} = \mathbf{R}\mathbf{S}\mathbf{S}^{T}\mathbf{R}^{T}$,
where $\mathbf{R}\in SO(3)$ represents rotation and $\mathbf{S}\in\mathbb{R}^{3\times3}$ encodes scaling. 
In practice, these parameters are stored as a quaternion $\mathbf{q}\in\mathbb{R}^4$ and a scale vector $\mathbf{s}\in\mathbb{R}^3$. 
Each primitive additionally carries an opacity parameter $\sigma \in \mathbb{R}$ and a view-dependent radiance function $\phi(\mathbf{d})$, commonly represented using spherical harmonics. Note that in this case, $\phi(\mathbf{d})$ depends only on the ray direction and not on the spatial position of the intersection point.

\paragraph{Primitive-based Volume Rendering.}
Given a camera ray $\mathbf{r}(\tau) = \mathbf{o} + \tau\mathbf{d}$ with origin $\mathbf{o} \in \mathbb{R}^3$ and direction $\mathbf{d} \in \mathbb{R}^3$, the rendered ray color $\mathbf{c}\in\mathbb{R}^3$ is obtained via volume rendering over the primitives $P_c$ intersecting the ray: 

\begin{equation}
\mathbf{c}(\mathbf{o},\mathbf{d}) 
    = \sum_{i \in P_c} \phi(\mathbf{d}) \cdot T_i \cdot \alpha_i,
\quad
T_i =
    \prod_{j=1}^{i-1} 1 -\alpha_j,
\label{eq:transmittance}
\end{equation}
where the opacity $\alpha_i$ is defined as $\alpha_i = \sigma_i \, \rho_i(\mathbf{o}+\tau\mathbf{d})$
and $T_i$ is the transmittance.

\paragraph{Positional Encoding:}
Neural networks exhibit a natural spectral bias that limits their ability to learn high-frequency functions in low-dimensional domains~\cite{tancik2020fourier}. Positional encodings mitigate this limitation by mapping low-dimensional spatial and directional coordinates into higher-dimensional representations before feeding them into the network~\cite{tancik2020fourier}. While early approaches relied on fixed \emph{frequency encodings}~\cite{harris2010digital,muller2019neural}, modern methods often use \emph{parametric encodings} that store learnable feature vectors in auxiliary spatial data structures, such as grids or trees~\cite{mueller2022instant,takikawa2021nglod}. Query coordinates retrieve and interpolate these features, effectively shifting much of the representational burden from the neural network to the encoding itself. This design trades increased memory usage for reduced computation: only a small subset of encoding parameters is updated per sample, allowing the neural network to remain compact and efficient while significantly accelerating convergence and maintaining high reconstruction quality.
\begin{figure*}[t]
    \includegraphics[width=\linewidth]{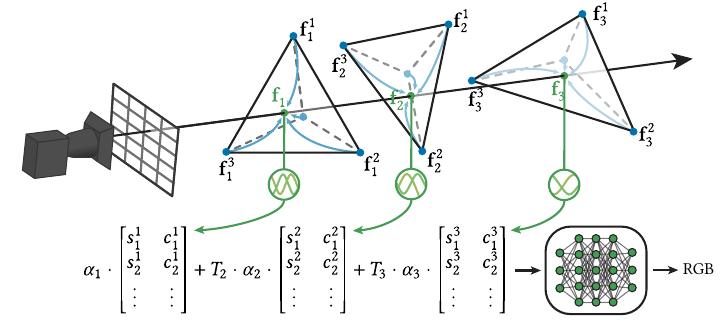}
    \caption{
        Neural Harmonic Textures applied to novel-view synthesis. We virtually attach feature vectors $\mathbf{f}_i$ to the vertices of tetrahedra inscribing the Gaussian primitives\protect\footnotemark.
        Following 3DGUT~\cite{3dgut}, we evaluate the point along the ray where the projected Gaussian has maximum response. 
        We barycentrically interpolate vertex features at that point, and encode them with sine and cosine functions into different channels. 
        These are then alpha blended along the rest of the ray, until the resulting sum of harmonics is decoded by a shallow MLP in a single image-space pass.
    }
    \label{fig:method}
\end{figure*}
\footnotetext{
    Gaussian particles are bounded by ellipsoids in world space, which become spheres after whitening.
    Our virtual bounding tetrahedra are defined in this canonical space.
}

\section{Method}
\label{sec_method}

We present the core components of our approach using 3D Gaussian primitives in the context of novel view synthesis, following the formulation of 3DGUT~\cite{3dgut}. We begin by introducing our \emph{primitive-bound feature embedding} in \cref{sec:representation}. Next, \cref{sec:encoding} presents \emph{harmonic texturing}, which increases per-primitive expressive power while preserving the locality and explicit form of the representation. Finally, \cref{sec:decoding} describes a \emph{neural deferred decoding} scheme that enables efficient signal reconstruction in a single image-space pass. Generalization to other primitive types and additional applications will be discussed in \cref{sec:other-applications}. \cref{fig:method} provides a schematic overview of our method.

\begin{figure*}[t]
    \centering
    \begin{subfigure}{0.32\textwidth}
        \includegraphics[width=\textwidth]{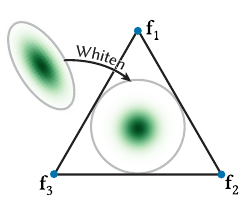}
        \subcaption{Latent vectors}
    \end{subfigure}
    \hfill
    \begin{subfigure}{0.32\textwidth}
        \includegraphics[width=\textwidth]{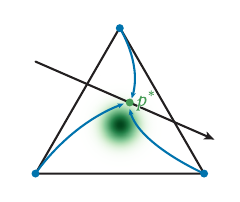}
        \subcaption{Interpolation}
    \end{subfigure}
    \hfill
    \begin{subfigure}{0.32\textwidth}
        \includegraphics[width=\textwidth]{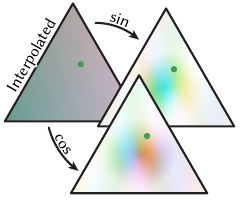}
        \subcaption{Harmonic decomposition}
    \end{subfigure}
    \caption{
        Illustrating our method in 2D.
        Each primitive is bounded by an ellipsoid in world space, which becomes a sphere in whitened canonical space \textbf{(a)}.
        Considering a virtual bounding tetrahedron in this canonical space, we attach one N-dimensional feature vector $\mathbf{f}^j$ to each vertex.
        The primitive's contribution is evaluated at the point of maximum response $\mathbf{p}^*$ of the projected Gaussian along the intersecting ray \textbf{(b)}.
        The feature vectors are barycentrically interpolated at $\mathbf{p}^*$ and encoded with sine and cosine periodic functions \textbf{(c)}.
    }
    \label{fig:gaussian_tetrahedron}
\end{figure*}

\subsection{Primitive-bound feature embedding}
\label{sec:representation}

Spatial structures such as triplanes \cite{duckworth2023smerf}, voxel grids \cite{SunSC22}, and multi-resolution hash grids~\cite{mueller2022instant} are commonly used in neural fields to hold latent feature vectors that embed query positions into a higher-dimensional space before decoding the signal. Despite their effectiveness, these representations rely on globally defined regular grids, which limits their scalability to large scenes or high-frequency detail. Moreover, because these structures are fixed in space, they struggle to represent scenes undergoing motion, deformation, or editing.

Instead, we exploit the Lagrangian nature and adaptivity of primitive-based representations by anchoring latent features to a virtual scaffold that encapsulates each primitive. 
Specifically, consider the isosurface ellipsoid defined by an anisotropic 3D Gaussian (\cref{eq:gaussian_kernel}) with bounded support, which becomes a sphere in canonical space after whitening. We define the virtual tetrahedron bounding this sphere and assign a feature vector $\mathbf{f}^j \in \mathbb{R}^{N_f}$ to each of its four vertices ($j \in \{0..3\}$) (\cref{fig:gaussian_tetrahedron}a). For each primitive intersected by a ray, we determine the point $\mathbf{p}^*$ corresponding to the maximum Gaussian response along the ray~\cite{3dgut}. The feature vector at this location, $\mathbf{f}$, is obtained via barycentric interpolation of the vertex features $\mathbf{f}^j$. A detailed derivation is provided in the supplementary material.

Compared to globally anchored feature representations, our formulation moves with the primitives and therefore naturally supports operations such as motion, deformation, or deletion during scene editing.

\subsection{Harmonic Texturing}
\label{sec:encoding}
\begin{wrapfigure}[14]{r}{0.35\textwidth}
    \vspace{-0.4\baselineskip}
    \includegraphics[width=\linewidth]{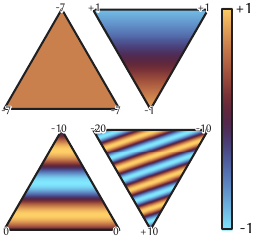}
    \caption{Harmonic textures.}
    \label{fig:harmonics}
    \vspace{0.4\baselineskip}
\end{wrapfigure}
In previous work, feature vectors are typically passed through a lightweight neural network to decode the primitive appearance \emph{prior} to volume rendering, requiring dozens of MLP evaluations per ray to produce the final pixel color~\cite{mueller2022instant,nestsplatting,radiancemesh}. In 3DGS, color decoding overhead (without neural networks) is reduced by approximating view-dependent color emission once per primitive, thereby reducing the complexity from the number of ray–primitive intersections to the number of primitives. However, this approximation assumes that the signal does not vary spatially within each primitive, which makes it unsuitable for our representation.

An alternative strategy is to blend the features along each ray and decode the signal in image space, in the spirit of deferred shading~\cite{hedman2021snerg}. Inspired by the Fourier transform, where a complex signal is expressed as a sum of harmonic components (i.e., periodic functions of different amplitudes), we encode the interpolated features using periodic functions before blending them along the ray. This yields a harmonic decomposition of the signal, which we term \emph{Harmonic Textures} (\cref{fig:gaussian_tetrahedron}c). Under this formulation, the interpolation function effectively acts as a frequency modulator: large differences between vertex features within a primitive produce rapidly oscillating, spatially varying textures. Each primitive additionally has a kernel-weighted opacity, which acts as the harmonic amplitude. This behavior is illustrated in the inset~\cref{fig:harmonics} (without the amplitude weighting, to better visualize the harmonics).

\subsection{Neural Deferred Shading}
\label{sec:decoding}
The rich high-dimensional signal yielded by the volume rendering of \emph{Harmonic Textures} enables a more efficient decoding strategy. We concatenate the accumulated harmonics with the ray direction $\mathbf{d}$ encoded using second-degree spherical harmonics $\mathrm{SH}_2(\mathbf{d}) \in \mathbb{R}^9$ as done in \cite{mueller2022instant}, and decode the final pixel color $\mathbf{c}$ using a shallow MLP, without further positional encoding. The resulting rendering equation is
\begin{equation}
\mathbf{c} =
\mathrm{MLP}_\theta \left(
    \sum_{i \in \mathcal{G}}
        \alpha_i \, T_i
        \begin{bmatrix}
            \sin(\mathbf{f}_i) \\
            \cos(\mathbf{f}_i)
        \end{bmatrix},
    \;
    k \cdot \mathrm{SH}_2(\mathbf{d})
\right),
\end{equation}
where $\mathbf{f}_i = \mathrm{interpolate}\!\left(
\left[\mathbf{f}_i^0, \mathbf{f}_i^1, \mathbf{f}_i^2, \mathbf{f}_i^3\right], \mathbf{p}_i^*
\right)$ denotes the primitive feature interpolated at the point $\mathbf{p}_i^*$ of maximum response, $\mathcal{G}$ is the set of primitives intersected by the ray, and $\alpha_i$ and $T_i$ correspond to the primitive opacity and accumulated transmittance defined in \cref{eq:transmittance}.
\section{Results}
\label{sec:results}

We present our results for radiance-field novel-view synthesis. Additionally, we demonstrate our method on two other reconstruction tasks: semantic scene reconstruction and high-resolution image fitting.

\subsection{Radiance Field Reconstruction}
\label{sec:results-nvs}

\subsubsection{Optimization}
We directly adopt the densification strategy, loss function and regularization terms of 3DGS-MCMC~\cite{3dgsmcmc}. Specifically, they define the regularization terms on primitive opacity and scale as
\begin{equation}
    \mathcal{R}_{\alpha} = \frac{1}{P}\sum_{i=1}^{P} \alpha_i\,, \qquad
    \mathcal{R}_{s} = \frac{1}{P}\sum_{i=1}^{P} \lVert \mathbf{s}_i \rVert_1\,,
\end{equation}
where $P$ is the number of primitives and $\mathbf{s}_i$ is the scale vector of primitive $i$.
Our final loss function is the same as 3DGS-MCMC:
\begin{equation}
    \mathcal{L} = (1 - \lambda)\,\mathcal{L}_{\mathrm{L}_1} + \lambda\,\mathcal{L}_{\text{D-SSIM}} + \lambda_{\alpha}\,\mathcal{R}_{\alpha} + \lambda_{s}\,\mathcal{R}_{s}\,.
\end{equation}
Similarly to the exponential scheduling applied to primitive positions~\cite{kerbl20233Dgaussians}, we use a cosine annealing scheduler for the feature and MLP learning rates.
We also apply an Exponential Moving Average (EMA) filter~\cite{mueller2022instant} to the MLP weights~$\theta$.
This enhances robustness to noise and avoids overfitting to individual frames:
\begin{equation}
    \bar{\theta}_{t} \leftarrow \gamma\, \bar{\theta}_{t-1} + (1 - \gamma)\, \theta_{t}\,,
\end{equation}
where $\bar{\theta}_{t}$ denotes the filtered weights at step $t$ and $\gamma$ is the decay factor.
Finally, we dedicate the last 3000 training iterations to optimize only the feature and MLP weights, disabling all regularization terms and freezing all other parameters. 
We found that this refinement step slightly improves color fidelity, particularly in large-scale scenes. Other hyperparameters are included in the supplementary material.
\subsubsection{Implementation}
\label{sec_implementation_details}
We implement our method in \emph{gsplat}~\cite{ye2025gsplat}, using the 3DGUT~\cite{3dgut} formulation, to which we add custom CUDA kernels for the forward and backward logic described in~\cref{sec_method}. To reduce register pressure, we use half-precision memory fetches on feature vectors during rasterization, for both forward and backward kernels. 
Our neural network uses tiny-cuda-nn's~\cite{mueller2022instant} JIT-compiled cooperative vector MLPs, trained and evaluated in half precision (FP16).
We apply an automatic scaling factor to the loss values in order to improve half-precision training stability~\cite{micikevicius2017mixed}. 

\subsubsection{Results}
\begin{figure}
    \centering
    \includegraphics{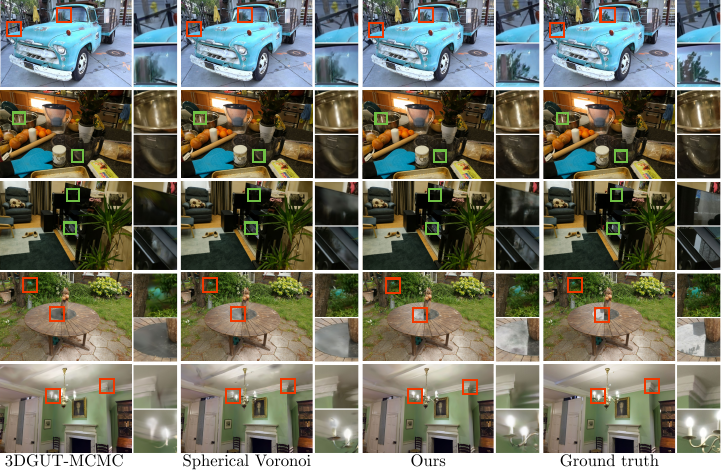}
    \caption{Comparison between our and previous works on radiance field reconstruction on scenes from MipNeRF360~\cite{barron2022mipnerf360}, Tanks and Temples~\cite{Knapitsch2017}. Our method models high frequency detail and view dependent effects to a higher degree than previous works.
    }
    \label{fig:inset_comparison}
\end{figure}

We present the results of our method against a number of previous works using pure neural fields~\cite{barron2022mipnerf360}, hashgrid-encoded neural fields~\cite{mueller2022instant,zipnerf}, pure primitive-based methods~\cite{2dgs, 3dgsmcmc,3dgut, betasplatting, trianglesplatting} with regular Spherical Harmonics (SH) appearance or more complex appearance models~\cite{sphericalVoronoi,texturedgaussians}, and closer to us, methods mixing primitives (e.g. for acceleration) and neural fields~\cite{nestsplatting, radiancemesh} (\cref{tab:extended_comparison}). In this table, we use the originally provided JPEG images for proper comparison with previous works in MipNeRF360; every other result presented in this section, unless otherwise stated, uses \emph{gsplat}'s default downscaling for all methods tested. We measure the performance delta of this choice in the supplementary (\cref{tab:jpeg_vs_downscaled}). We consistently outperform all previous works in a varied set of real datasets~\cite{barron2022mipnerf360, Knapitsch2017,hedman2018deep}. We include a selection of views in \cref{fig:inset_comparison}. Our method particularly excels at high-frequency detail modelling, capturing specular highlights and reflections with superior fidelity.

\begin{table*}[t]
  \centering
  \caption{Comparison on MipNeRF360~\cite{barron2022mipnerf360}, Tanks \& Temples~\cite{Knapitsch2017}, and Deep Blending~\cite{hedman2018deep}, on Neural Field-style methods,
  primitive-based methods, and mixed neural field/primitive-based methods. For the MipNeRF360 datasets, we disable \emph{gsplat}’s default downscaling and instead train and evaluate directly on the provided JPEG-compressed reference images following prior work. We measure the effect in \cref{tab:jpeg_vs_downscaled}. For our method, we use 64 features per primitive (16 per vertex), with a 128-wide $\times$ 3 hidden layers MLP, which results in a roughly similar number of total parameters as previous primitive-based approaches on average. We use 2M primitives for indoor scenes and 5M for outdoors. To regularize training in high-parallax regimes (e.g., indoor datasets), we encode the central camera ray rather than the individual ray direction. Finally, to ensure similar training epochs, we train indoor datasets longer (45k iterations for $\sim290$ images) than outdoor (25k for $\sim175$ images), following Spherical Voronoi~\cite{sphericalVoronoi} (SV). Unlike SV, we do not fine-tune learning rates per dataset. Note that convergence of an MLP-based appearance model can be slower than SH-based ones. 
  }
  \label{tab:extended_comparison}
  \setlength{\tabcolsep}{3pt}
  \resizebox{\textwidth}{!}{%
  \begin{tabular}{l*{3}{c}*{3}{c}*{3}{c}}
    \toprule
    & \multicolumn{3}{c}{MipNeRF360}
    & \multicolumn{3}{c}{Tanks \& Temples}
    & \multicolumn{3}{c}{Deep Blending} \\
    \cmidrule(lr){2-4} \cmidrule(lr){5-7} \cmidrule(lr){8-10}
    Method
      & PSNR$\uparrow$ & SSIM$\uparrow$ & LPIPS$\downarrow$
      & PSNR$\uparrow$ & SSIM$\uparrow$ & LPIPS$\downarrow$
      & PSNR$\uparrow$ & SSIM$\uparrow$ & LPIPS$\downarrow$ \\
    \midrule
    Instant NGP-Big~\cite{mueller2022instant}
      & 25.59 & 0.695 & 0.375 & 21.92 & 0.740 & 0.342 & 24.96 & 0.815 & 0.459 \\
    Mip-NeRF 360~\cite{barron2022mipnerf360}
      & 27.60 & 0.788 & 0.275 & 22.22 & 0.754 & 0.290 & 29.40 & 0.899 & \cellcolor{thirdcell}0.306 \\
    ZipNeRF~\cite{zipnerf}
      & \cellcolor{thirdcell}28.55 & 0.829 & \cellcolor{secondcell}0.218 & 23.64 & 0.836 & 0.179 & -- & -- & -- \\
    \midrule
    2DGS~\cite{2dgs}
      & 27.22 & 0.804 & 0.275 & 22.85 & 0.827 & 0.244 & 29.56 & 0.904 & 0.325 \\
    3DGS-MCMC~\cite{3dgsmcmc}
      & 27.99 & 0.830 & 0.229 & 24.46 & \cellcolor{thirdcell}0.866 & \cellcolor{thirdcell}0.174 & 29.49 & 0.912 & \cellcolor{thirdcell}0.306 \\
    3DGUT-MCMC~\cite{3dgsmcmc}
      & 27.82 & 0.826 & 0.233 & 24.20 & 0.861 & 0.180 & \cellcolor{thirdcell}29.87 & \cellcolor{thirdcell}0.913 & 0.309 \\
    Beta Splatting-MCMC~\cite{betasplatting}
      & 28.12 & \cellcolor{thirdcell}0.831 & 0.238 & \cellcolor{thirdcell}24.54 & 0.866 & 0.196 & 29.56 & 0.907 & 0.316 \\
    Spherical Voronoi~\cite{sphericalVoronoi}
      & \cellcolor{secondcell}28.56 & \cellcolor{bestcell}\textbf{0.835} & \cellcolor{thirdcell}0.228 & \cellcolor{secondcell}24.80 & \cellcolor{secondcell}0.871 & \cellcolor{secondcell}0.172 & \cellcolor{secondcell}30.34 & \cellcolor{secondcell}0.914 & \cellcolor{bestcell}\textbf{0.299} \\
    Triangle Splatting~\cite{trianglesplatting}
      & 27.00 & 0.808 & 0.231 & 23.05 & 0.843 & 0.191 & 28.92 & 0.891 & 0.308 \\
    Textured Gaussians~\cite{texturedgaussians}
      & 27.35 & 0.827 & -- & 24.26 & 0.854 & -- & 28.33 & 0.891 & -- \\
    \midrule
    NeST~\cite{nestsplatting}
      & 26.54 & 0.776 & 0.260 & -- & -- & -- & -- & -- & -- \\
    Radiance Meshes~\cite{radiancemesh}
      & 27.15 & 0.810 & 0.274 & 23.13 & 0.851 & 0.200 & 29.39 & 0.901 & 0.362 \\
    Neural Harmonic Textures (Ours)
      & \cellcolor{bestcell}\textbf{28.74} & \cellcolor{secondcell}0.834 & \cellcolor{bestcell}\textbf{0.216} & \cellcolor{bestcell}\textbf{25.68} & \cellcolor{bestcell}\textbf{0.882} & \cellcolor{bestcell}\textbf{0.141} & \cellcolor{bestcell}\textbf{30.94} & \cellcolor{bestcell}\textbf{0.919} & \cellcolor{secondcell}0.302 \\
    \bottomrule
  \end{tabular}%
  }
\end{table*}

In order to isolate the impact of our method, we include a controlled experiment in \cref{tab:comparison}. We compare between 3DGS-MCMC, 3DGUT-MCMC, and different appearance models for the latter: regular SH, current state-of-the-art Spherical Voronoi~\cite{sphericalVoronoi} (SV) and ours (NHT). We implement all four under the same framework (\emph{gsplat}), 1M primitives cap, 30k iterations, identical per-scene hyperparameters, and the same opacity and scale regularizers. This isolates the appearance model from training-schedule choices in \cref{tab:extended_comparison}. \cref{tab:comparison} shows that our method consistently outperforms both SH and SV across all benchmarks, while using less storage, without sacrificing real-time performance. NHT takes around 14.5 minutes to train up to 30k iterations on MipNeRF360 on average (1M primitives, RTX 5090); training times for all methods across datasets are reported in the supplementary material (\cref{tab:training_times}). Per-scene breakdowns and more details are provided in the supplementary material (\cref{tab:perscene,tab:ours_perscene}).
We also compare against a close texture-based method (BBSplat~\cite{bbsplat}) in the supplementary material (\cref{tab:bbsplat}). NHT achieves substantially higher quality and faster rendering not only at equal memory, but also at equal primitive count.
\begin{table*}[t]
  \centering
  \caption{Quantitative results of our approach and baselines on the MipNeRF360~\cite{barron2022mipnerf360}, Tanks \& Temples~\cite{Knapitsch2017}, and Deep Blending~\cite{hedman2018deep} datasets, comparing our method against Spherical Voronoi~\cite{sphericalVoronoi} (SV) and regular SH models. We isolate the effect of our approach by implementing all methods in the same framework (\emph{gsplat}~\cite{gsplat} with its default downsampling), using the same number of primitives (1M), training for the same number of iterations (30k), allocating the same number of parameters per primitive for appearance (48) and using the same hyperparameters for all scenes. Our method uses a 128$\times$3 hidden MLP. All MCMC-based methods share the same opacity and scale regularizers from 3DGS-MCMC~\cite{3dgsmcmc}. We improve reconstruction quality while still managing real-time performance (RTX A6000 Ada).}
  \label{tab:comparison}
  \setlength{\tabcolsep}{3pt}
  \resizebox{\textwidth}{!}{%
  \begin{tabular}{l*{4}{c}*{4}{c}*{4}{c}}
    \toprule
    & \multicolumn{4}{c}{MipNeRF360}
    & \multicolumn{4}{c}{Tanks \& Temples}
    & \multicolumn{4}{c}{Deep Blending} \\
    \cmidrule(lr){2-5} \cmidrule(lr){6-9} \cmidrule(lr){10-13}
    Method (w/ MCMC)
      & PSNR$\uparrow$ & SSIM$\uparrow$ & LPIPS$\downarrow$ & FPS$\uparrow$
      & PSNR$\uparrow$ & SSIM$\uparrow$ & LPIPS$\downarrow$ & FPS$\uparrow$
      & PSNR$\uparrow$ & SSIM$\uparrow$ & LPIPS$\downarrow$ & FPS$\uparrow$ \\
    \midrule
    3DGS + SH
      & \cellcolor{thirdcell}27.94 & \cellcolor{secondcell}0.829 & \cellcolor{secondcell}0.246 & \cellcolor{bestcell}\textbf{251}
      & \cellcolor{secondcell}24.25 & \cellcolor{secondcell}0.861 & \cellcolor{thirdcell}0.188 & \cellcolor{bestcell}\textbf{294}
      & 29.98 & \cellcolor{thirdcell}0.912 & \cellcolor{secondcell}0.317 & \cellcolor{bestcell}\textbf{331} \\
    3DGUT + SH
      & 27.93 & \cellcolor{thirdcell}0.828 & \cellcolor{thirdcell}0.247 & 201
      & 23.99 & 0.859 & 0.192 & \cellcolor{secondcell}245
      & \cellcolor{thirdcell}30.21 & \cellcolor{secondcell}0.913 & \cellcolor{thirdcell}0.318 & \cellcolor{secondcell}282 \\
    3DGUT + SV
      & \cellcolor{secondcell}28.15 & 0.823 & 0.248 & \cellcolor{secondcell}202
      & \cellcolor{thirdcell}24.18 & \cellcolor{thirdcell}0.861 & \cellcolor{secondcell}0.187 & 242
      & \cellcolor{secondcell}30.29 & 0.912 & 0.320 & 267 \\
    3DGUT + NHT (Ours)
      & \cellcolor{bestcell}\textbf{28.46} & \cellcolor{bestcell}\textbf{0.830} & \cellcolor{bestcell}\textbf{0.232} & 140
      & \cellcolor{bestcell}\textbf{24.79} & \cellcolor{bestcell}\textbf{0.875} & \cellcolor{bestcell}\textbf{0.169} & 226
      & \cellcolor{bestcell}\textbf{30.88} & \cellcolor{bestcell}\textbf{0.918} & \cellcolor{bestcell}\textbf{0.311} & 240 \\
    \bottomrule
  \end{tabular}%
  }
\end{table*}

\begin{table*}[t]
    \centering
    \begin{minipage}[t]{0.56\textwidth}
        \centering
        \caption{Generality of NHT: we apply it to different primitive-based representations, evaluated on MipNeRF360~\cite{barron2022mipnerf360}. Triangle Splatting methods use JPEG references as in its source implementation. We would like to note that our results for Triangle Splatting are just a proof-of-concept and were not extensively optimized.}
        \label{tab:generality}
        \vspace{0.5em}
        \resizebox{\linewidth}{!}{
  \setlength{\tabcolsep}{5pt}
  \begin{tabular}{lccc}
    \toprule
    Method & PSNR$\uparrow$ & SSIM$\uparrow$ & LPIPS$\downarrow$ \\
    \midrule
    Triangle Splatting
      & 27.00 & \cellcolor{bestcell}\textbf{0.808} & 0.231 \\
    Triangle Splatting + NHT
      & \cellcolor{bestcell}\textbf{27.52} & 0.807 & \cellcolor{bestcell}\textbf{0.227} \\
    \midrule
    2DGS
      & 27.48 & 0.816 & 0.263 \\
    2DGS + NHT
      & \cellcolor{bestcell}\textbf{28.27} & \cellcolor{bestcell}\textbf{0.820} & \cellcolor{bestcell}\textbf{0.238} \\
    \midrule
    3DGUT-MCMC
      & 27.93 & 0.828 & 0.247 \\
    3DGUT-MCMC + NHT
      & \cellcolor{bestcell}\textbf{28.46} & \cellcolor{bestcell}\textbf{0.830} & \cellcolor{bestcell}\textbf{0.232} \\
    \bottomrule
  \end{tabular}

        }
    \end{minipage}
    \hfill
    \begin{minipage}[t]{0.40\textwidth}
        \centering
        \vspace{0.5em}
        \includegraphics[width=\linewidth]{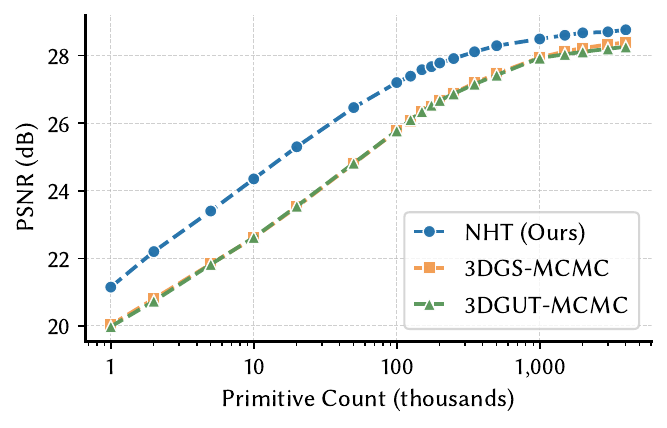}
        \captionof{figure}{
            Our method outperforms 3DGS and 3DGUT at all primitive counts. The improvement is particularly pronounced in the low-primitive regime ($\leq 100k$), with deltas upwards of 2dB of PSNR.
        }
        \label{fig:compactness_study}
    \end{minipage}
\end{table*}

\subsubsection{Ablations}
\label{sec:ablations}
We ablate the design of our method on the MipNeRF360 dataset.

\paragraph{Optimization choices.}
Supplementary~\cref{tab:ablation_opt} isolates the individual contributions of the training strategies: learning rate scheduling, EMA on the MLP weights, loss scaling, the color-refinement phase, and opacity and scale regularization.

\begin{figure}[t]
    \centering
    \begin{minipage}[t]{0.48\linewidth}
        \centering
        \includegraphics[width=\linewidth]{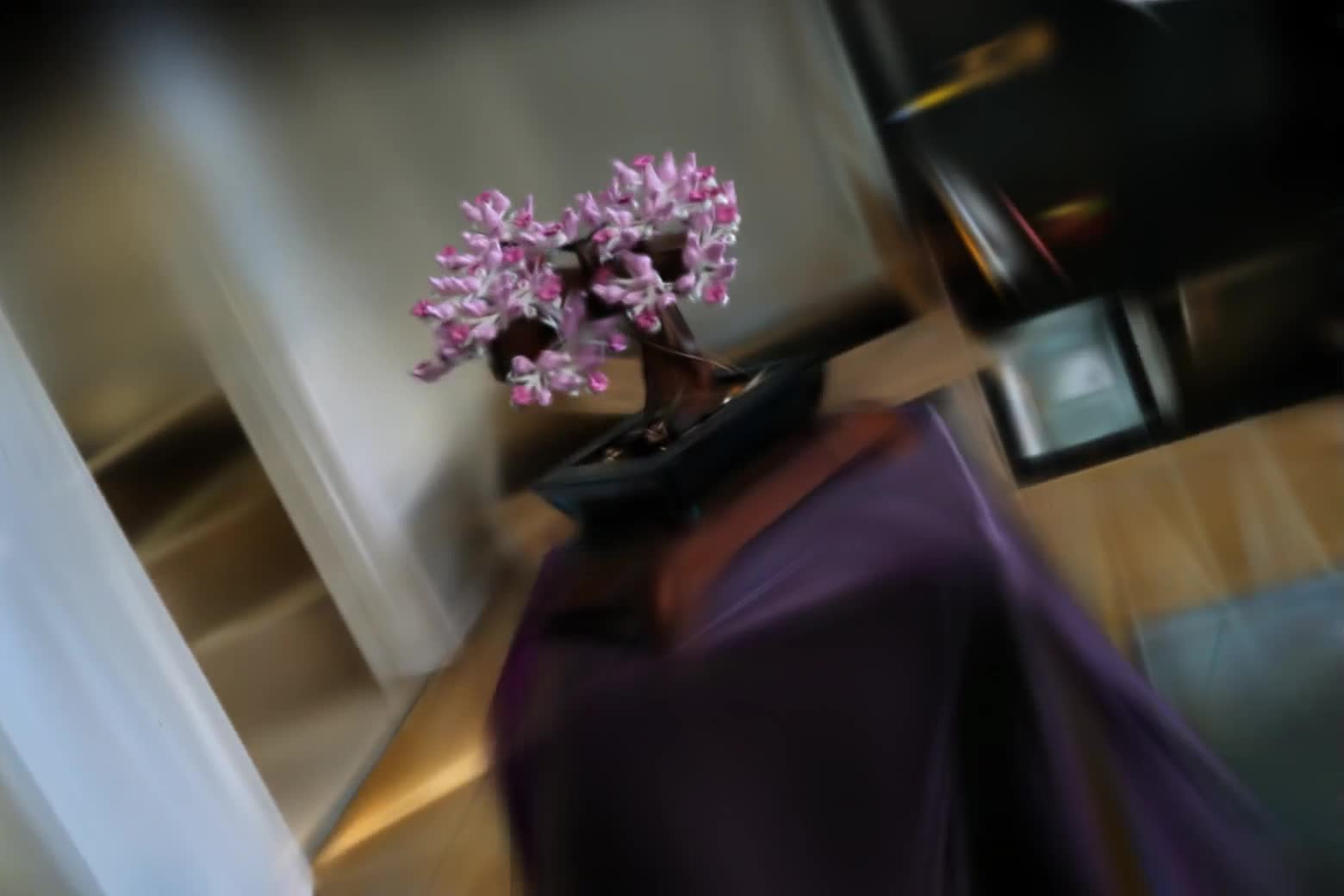}
        \centerline{\small 3DGUT-MCMC}
        \centerline{\small 23.97\,dB / 0.724 / 0.548}
    \end{minipage}
    \hfill
    \begin{minipage}[t]{0.48\linewidth}
        \centering
        \includegraphics[width=\linewidth]{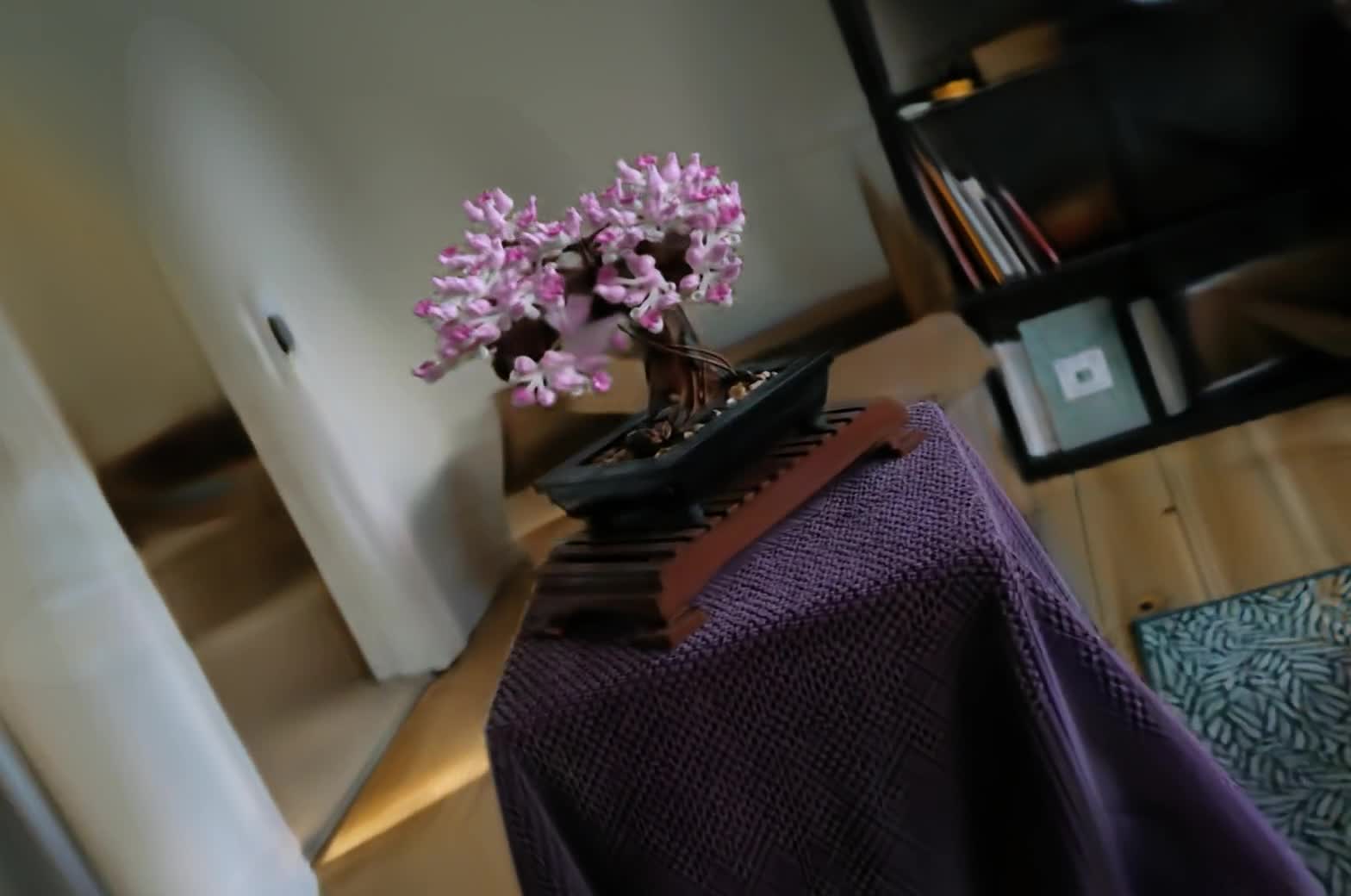}
        \centerline{\small Neural Harmonic Textures (Ours)}
        \centerline{\small 27.63\,dB / 0.856 / 0.410}
    \end{minipage}
    \caption{Qualitative comparison on MipNeRF360 \emph{bonsai} with 10K Gaussians. Metrics are PSNR / SSIM / LPIPS averages over the test set. NHT recovers substantially more detail than 3DGUT-MCMC at equal primitive budget (+3.7\,dB PSNR).}
    \label{fig:bonsai_10k}
\end{figure}

\paragraph{Primitive count.}
\cref{fig:compactness_study} illustrates how reconstruction quality scales with the number of primitives, ranging from 1K to 4M. Our method improves over pior art across this entire range, excelling particularly in the low primitive count regime. For instance, we achieve performance comparable to previous methods using 1M primitives with only a third of that amount, offering more flexibility in the memory-speed-quality trade-off; Supplementary~\cref{fig:bonsai_10k} shows a side-by-side comparison against 3DGUT-MCMC on \emph{bonsai} with only 10K primitives (+3.7\,dB PSNR on average).
\cref{tab:ablation_featuredim,tab:ablation_modelsize} in the supplementary material explore the impact of varying the per-primitive feature dimension and the MLP architecture, respectively, to illustrate the resulting quality-speed trade-offs.

\paragraph{Feature encoding.}
We ablate the choice of feature encoding function applied to the interpolated features in the Supplementary \cref{tab:ablation_encoding}, showing how our choice of harmonic functions provides the best results.

\subsection{Other Applications}
\label{sec:other-applications}

\paragraph{Alternative primitive-based methods.}
We demonstrate the generality of our approach by incorporating Neural Harmonic Textures into 2DGS~\cite{2dgs} and Triangle Splatting~\cite{trianglesplatting}.
Our method is readily adapted to these 2D primitives by simply replacing the bounding tetrahedra with virtual triangles (2DGS) or directly anchoring feature vectors to explicit triangle geometry (Triangle Splatting), therefore using three feature vectors per primitive instead of four.
The results are shown in~\cref{tab:generality}, with further details in the Supplementary.

\paragraph{Semantic field reconstruction.}
\label{sec:app-semantic-field-reconstruction}
Our method can easily model higher dimensional data as well. 
By simply expanding the decoding head of our MLP, we can fit signals of arbitrary dimension.
As the rest of the MLP layers remain small, and the size of feature vectors remains the same, the model is forced to exploit potential correlations between dimensions. 
A particularly challenging case is semantic scene reconstruction, where fields are upwards of 512-dimensional. 3DGS-based methods relying on high-dimensional explicit features generally struggle with such scale, even when paired with feature upsampling networks.
We thus test our method on joint novel-view synthesis of RGB radiance and LSEG~\cite{LSEG} 512-wide semantic features.
Due to the limited resolution of LSEG feature maps, we render and predict at the RGB resolution, but bilinearly downsample the spatial dimension of the resulting feature maps in order to supervise on ground truth features. 
We compare against Feature 3DGS~\cite{feature3dgs}, a hybrid method relying on 128-wide semantic feature vector per primitive plus a CNN to jointly downsample the spatial dimension and upsample the feature dimension of the resulting rasterized image up to the full 512 dimensions, and another set of 48 SH basis weights for regular RGB color. Results for the aggregated MipNeRF360 dataset can be seen in~\cref{tab:lseg_comparison}. 
We afford Feature-3DGS twice our training time budget, up to 7K iterations, slightly more than in the original paper (5K), to accommodate for the larger-scale datasets. We further ablate the per-primitive feature dimensionality for this task in \cref{tab:ablation_lseg_featuredim}. Our results show substantial improvement over Feature 3DGS, and the capability to easily scale upwards or downwards in performance or memory, in a completely detached manner from the nature of the data being encoded. It is particularly interesting to note how small the drop in RGB quality is, which points to either high correlation among feature vectors or excessive free bandwidth in our RGB experiments. Our features can then be used in downstream tasks such as scene editing or semantic segmentation.

\begin{table}[t]
  \centering
  \caption{Joint RGB and LSEG semantic feature reconstruction on MipNeRF360~\cite{barron2022mipnerf360}, (RTX A6000 Ada). LSEG$_\text{P}$ and LSEG$_\text{C}$ denote PSNR and cosine similarity of the 512-dim semantic features, respectively. We use 80 features per primitive and a 128$\times$3 MLP, while Feature 3DGS uses 176 features per primitive, 512$\times$128 CNN and 1.5$\times$ the number of primitives we use, resulting in an approximately 3$\times$ higher memory footprint.}
  \label{tab:lseg_comparison}
  \setlength{\tabcolsep}{4pt}
  \resizebox{\linewidth}{!}{
  \begin{tabular}{lcccccc}
    \toprule
    Method
      & RGB$_\text{PSNR}$$\uparrow$
      & RGB$_\text{SSIM}$$\uparrow$
      & RGB$_\text{LPIPS}$$\downarrow$
      & LSEG$_\text{P}$$\uparrow$
      & LSEG$_\text{C}$$\uparrow$
      & FPS$\uparrow$ \\
    \midrule
    Feature 3DGS~\cite{feature3dgs}
      & 26.27
      & 0.785
      & 0.262
      & 43.68
      & 0.985
      & 3.02 \\
    Neural Harmonic Textures (Ours)
      & \cellcolor{bestcell}\textbf{28.16}
      & \cellcolor{bestcell}\textbf{0.824}
      & \cellcolor{bestcell}\textbf{0.228}
      & \cellcolor{bestcell}\textbf{46.90}	
      & \cellcolor{bestcell}\textbf{0.993}
      & \cellcolor{bestcell}\textbf{28.11} \\
    \bottomrule
  \end{tabular}}
\end{table}

\paragraph{2D image reconstruction.}
\label{sec:app-image-reconstruction}
Finally, we show results on 2D image fitting. 
To adapt our method to this application, we make two key changes with respect to the primitive-based approaches. 
First, we employ a connected 2D triangle mesh topology.
This mesh is fully opaque: there is no alpha blending, no Gaussian or window kernel, simply interpolation of local features for a given pixel.
Second, we change our barycentric interpolation to Clough-Tocher cubic interpolation. 
This makes interpolated features continuous over the triangle edges, which removes high frequency artifacts. 
The rest of the architecture is the same as described in~\cref{sec_method}, and we include further details on the optimization strategy in the Supplementary.
We evaluate image reconstruction performance on a new curated dataset of high-resolution (45.7\,MP) 14-bit HDR RAW images, and compare it against Instant NGP~\cite{mueller2022instant} and the JPEG-XL encoder~\cite{jpegxl}, which supports high-dynamic range.
We measure reconstruction quality at two compression ratios ($10\times$ and $100\times$), computing PSNR in $\mu$-law transformed space (PSNR$_\mu$), standard in HDR imaging pipelines. 
We also report metrics on tonemapped images (subscript ``tm'').
Results are shown in~\cref{tab:results_images}.
Our method achieves comparable pixel-error quality to Instant NGP in both HDR and linear spaces, while substantially reducing perceptual error: at $100\times$ compression, NHT achieves an LPIPS of 0.048 compared to 0.078 for Instant NGP, a $38\%$ relative improvement. 
JPEG-XL however still comes out on top, particularly in perceptual metrics, which is to be expected given its human-perception-based inductive biases.
We can further compress our representation by a factor of 3 without noticeable loss in quality by aggressively reducing floating point precision of the implicit features, among other strategies. We include more details on this in the supplementary.

\begin{table}[t]
  \centering
  \caption{High-resolution (45.7\,MP) 14-bit HDR RAW image fitting, averaged over 15 images. We report PSNR in $\mu$-law transformed space (PSNR$_\mu$), common in HDR pipelines. PSNR, SSIM, and LPIPS are also computed on tonemapped images (subscript ``tm''). All methods use the same memory budget per compression ratio.}
  \label{tab:results_images}
  \setlength{\tabcolsep}{4pt}
  \begin{tabular}{clcccc}
    \toprule
    Ratio & Method
      & PSNR$_\mu$$\uparrow$
      & PSNR$_\text{tm}$$\uparrow$
      & SSIM$_\text{tm}$$\uparrow$
      & LPIPS$_\text{tm}$$\downarrow$ \\
    \midrule
    \multirow{3}{*}{$10\times$}
      & JPEG-XL~\cite{jpegxl}
        & \cellcolor{bestcell}\textbf{45.31}
        & \cellcolor{bestcell}\textbf{44.66}
        & \cellcolor{bestcell}\textbf{0.994}
        & \cellcolor{bestcell}\textbf{0.001} \\
      & Instant NGP~\cite{mueller2022instant}
        & \cellcolor{secondcell}37.63
        & \cellcolor{thirdcell}38.91
        & \cellcolor{thirdcell}0.949
        & \cellcolor{thirdcell}0.061 \\
      & NHT (Ours)
        & \cellcolor{thirdcell}37.16
        & \cellcolor{secondcell}39.87
        & \cellcolor{secondcell}0.963
        & \cellcolor{secondcell}0.023 \\
    \midrule
    \multirow{3}{*}{$100\times$}
      & JPEG-XL~\cite{jpegxl}
        & \cellcolor{bestcell}\textbf{38.53}
        & \cellcolor{thirdcell}35.54
        & \cellcolor{bestcell}\textbf{0.971}
        & \cellcolor{bestcell}\textbf{0.017} \\
      & Instant NGP~\cite{mueller2022instant}
        & \cellcolor{secondcell}35.17
        & \cellcolor{secondcell}36.39
        & \cellcolor{thirdcell}0.923
        & \cellcolor{thirdcell}0.078 \\
      & NHT (Ours)
        & \cellcolor{thirdcell}35.06
        & \cellcolor{bestcell}\textbf{36.43}
        & \cellcolor{secondcell}0.927
        & \cellcolor{secondcell}0.048 \\
    \bottomrule
  \end{tabular}
\end{table}

\begin{figure}
    \centering
    \includegraphics{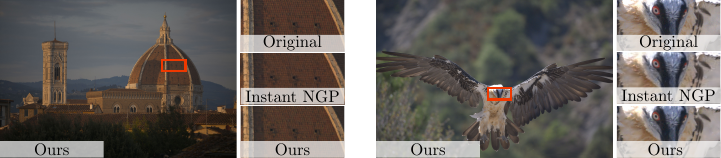}
    \caption{Comparison of our work vs Instant NGP on a 100$\times$ compression task. Original images are 45.7MP 14-bit HDR RAW files. We achieve substantially superior perceptual quality at equal compression and similar training times.}
    \label{fig:image_compression}
\end{figure}

\section{Discussion}
\label{sec_conclusion}

\paragraph{Limitations and future work}
Our method focuses on increasing per-primitive expressivity, which has inherent downsides.
In particular, our method can overfit to individual views in scenarios with very sparse supervision, leading to lower novel view synthesis performance.
Furthermore, while our method consistently achieves 140+ frames per second at inference time, the additional neural decoder results in slightly slower rendering than pure 3DGS methods(\cref{tab:comparison}).

In future work, we would like to investigate the feasibility of extracting automatic Level of Detail from a finer-grained harmonic decomposition.
Given the generality of Neural Harmonic Textures, applications beyond novel view synthesis such as radiance caching, neural physically-based rendering and geometric reconstruction are also of great interest. Finally, our work opens the possibility of removing kernel functions entirely, as we can still propagate derivatives to explicit geometry without the spatially decaying opacity function, potentially enabling drastic performance gains.

\paragraph{Conclusion}
We presented Neural Harmonic Textures, which combine particle-anchored feature vectors, harmonic activations, and a single image-space decoding pass to achieve state-of-the-art results in novel view synthesis.
Our method consistently outperforms baselines across different particle counts, trains rapidly, and has high inference performance. Moreover, it is compatible with existing deferred-shading rendering pipelines and supports a wide range of applications, from lifting semantic fields to high-resolution image reconstruction and beyond.

\paragraph{Acknowledgements}
\label{sec_ackw}
We would like to thank Ruilong Li and Martin Bisson for helpful discussions. Jorge Condor and Piotr Didyk acknowledge funding from the
Swiss National Science Foundation (SNSF, Grant 200502) and an
academic gift from Meta.

\bibliographystyle{splncs04}
\bibliography{src_bib}
\newpage
\title{Supplementary Material\\[2pt]
\large Neural Harmonic Textures for High-Quality Primitive Based Neural Reconstruction}

\titlerunning{Neural Harmonic Textures --- Supplementary}

\author{Jorge Condor\inst{1,2}\orcidlink{0000-0002-9958-0118} \and
Nicolas Mo\"enne-Loccoz\inst{1}\orcidlink{0000-0002-2312-9275}\and
Merlin Nimier-David\inst{1}\orcidlink{0000-0002-6234-3143} \and
Piotr Didyk\inst{2}\orcidlink{0000-0003-0768-8939} \and
Zan Gojcic\inst{1}\orcidlink{0000-0001-6392-2158} \and
Qi Wu\inst{1}\orcidlink{0000-0003-0342-9366}}

\authorrunning{J.~Condor et al.}

\institute{NVIDIA\\
\email{\{nicolasm,mnimierdavid,zgojcic,qiwu\}@nvidia.com}
 \and
 Universit\`a della Svizzera italiana, Lugano, Switzerland\\
\email{\{jorge.condor,piotr.didyk\}@usi.ch}}

\maketitle

\section{Additional Results}
\label{sec:additional_results}

\paragraph{Comparison with BBSplat.}
Both BBSplat~\cite{bbsplat} and NHT are primitive-texturing methods, but BBSplat uses an \emph{explicit} per-primitive RGBA grid, whereas NHT uses \emph{implicit} textures obtained by interpolating features from a virtual scaffold and decoding them with a shared MLP in a single deferred pass. \cref{tab:bbsplat} compares the two on MipNeRF360 across primitive budgets, under both equal-primitive and equal-memory settings. NHT achieves substantially higher quality and faster rendering not only at equal memory, but also at equal primitive count.

\begin{table*}[t]
\centering
\caption{Comparison with BBSplat~\cite{bbsplat} on MipNeRF360 (average over all 9 scenes). We follow BBSplat's default training strategy and vary only the primitive cap; at 15k primitives we allocate 2k to the skybox following their supplement. Memory is reported after all compression schemes are applied. For NHT, primitive geometry is stored in \texttt{fp32} and vertex features plus MLP weights in \texttt{fp16}. Performance measured on an A100 80GB (Ampere).}
\label{tab:bbsplat}
\setlength{\tabcolsep}{3pt}
\footnotesize
\resizebox{\textwidth}{!}{%
\begin{tabular}{l|cccc|cccc|cccc}
\toprule
 & \multicolumn{4}{c|}{BBSplat} & \multicolumn{4}{c|}{NHT (equal primitives)} & \multicolumn{4}{c}{NHT ($\sim$ equal memory)} \\
\cmidrule(lr){2-5} \cmidrule(lr){6-9} \cmidrule(lr){10-13}
Cap & PSNR$\uparrow$ & LPIPS$\downarrow$ & MB$\downarrow$ & FPS$\uparrow$
    & PSNR$\uparrow$ & LPIPS$\downarrow$ & MB$\downarrow$ & FPS$\uparrow$
    & PSNR$\uparrow$ & LPIPS$\downarrow$ & MB$\downarrow$ & FPS$\uparrow$ \\
\midrule
15K  & 24.35 & 0.416 &  13.2 &  86.9 & 24.54 & 0.448 &   \textbf{2.5} & 222.2 & \textbf{26.29} & \textbf{0.358} &   9.0 & \textbf{245.4} \\
30K  & 24.95 & 0.362 &  25.6 &  64.3 & 25.49 & 0.403 &   \textbf{5.0} & \textbf{214.5} & \textbf{27.03} & \textbf{0.318} &  18.1 & 192.5 \\
60K  & 26.10 & 0.294 &  50.2 &  55.0 & 26.40 & 0.358 &   \textbf{9.9} & \textbf{211.3} & \textbf{27.64} & \textbf{0.283} &  36.3 & 165.4 \\
160K & 26.85 & 0.264 & 124.6 &  33.2 & 27.39 & 0.302 &  \textbf{26.3} & \textbf{170.3} & \textbf{28.16} & \textbf{0.247} &  93.8 & 111.5 \\
300K & 27.09 & 0.255 & 215.5 &  17.3 & 27.83 & 0.273 &  \textbf{49.3} & \textbf{137.8} & \textbf{28.36} & \textbf{0.232} & 161.6 &  84.0 \\
\bottomrule
\end{tabular}%
}
\end{table*}

In \cref{tab:perscene} we provide a detailed breakdown of \cref{tab:comparison} from the main paper, with per-scene results on MipNeRF360, Tanks and Temples and Deep Blending. 
For this comparison, we isolate completely the effect of our method by ensuring an even setup for all methods. 
All four approaches are implemented under the same framework (\emph{gsplat}), trained for 30k iterations, using 1M primitives for all scenes, sharing training parameters among all scenes, and limiting to 48 features per primitive reserved for appearance. 
All results presented in this paper compute the LPIPs metric using the VGG model with normalized inputs to the range of $[-1,1]$.
\cref{tab:training_times} reports the average wall-clock training time per dataset for all four methods on an RTX 5090, under this same controlled setup.

\begin{table}[t]
  \centering
  \caption{
    Average wall-clock training time (minutes) to reach 30k iterations with 1M primitives, measured on an NVIDIA RTX 5090 and averaged per dataset.
    All methods are implemented in the same \emph{gsplat}~\cite{gsplat} framework and use the same hyperparameters as in \cref{tab:comparison}.
    NHT timings use our fused kernels.
  }
  \label{tab:training_times}
  \setlength{\tabcolsep}{6pt}
  \begin{tabular}{lccc}
    \toprule
    Method (w/ MCMC) & MipNeRF360 & Tanks \& Temples & Deep Blending \\
    \midrule
    3DGS + SH        & 6.0  & 5.1  & 5.2  \\
    3DGUT + SH       & 7.1  & 5.8  & 6.0  \\
    3DGUT + SV       & 7.7  & 6.4  & 6.6  \\
    3DGUT + NHT (Ours) & 14.5 & 9.4  & 10.6 \\
    \bottomrule
  \end{tabular}
\end{table}

Furthermore, in \cref{sup:tab:comparison_standardized} we provide another standardized comparison on all three benchmarks, only this time we match the same primitive count in each scene as the high-quality reported baselines in their respective papers, training for the same amount of iterations.
We include our method with both 48 and 64 features per primitive (48F, 64F).

\begin{table*}[t]
  \centering
  \caption{Standardized comparison on MipNeRF360~\cite{barron2022mipnerf360}, Tanks \& Temples~\cite{Knapitsch2017}, and Deep Blending~\cite{hedman2018deep}: 3DGS-MCMC, 3DGUT-MCMC, and ours with 48 and 64 features per primitive (48F, 64F). All methods use the same primitive count (matching the high-quality preset in 3DGS-MCMC, which replicates the number of primitives in each scene from original 3DGS), 30k iterations, and the same hyperparameters across all scenes. Extended version of \cref{tab:comparison} of the main paper. While our work particularly excels at extracting higher quality from less primitives, we still outperform previous works at high primitive counts across the board.}
  \label{sup:tab:comparison_standardized}
  \setlength{\tabcolsep}{3pt}
  \resizebox{\textwidth}{!}{%
  \begin{tabular}{l*{3}{c}*{3}{c}*{3}{c}}
    \toprule
    & \multicolumn{3}{c}{MipNeRF360}
    & \multicolumn{3}{c}{Tanks \& Temples}
    & \multicolumn{3}{c}{Deep Blending} \\
    \cmidrule(lr){2-4} \cmidrule(lr){5-7} \cmidrule(lr){8-10}
    Method
      & PSNR$\uparrow$ & SSIM$\uparrow$ & LPIPS$\downarrow$ 
      & PSNR$\uparrow$ & SSIM$\uparrow$ & LPIPS$\downarrow$ 
      & PSNR$\uparrow$ & SSIM$\uparrow$ & LPIPS$\downarrow$\\
    \midrule
    3DGS-MCMC + SH
      & \cellcolor{thirdcell}28.21 & \cellcolor{bestcell}\textbf{0.841} & \cellcolor{thirdcell}0.214 
      & \cellcolor{thirdcell}24.46 & \cellcolor{thirdcell}0.866 & \cellcolor{thirdcell}0.174
      & 29.49 & \cellcolor{thirdcell}0.912 & \cellcolor{thirdcell}0.306 \\
    3DGUT-MCMC + SH
      & 28.08 & \cellcolor{secondcell}0.837 & 0.218
      & 24.20 & 0.861 & 0.180
      & \cellcolor{thirdcell}29.87 & \cellcolor{secondcell}0.913 & 0.309 \\
    3DGUT-MCMC + NHT (Ours, 48F)
      & \cellcolor{secondcell}28.65 & \cellcolor{thirdcell}0.834 & \cellcolor{secondcell}0.212
      & \cellcolor{secondcell}25.41 & \cellcolor{secondcell}0.878 & \cellcolor{secondcell}0.156
      & \cellcolor{secondcell}30.67 & \cellcolor{bestcell}\textbf{0.918} & \cellcolor{secondcell}0.302 \\
    3DGUT-MCMC + NHT (Ours, 64F)
      & \cellcolor{bestcell}\textbf{28.68} & 0.832 & \cellcolor{bestcell}\textbf{0.211}
      & \cellcolor{bestcell}\textbf{25.64} & \cellcolor{bestcell}\textbf{0.879} & \cellcolor{bestcell}\textbf{0.154}
      & \cellcolor{bestcell}\textbf{30.69} & \cellcolor{bestcell}\textbf{0.918} & \cellcolor{bestcell}\textbf{0.301} \\
    \bottomrule
  \end{tabular}%
  }
\end{table*}

\begin{table*}[t]
  \centering
  \caption{Per-scene results for the standardized comparison (\cref{sup:tab:comparison_standardized}). All methods use the same primitive count (matching the baseline default), 30k iterations, and the same hyperparameters across all scenes.}
  \label{sup:tab:perscene_standardized}
  \setlength{\tabcolsep}{3pt}
  \resizebox{\textwidth}{!}{%
  \begin{tabular}{l*{3}{c}*{3}{c}*{3}{c}*{3}{c}}
    \toprule
    & \multicolumn{3}{c}{3DGS-MCMC + SH}
    & \multicolumn{3}{c}{3DGUT-MCMC + SH}
    & \multicolumn{3}{c}{NHT (Ours, 48F)}
    & \multicolumn{3}{c}{NHT (Ours, 64F)} \\
    \cmidrule(lr){2-4} \cmidrule(lr){5-7} \cmidrule(lr){8-10} \cmidrule(lr){11-13}
    Scene
      & PSNR$\uparrow$ & SSIM$\uparrow$ & LPIPS$\downarrow$
      & PSNR$\uparrow$ & SSIM$\uparrow$ & LPIPS$\downarrow$
      & PSNR$\uparrow$ & SSIM$\uparrow$ & LPIPS$\downarrow$
      & PSNR$\uparrow$ & SSIM$\uparrow$ & LPIPS$\downarrow$ \\
    \midrule
    \multicolumn{13}{l}{\textit{MipNeRF360 -- Outdoor}} \\
    \midrule
    bicycle
      & \cellcolor{bestcell}\textbf{25.87} & \cellcolor{bestcell}\textbf{0.801} & \cellcolor{bestcell}\textbf{0.183}
      & \cellcolor{secondcell}25.69 & \cellcolor{secondcell}0.790 & \cellcolor{secondcell}0.193
      & 25.53 & \cellcolor{thirdcell}0.779 & 0.202
      & \cellcolor{thirdcell}25.62 & 0.776 & 0.202 \\
    garden
      & \cellcolor{thirdcell}28.11 & \cellcolor{bestcell}\textbf{0.880} & \cellcolor{thirdcell}0.107
      & 27.84 & 0.874 & 0.111
      & \cellcolor{bestcell}\textbf{28.33} & \cellcolor{secondcell}0.878 & \cellcolor{bestcell}\textbf{0.102}
      & \cellcolor{secondcell}28.23 & \cellcolor{thirdcell}0.876 & \cellcolor{secondcell}0.103 \\
    stump
      & \cellcolor{bestcell}\textbf{27.32} & \cellcolor{bestcell}\textbf{0.808} & \cellcolor{bestcell}\textbf{0.203}
      & \cellcolor{secondcell}27.15 & \cellcolor{secondcell}0.802 & \cellcolor{secondcell}0.215
      & \cellcolor{thirdcell}27.06 & \cellcolor{thirdcell}0.788 & \cellcolor{secondcell}0.215
      & 27.02 & 0.786 & 0.215 \\
    treehill
      & \cellcolor{bestcell}\textbf{23.29} & \cellcolor{bestcell}\textbf{0.670} & \cellcolor{secondcell}0.310
      & \cellcolor{secondcell}23.13 & \cellcolor{secondcell}0.664 & 0.313
      & 21.96 & 0.639 & \cellcolor{thirdcell}0.311
      & \cellcolor{thirdcell}22.27 & \cellcolor{thirdcell}0.640 & \cellcolor{bestcell}\textbf{0.308} \\
    flowers
      & \cellcolor{bestcell}\textbf{22.25} & \cellcolor{bestcell}\textbf{0.652} & \cellcolor{bestcell}\textbf{0.312}
      & \cellcolor{bestcell}\textbf{22.25} & \cellcolor{secondcell}0.648 & 0.318
      & \cellcolor{thirdcell}21.90 & \cellcolor{thirdcell}0.637 & \cellcolor{thirdcell}0.315
      & 21.77 & 0.631 & \cellcolor{bestcell}\textbf{0.312} \\
    \midrule
    \multicolumn{13}{l}{\textit{MipNeRF360 -- Indoor}} \\
    \midrule
    room
      & \cellcolor{thirdcell}32.56 & \cellcolor{thirdcell}0.939 & \cellcolor{thirdcell}0.240
      & 32.43 & 0.939 & 0.241
      & \cellcolor{bestcell}\textbf{33.82} & \cellcolor{secondcell}0.945 & \cellcolor{secondcell}0.229
      & \cellcolor{secondcell}33.75 & \cellcolor{bestcell}\textbf{0.946} & \cellcolor{bestcell}\textbf{0.227} \\
    counter
      & 29.53 & \cellcolor{thirdcell}0.925 & \cellcolor{thirdcell}0.219
      & \cellcolor{thirdcell}29.55 & 0.925 & 0.220
      & \cellcolor{secondcell}30.80 & \cellcolor{bestcell}\textbf{0.931} & \cellcolor{secondcell}0.205
      & \cellcolor{bestcell}\textbf{30.92} & \cellcolor{bestcell}\textbf{0.931} & \cellcolor{bestcell}\textbf{0.203} \\
    kitchen
      & \cellcolor{thirdcell}32.10 & \cellcolor{thirdcell}0.938 & \cellcolor{thirdcell}0.136
      & 31.96 & 0.938 & 0.136
      & \cellcolor{secondcell}33.43 & \cellcolor{bestcell}\textbf{0.944} & \cellcolor{secondcell}0.127
      & \cellcolor{bestcell}\textbf{33.44} & \cellcolor{bestcell}\textbf{0.944} & \cellcolor{bestcell}\textbf{0.124} \\
    bonsai
      & \cellcolor{thirdcell}32.84 & \cellcolor{thirdcell}0.955 & 0.216
      & 32.76 & 0.955 & \cellcolor{thirdcell}0.215
      & \cellcolor{secondcell}34.99 & \cellcolor{bestcell}\textbf{0.963} & \cellcolor{secondcell}0.202
      & \cellcolor{bestcell}\textbf{35.07} & \cellcolor{secondcell}0.962 & \cellcolor{bestcell}\textbf{0.201} \\
    \midrule
    \multicolumn{13}{l}{\textit{Tanks \& Temples}} \\
    \midrule
    train
      & \cellcolor{thirdcell}22.48 & \cellcolor{thirdcell}0.832 & \cellcolor{thirdcell}0.221
      & 22.19 & 0.826 & 0.229
      & \cellcolor{secondcell}23.88 & \cellcolor{secondcell}0.856 & \cellcolor{secondcell}0.198
      & \cellcolor{bestcell}\textbf{24.33} & \cellcolor{bestcell}\textbf{0.858} & \cellcolor{bestcell}\textbf{0.194} \\
    truck
      & \cellcolor{thirdcell}26.43 & \cellcolor{thirdcell}0.899 & \cellcolor{thirdcell}0.126
      & 26.21 & 0.896 & 0.131
      & \cellcolor{secondcell}26.94 & \cellcolor{bestcell}\textbf{0.901} & \cellcolor{secondcell}0.115
      & \cellcolor{bestcell}\textbf{26.96} & \cellcolor{bestcell}\textbf{0.901} & \cellcolor{bestcell}\textbf{0.114} \\
    \midrule
    \multicolumn{13}{l}{\textit{Deep Blending}} \\
    \midrule
    drjohnson
      & \cellcolor{thirdcell}29.25 & \cellcolor{thirdcell}0.910 & \cellcolor{thirdcell}0.310
      & 29.13 & 0.910 & 0.315
      & \cellcolor{secondcell}30.06 & \cellcolor{bestcell}\textbf{0.916} & \cellcolor{secondcell}0.308
      & \cellcolor{bestcell}\textbf{30.09} & \cellcolor{bestcell}\textbf{0.916} & \cellcolor{bestcell}\textbf{0.306} \\
    playroom
      & 29.72 & 0.913 & \cellcolor{thirdcell}0.301
      & \cellcolor{thirdcell}30.60 & \cellcolor{thirdcell}0.916 & 0.302
      & \cellcolor{bestcell}\textbf{31.28} & \cellcolor{bestcell}\textbf{0.920} & \cellcolor{secondcell}0.297
      & \cellcolor{bestcell}\textbf{31.28} & \cellcolor{bestcell}\textbf{0.920} & \cellcolor{bestcell}\textbf{0.296} \\
    \midrule
    \midrule
    M360 Outdoor Avg
      & \cellcolor{bestcell}\textbf{25.37} & \cellcolor{bestcell}\textbf{0.762} & \cellcolor{bestcell}\textbf{0.223}
      & \cellcolor{secondcell}25.21 & \cellcolor{secondcell}0.756 & 0.230
      & 24.96 & \cellcolor{thirdcell}0.744 & \cellcolor{thirdcell}0.229
      & \cellcolor{thirdcell}24.98 & 0.742 & \cellcolor{secondcell}0.228 \\
    M360 Indoor Avg
      & \cellcolor{thirdcell}31.76 & \cellcolor{thirdcell}0.939 & \cellcolor{thirdcell}0.203
      & 31.67 & 0.939 & 0.203
      & \cellcolor{secondcell}33.26 & \cellcolor{bestcell}\textbf{0.946} & \cellcolor{secondcell}0.191
      & \cellcolor{bestcell}\textbf{33.30} & \cellcolor{bestcell}\textbf{0.946} & \cellcolor{bestcell}\textbf{0.189} \\
    M360 Total Avg
      & \cellcolor{thirdcell}28.21 & \cellcolor{bestcell}\textbf{0.841} & \cellcolor{thirdcell}0.214
      & 28.08 & \cellcolor{secondcell}0.837 & 0.218
      & \cellcolor{secondcell}28.65 & \cellcolor{thirdcell}0.834 & \cellcolor{secondcell}0.212
      & \cellcolor{bestcell}\textbf{28.68} & 0.832 & \cellcolor{bestcell}\textbf{0.211} \\
    T\&T Avg
      & \cellcolor{thirdcell}24.46 & \cellcolor{thirdcell}0.866 & \cellcolor{thirdcell}0.174
      & 24.20 & 0.861 & 0.180
      & \cellcolor{secondcell}25.41 & \cellcolor{secondcell}0.879 & \cellcolor{secondcell}0.157
      & \cellcolor{bestcell}\textbf{25.65} & \cellcolor{bestcell}\textbf{0.879} & \cellcolor{bestcell}\textbf{0.154} \\
    DB Avg
      & 29.49 & 0.912 & \cellcolor{thirdcell}0.306
      & \cellcolor{thirdcell}29.87 & \cellcolor{thirdcell}0.913 & 0.309
      & \cellcolor{secondcell}30.67 & \cellcolor{bestcell}\textbf{0.918} & \cellcolor{secondcell}0.302
      & \cellcolor{bestcell}\textbf{30.69} & \cellcolor{bestcell}\textbf{0.918} & \cellcolor{bestcell}\textbf{0.301} \\
    \bottomrule
  \end{tabular}%
  }
\end{table*}

\begin{figure}[ht]
    \centering
    \includegraphics[width=\linewidth]{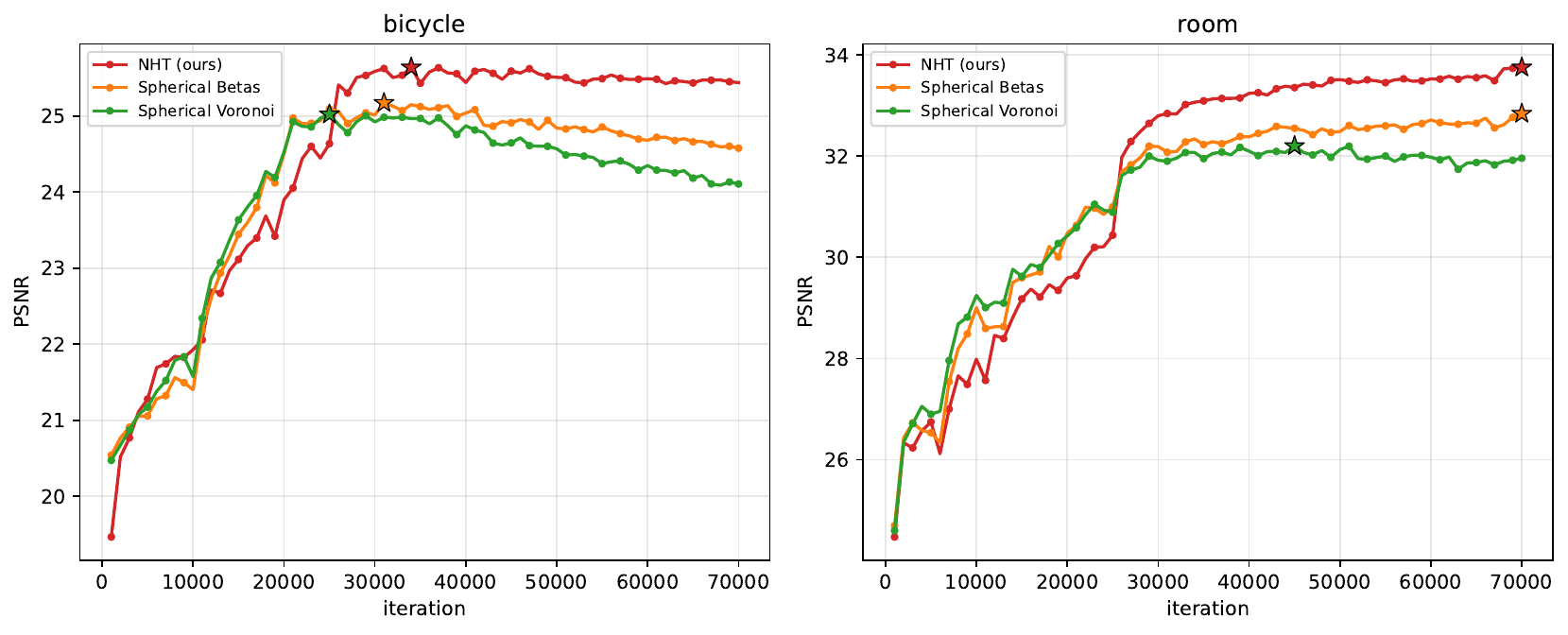}
    \caption{Eval-set PSNR during training on MipNeRF360 scenes \emph{bicycle} (outdoors, 194 images) and \emph{room} (indoors, 311 images). Overfitting can manifest when scenes are sparsely captured and trained for a higher number of epochs.}
    \label{fig:overfitting}
\end{figure}

\paragraph{Training dynamics and overfitting.}
\cref{fig:overfitting} shows eval-set PSNR during training on representative MipNeRF360 scenes. Highly expressive primitive appearance models can overfit when training for many epochs relative to dataset size, motivating the need to ensure equal training epochs rather than single frame iterations in \cref{tab:extended_comparison}.

\cref{tab:ours_perscene} details the per-scene results of main \cref{tab:extended_comparison}, which uses our split indoor/outdoor training configuration.
This is our higher quality setup, which adapts to the different scene scale and dataset sizes of outdoor and indoor datasets. 

\paragraph{Effect of reference image format.}
As noted in \cref{tab:extended_comparison}, MipNeRF360 images are distributed as pre-downscaled, re-compressed JPEG files.
The default \emph{gsplat} pipeline downscales and converts these to lossless PNGs before training and evaluation, which removes JPEG compression artifacts from the ground truth.
In \cref{tab:jpeg_vs_downscaled} we compare NHT results under both protocols.
Training and evaluating directly on the original JPEG references yields slightly lower PSNR ($-$0.28\,dB), SSIM ($-$0.011), and higher LPIPS ($+$0.014) on average, as JPEG compression artifacts in the ground truth lower the effective quality ceiling.

\begin{table}[t]
  \centering
  \caption{Effect of reference image format on MipNeRF360 evaluation for NHT (Ours). \emph{Downscaled PNGs} uses \emph{gsplat}'s default strategy, which downscales and converts the provided JPEG references to lossless PNGs before training and evaluation. \emph{JPEG (original)} trains and evaluates directly on the provided JPEG-compressed references without any preprocessing. JPEG compression artifacts in the ground truth lower the effective quality ceiling, slightly degrading all three metrics.}
  \label{tab:jpeg_vs_downscaled}
  \setlength{\tabcolsep}{5pt}
  \begin{tabular}{lccc}
    \toprule
    References & PSNR$\uparrow$ & SSIM$\uparrow$ & LPIPS$\downarrow$ \\
    \midrule
    Downscaled PNGs  & \cellcolor{bestcell}\textbf{29.022} & \cellcolor{bestcell}\textbf{0.8447} & \cellcolor{bestcell}\textbf{0.2026} \\
    JPEG (original)  & 28.738 & 0.8341 & 0.2164 \\
    \midrule
    $\Delta$         & $-$0.284 & $-$0.0106 & $+$0.0138 \\
    \bottomrule
  \end{tabular}
\end{table}

\begin{table*}[h!]
  \centering
  \caption{Per-scene results on MipNeRF360~\cite{barron2022mipnerf360}, Tanks \& Temples~\cite{Knapitsch2017}, and Deep Blending~\cite{hedman2018deep}, from \cref{tab:comparison}. This is the standardized setup limited to 1M primitives, 30k iterations, same number of appearance features per primitive (48) and same hyperparameters for all scenes, isolating entirely the effect of our approach from other variables.}
  \label{tab:perscene}
  \setlength{\tabcolsep}{3pt}
  \resizebox{\textwidth}{!}{%
  \begin{tabular}{l*{3}{c}*{3}{c}*{3}{c}*{3}{c}}
    \toprule
    & \multicolumn{3}{c}{3DGS-MCMC}
    & \multicolumn{3}{c}{3DGUT-MCMC + SH}
    & \multicolumn{3}{c}{3DGUT-MCMC + SV}
    & \multicolumn{3}{c}{3DGUT-MCMC + NHT (Ours)} \\
    \cmidrule(lr){2-4} \cmidrule(lr){5-7} \cmidrule(lr){8-10} \cmidrule(lr){11-13}
    Scene
      & PSNR$\uparrow$ & SSIM$\uparrow$ & LPIPS$\downarrow$
      & PSNR$\uparrow$ & SSIM$\uparrow$ & LPIPS$\downarrow$
      & PSNR$\uparrow$ & SSIM$\uparrow$ & LPIPS$\downarrow$
      & PSNR$\uparrow$ & SSIM$\uparrow$ & LPIPS$\downarrow$ \\
    \midrule
    \multicolumn{13}{l}{\textit{MipNeRF360 -- Outdoor}} \\
    \midrule
    bicycle
      & \cellcolor{bestcell}\textbf{25.61} & \cellcolor{bestcell}\textbf{0.774} & \cellcolor{secondcell}0.246
      & \cellcolor{secondcell}25.51 & \cellcolor{secondcell}0.771 & \cellcolor{thirdcell}0.250
      & 25.21 & 0.758 & 0.255
      & \cellcolor{thirdcell}25.43 & \cellcolor{secondcell}0.771 & \cellcolor{bestcell}\textbf{0.243} \\
    garden
      & \cellcolor{thirdcell}27.29 & \cellcolor{thirdcell}0.855 & \cellcolor{secondcell}0.155
      & 27.21 & 0.854 & \cellcolor{thirdcell}0.156
      & \cellcolor{secondcell}27.48 & \cellcolor{secondcell}0.856 & \cellcolor{thirdcell}0.156
      & \cellcolor{bestcell}\textbf{28.04} & \cellcolor{bestcell}\textbf{0.872} & \cellcolor{bestcell}\textbf{0.122} \\
    stump
      & \cellcolor{bestcell}\textbf{26.93} & \cellcolor{secondcell}0.788 & \cellcolor{secondcell}0.250
      & \cellcolor{bestcell}\textbf{26.93} & \cellcolor{bestcell}\textbf{0.789} & \cellcolor{thirdcell}0.251
      & 25.95 & 0.757 & 0.270
      & \cellcolor{thirdcell}26.58 & \cellcolor{thirdcell}0.772 & \cellcolor{bestcell}\textbf{0.249} \\
    treehill
      & \cellcolor{bestcell}\textbf{23.45} & \cellcolor{bestcell}\textbf{0.665} & \cellcolor{secondcell}0.366
      & \cellcolor{secondcell}23.30 & \cellcolor{secondcell}0.663 & \cellcolor{thirdcell}0.368
      & \cellcolor{thirdcell}23.16 & 0.653 & 0.371
      & 23.13 & \cellcolor{thirdcell}0.660 & \cellcolor{bestcell}\textbf{0.340} \\
    flowers
      & \cellcolor{secondcell}22.04 & \cellcolor{bestcell}\textbf{0.628} & \cellcolor{secondcell}0.362
      & \cellcolor{bestcell}\textbf{22.07} & \cellcolor{secondcell}0.627 & \cellcolor{thirdcell}0.363
      & \cellcolor{thirdcell}21.76 & \cellcolor{thirdcell}0.616 & 0.367
      & 21.60 & 0.615 & \cellcolor{bestcell}\textbf{0.351} \\
    \midrule
    \multicolumn{13}{l}{\textit{MipNeRF360 -- Indoor}} \\
    \midrule
    room
      & 32.33 & \cellcolor{thirdcell}0.938 & \cellcolor{thirdcell}0.245
      & \cellcolor{thirdcell}32.36 & 0.937 & 0.246
      & \cellcolor{secondcell}32.67 & \cellcolor{secondcell}0.939 & \cellcolor{secondcell}0.243
      & \cellcolor{bestcell}\textbf{33.03} & \cellcolor{bestcell}\textbf{0.943} & \cellcolor{bestcell}\textbf{0.234} \\
    counter
      & \cellcolor{thirdcell}29.48 & 0.924 & \cellcolor{thirdcell}0.223
      & 29.47 & 0.923 & 0.224
      & \cellcolor{secondcell}30.69 & \cellcolor{bestcell}\textbf{0.932} & \cellcolor{secondcell}0.213
      & \cellcolor{bestcell}\textbf{30.83} & \cellcolor{secondcell}0.931 & \cellcolor{bestcell}\textbf{0.208} \\
    kitchen
      & 31.67 & \cellcolor{thirdcell}0.935 & \cellcolor{thirdcell}0.144
      & \cellcolor{thirdcell}31.72 & \cellcolor{thirdcell}0.935 & \cellcolor{thirdcell}0.144
      & \cellcolor{secondcell}32.38 & \cellcolor{secondcell}0.938 & \cellcolor{secondcell}0.140
      & \cellcolor{bestcell}\textbf{32.98} & \cellcolor{bestcell}\textbf{0.942} & \cellcolor{bestcell}\textbf{0.132} \\
    bonsai
      & 32.69 & \cellcolor{thirdcell}0.953 & 0.220
      & \cellcolor{thirdcell}32.76 & \cellcolor{thirdcell}0.953 & \cellcolor{thirdcell}0.220
      & \cellcolor{secondcell}34.07 & \cellcolor{secondcell}0.959 & \cellcolor{secondcell}0.215
      & \cellcolor{bestcell}\textbf{34.55} & \cellcolor{bestcell}\textbf{0.961} & \cellcolor{bestcell}\textbf{0.208} \\
    \midrule
    \multicolumn{13}{l}{\textit{Tanks \& Temples}} \\
    \midrule
    train
      & \cellcolor{secondcell}22.50 & \cellcolor{secondcell}0.832 & \cellcolor{thirdcell}0.224
      & 22.10 & 0.827 & 0.231
      & \cellcolor{thirdcell}22.32 & \cellcolor{thirdcell}0.831 & \cellcolor{secondcell}0.223
      & \cellcolor{bestcell}\textbf{23.11} & \cellcolor{bestcell}\textbf{0.851} & \cellcolor{bestcell}\textbf{0.202} \\
    truck
      & \cellcolor{thirdcell}26.00 & \cellcolor{secondcell}0.891 & \cellcolor{secondcell}0.151
      & 25.88 & 0.890 & 0.153
      & \cellcolor{secondcell}26.04 & \cellcolor{secondcell}0.891 & \cellcolor{secondcell}0.151
      & \cellcolor{bestcell}\textbf{26.48} & \cellcolor{bestcell}\textbf{0.898} & \cellcolor{bestcell}\textbf{0.135} \\
    \midrule
    \multicolumn{13}{l}{\textit{Deep Blending}} \\
    \midrule
    drjohnson
      & 29.50 & \cellcolor{secondcell}0.909 & \cellcolor{secondcell}0.320
      & \cellcolor{thirdcell}29.60 & \cellcolor{secondcell}0.909 & 0.321
      & \cellcolor{secondcell}29.69 & \cellcolor{secondcell}0.909 & \cellcolor{thirdcell}0.321
      & \cellcolor{bestcell}\textbf{30.24} & \cellcolor{bestcell}\textbf{0.915} & \cellcolor{bestcell}\textbf{0.317} \\
    playroom
      & 30.45 & \cellcolor{thirdcell}0.915 & \cellcolor{secondcell}0.313
      & \cellcolor{thirdcell}30.82 & \cellcolor{secondcell}0.916 & \cellcolor{thirdcell}0.315
      & \cellcolor{secondcell}30.89 & \cellcolor{thirdcell}0.915 & 0.318
      & \cellcolor{bestcell}\textbf{31.52} & \cellcolor{bestcell}\textbf{0.920} & \cellcolor{bestcell}\textbf{0.306} \\
    \midrule
    \midrule
    M360 Outdoor Avg
      & \cellcolor{bestcell}\textbf{25.06} & \cellcolor{bestcell}\textbf{0.742} & \cellcolor{secondcell}0.276
      & \cellcolor{secondcell}25.00 & \cellcolor{secondcell}0.741 & \cellcolor{thirdcell}0.278
      & 24.71 & 0.728 & 0.284
      & \cellcolor{thirdcell}24.96 & \cellcolor{thirdcell}0.738 & \cellcolor{bestcell}\textbf{0.261} \\
    M360 Indoor Avg
      & 31.54 & \cellcolor{thirdcell}0.938 & \cellcolor{thirdcell}0.208
      & \cellcolor{thirdcell}31.58 & 0.937 & 0.209
      & \cellcolor{secondcell}32.45 & \cellcolor{secondcell}0.942 & \cellcolor{secondcell}0.203
      & \cellcolor{bestcell}\textbf{32.85} & \cellcolor{bestcell}\textbf{0.944} & \cellcolor{bestcell}\textbf{0.196} \\
    M360 Total Avg
      & \cellcolor{thirdcell}27.94 & \cellcolor{secondcell}0.829 & \cellcolor{secondcell}0.246
      & 27.93 & \cellcolor{thirdcell}0.828 & \cellcolor{thirdcell}0.247
      & \cellcolor{secondcell}28.15 & 0.823 & 0.248
      & \cellcolor{bestcell}\textbf{28.46} & \cellcolor{bestcell}\textbf{0.830} & \cellcolor{bestcell}\textbf{0.232} \\
    T\&T Avg
      & \cellcolor{secondcell}24.25 & \cellcolor{secondcell}0.861 & \cellcolor{secondcell}0.188
      & 23.99 & 0.859 & 0.192
      & \cellcolor{thirdcell}24.18 & \cellcolor{thirdcell}0.861 & \cellcolor{thirdcell}0.187
      & \cellcolor{bestcell}\textbf{24.79} & \cellcolor{bestcell}\textbf{0.875} & \cellcolor{bestcell}\textbf{0.169} \\
    DB Avg
      & 29.98 & \cellcolor{thirdcell}0.912 & \cellcolor{secondcell}0.317
      & \cellcolor{thirdcell}30.21 & \cellcolor{secondcell}0.913 & \cellcolor{thirdcell}0.318
      & \cellcolor{secondcell}30.29 & \cellcolor{thirdcell}0.912 & 0.320
      & \cellcolor{bestcell}\textbf{30.88} & \cellcolor{bestcell}\textbf{0.918} & \cellcolor{bestcell}\textbf{0.311} \\
    \bottomrule
  \end{tabular}%
  }
\end{table*}

\begin{table}[t]
  \centering
  \caption{Per-scene results of NHT (Ours) with split training configurations on MipNeRF360~\cite{barron2022mipnerf360}, Tanks \& Temples~\cite{Knapitsch2017}, and Deep Blending~\cite{hedman2018deep}. We adapt primitive count and training length per dataset to accommodate different dataset sizes and spatial scales, but keep the same learning rates and other hyperparameters. Indoor datasets benefit from encoding the central camera ray, instead of per-ray directions, due to their high parallax, acting as a regularization on ray orientation. Trained and evaluated on the original JPEG-compressed reference images.}
  \label{tab:ours_perscene}
  \setlength{\tabcolsep}{5pt}
  \begin{tabular}{lccc}
    \toprule
    Scene & PSNR$\uparrow$ & SSIM$\uparrow$ & LPIPS$\downarrow$ \\
    \midrule
    \multicolumn{4}{l}{\textit{M360 Outdoor (5M, 25k iters, per-ray)}} \\
    \midrule
    garden    & 28.06 & 0.875 & 0.108 \\
    bicycle   & 25.53 & 0.782 & 0.203 \\
    stump     & 27.10 & 0.797 & 0.206 \\
    treehill  & 23.09 & 0.652 & 0.298 \\
    flowers   & 22.08 & 0.642 & 0.304 \\
    \midrule
    \multicolumn{4}{l}{\textit{M360 Indoor (2M, 45k iters, center ray)}} \\
    \midrule
    bonsai    & 35.09 & 0.957 & 0.222 \\
    counter   & 30.83 & 0.925 & 0.220 \\
    kitchen   & 33.40 & 0.940 & 0.138 \\
    room      & 33.43 & 0.936 & 0.250 \\
    \midrule
    \multicolumn{4}{l}{\textit{Tanks \& Temples (2.5M, 40k iters, center ray)}} \\
    \midrule
    truck     & 26.91 & 0.900 & 0.112 \\
    train     & 24.45 & 0.865 & 0.169 \\
    \midrule
    \multicolumn{4}{l}{\textit{Deep Blending (2M, 30k iters, center ray)}} \\
    \midrule
    drjohnson & 30.43 & 0.918 & 0.309 \\
    playroom  & 31.45 & 0.921 & 0.296 \\
    \midrule
    \midrule
    M360 Outdoor Avg & 25.17 & 0.749 & 0.224 \\
    M360 Indoor Avg  & 33.19 & 0.940 & 0.207 \\
    \textbf{M360 Total} & \textbf{28.74} & \textbf{0.834} & \textbf{0.216} \\
    \textbf{T\&T Avg} & \textbf{25.68} & \textbf{0.882} & \textbf{0.141} \\
    \textbf{DB Avg} & \textbf{30.94} & \textbf{0.919} & \textbf{0.302} \\
    \bottomrule
  \end{tabular}
\end{table}

\subsection{Semantic Reconstruction}
\label{sup:sec:lseg}

\paragraph{Semantic Reconstruction Details}
\label{sup:sec:lseg_details}
For our semantic reconstruction task, following a similar setup to Feature 3DGS~\cite{feature3dgs}, we train a joint RGB and LSEG semantic feature field.
Joint RGB and LSEG training uses the same approach, densification, and regularization as standard radiance-field training, with the differences being in the addition of semantic supervision losses 
and an expanded decoding head in our MLP. The deferred MLP is extended to predict both RGB (3 channels) and LSEG features (512 dimensions) via addition of an extra Pytorch linear layer. Rendering speed
becomes bottlenecked by this linear layer, and more efficient fused alternatives or upsampling strategies could be explored in the future---we consider this a proof-of-concept implementation.

The total loss adds an LSEG term to the usual $\mathrm{L}_1$, D-SSIM, opacity and scale regularizers terms:
\begin{equation}
\mathcal{L}_{\text{LSEG}}
    = \lambda_{\text{LSEG}}\, \mathcal{L}_{\mathrm{L}_1} \left(\hat{f}, f^* \right) 
      + \lambda_{\text{cos}}\,\left(1 - \cos(\hat{f}, f^*)\right),
\end{equation}
where $\hat{f}$ and $f^*$ are predicted and ground-truth LSEG feature maps, and $cos$ their cosine similarity. We compute these losses at the native LSEG resolution: the model renders at RGB resolution (high); then, we bilinearly downsample 
the predicted feature map to match the (lower-resolution) LSEG targets. \cref{sup:tab:lseg_hyperparameters} lists the LSEG-specific hyperparameters used in our final training. We largely kept the same configuration from our pure RGB radiance field experiments; further tuning for this specific use case may render higher performance.

\paragraph{Differences to RGB-only training.}
Aside from the extra output head and loss terms, training proceeds as in the main paper: same optimizer, learning-rate schedules, EMA on the MLP, and color-refinement phase. The only architectural change is the increased output dimension (3 + 512 when fused, or a separate 512-dim linear layer).
 We do not use a separate training stage for LSEG; RGB and LSEG are optimized jointly from the start.

\begin{figure*}[t]
    \centering
    \setlength{\tabcolsep}{1pt}
    \renewcommand{\arraystretch}{0.3}
    \newcommand{\imgw}{0.24\textwidth}
    \begin{tabular}{cccc}
        \small GT & \small Ours & \small GT & \small Ours \\[2pt]
        \includegraphics[width=\imgw]{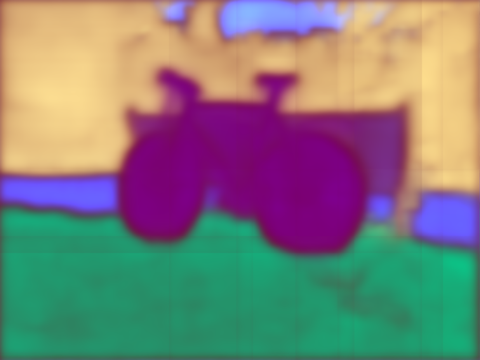} &
        \includegraphics[width=\imgw]{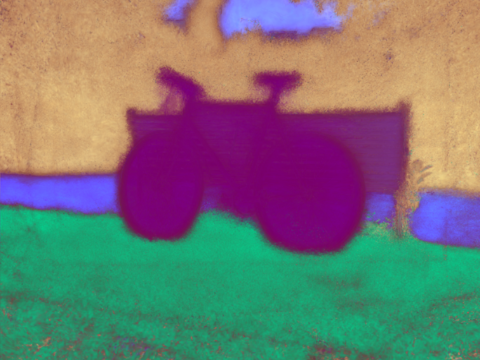} &
        \includegraphics[width=\imgw]{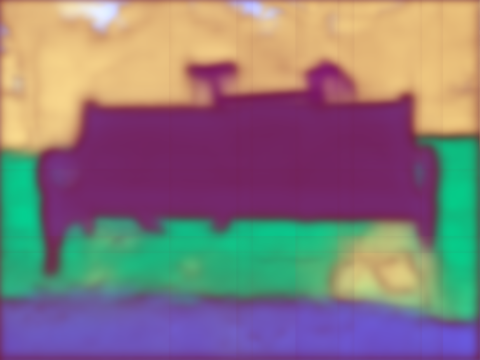} &
        \includegraphics[width=\imgw]{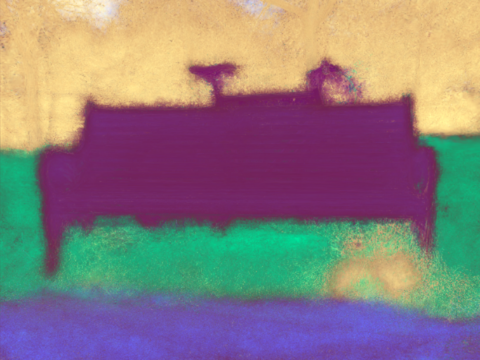} \\[-1pt]
        \includegraphics[width=\imgw]{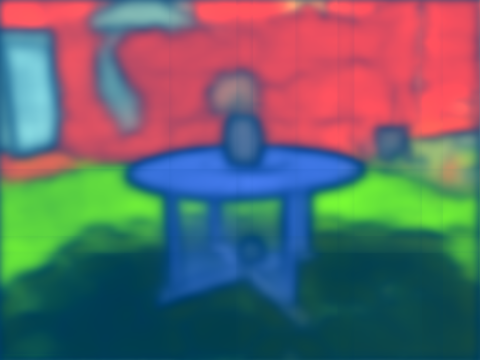} &
        \includegraphics[width=\imgw]{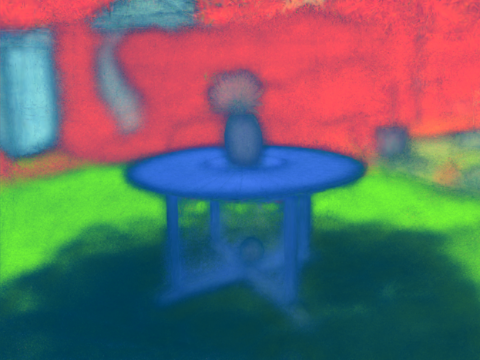} &
        \includegraphics[width=\imgw]{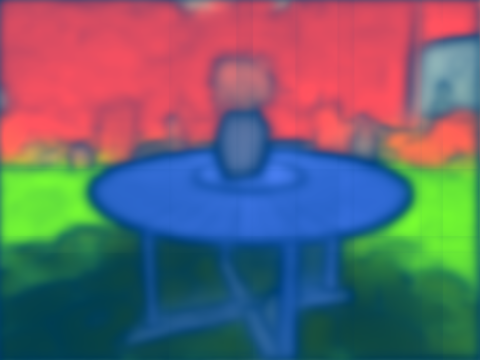} &
        \includegraphics[width=\imgw]{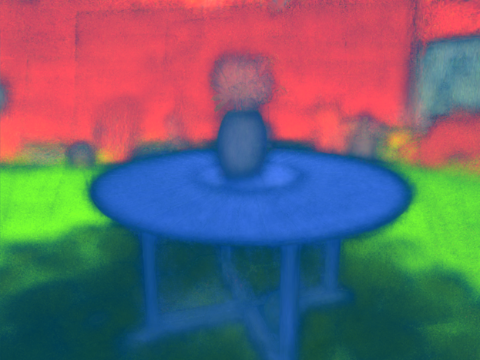} \\[-1pt]
        \includegraphics[width=\imgw]{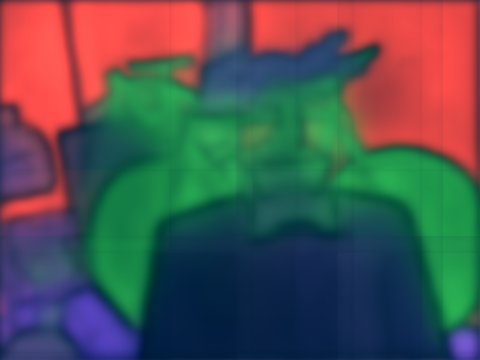} &
        \includegraphics[width=\imgw]{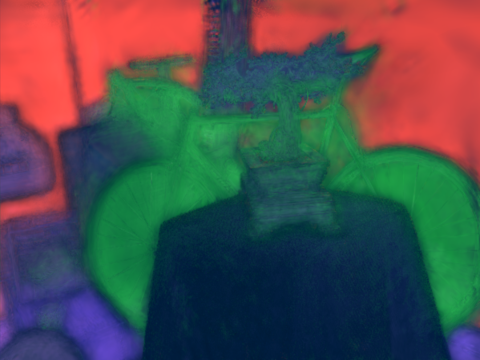} &
        \includegraphics[width=\imgw]{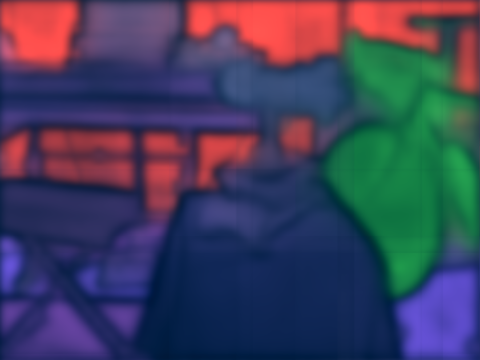} &
        \includegraphics[width=\imgw]{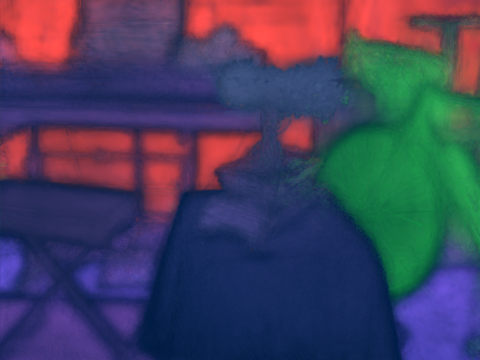} \\[-1pt]
        \includegraphics[width=\imgw]{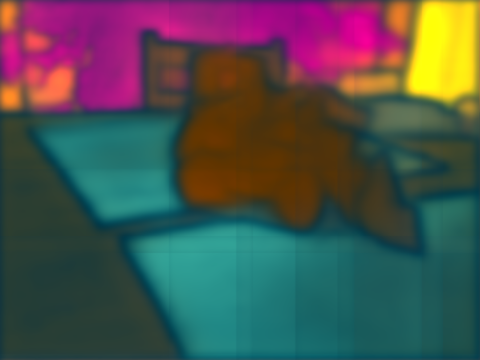} &
        \includegraphics[width=\imgw]{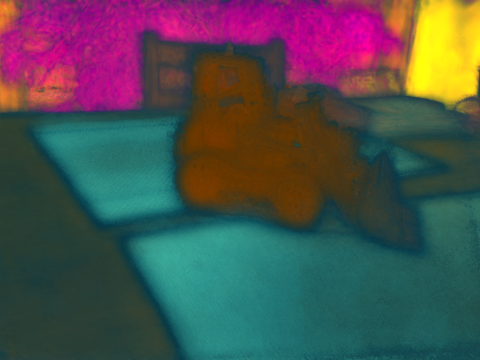} &
        \includegraphics[width=\imgw]{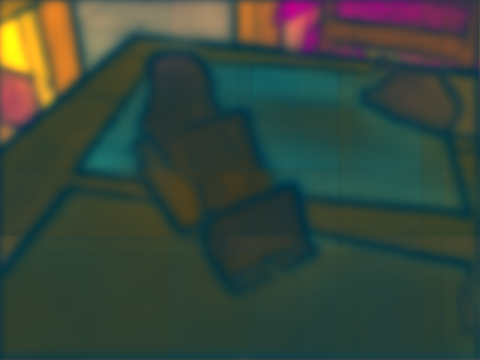} &
        \includegraphics[width=\imgw]{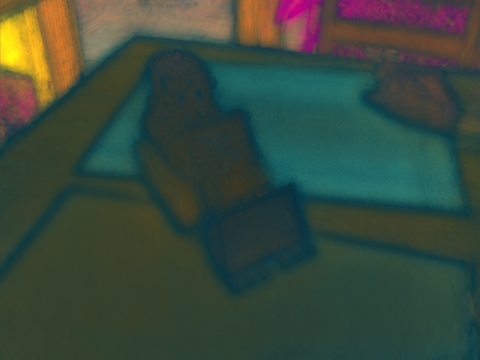} \\
    \end{tabular}
    \caption{LSEG feature PCA visualizations on test views from MipNeRF360. Rows from top to bottom: \textit{bicycle}, \textit{garden}, \textit{bonsai}, \textit{kitchen}. Each pair shows the ground-truth LSEG features (GT) alongside our rendered features (Ours). Our method faithfully reconstructs semantic feature maps while preserving sharp boundaries, at higher resolutions, and real-time.}
    \label{sup:fig:lseg_pca}
\end{figure*}

\paragraph{PCA visualization.}
LSEG feature maps are 512-dimensional and therefore not directly viewable.
For qualitative inspection, we project both ground-truth and predicted features to RGB via PCA.
The PCA basis is computed from the first image, then applied the to all frames.
\cref{sup:fig:lseg_pca} shows example PCA visualizations. 
They allow a direct visual comparison of semantic structure and show that the predicted features align well with the target, while being substantially faster to generate and higher resolution than inferring the RGB frame only and running the LSEG model to obtain its feature map. 
Semantic scene reconstruction extracts detail from the multiple training views, improving the granularity of the semantic information available for downstream applications.

Finally, in \cref{tab:ablation_lseg_featuredim} we ablate the feature dimensionality for the joint RGB and LSEG semantic reconstruction task. 

\begin{table}[t]
  \centering
  \caption{Feature dimensionality ablation for joint RGB + LSEG reconstruction on the \emph{truck} scene (Tanks \& Temples). LSEG$_\text{P}$ and LSEG$_\text{C}$ denote PSNR and cosine similarity of the 512-dim semantic features, respectively. Measured on an RTX A6000 Ada.}
  \label{tab:ablation_lseg_featuredim}
  \setlength{\tabcolsep}{4pt}
  \begin{tabular}{ccccccc}
    \toprule
    Dim & PSNR$\uparrow$ & SSIM$\uparrow$ & LPIPS$\downarrow$ & LSEG$_\text{P}$$\uparrow$ & LSEG$_\text{C}$$\uparrow$ & FPS$\uparrow$ \\
    \midrule
     4 & 23.81 & 0.853 & 0.254 & 44.33 & 0.988 & 76 \\
     8 & 25.40 & 0.876 & 0.200 & 47.10 & 0.994 & 75 \\
    16 & 26.06 & 0.885 & 0.178 & 48.23 & 0.995 & 72 \\
    32 & 26.34 & 0.887 & 0.166 & 48.68 & 0.996 & 72 \\
    64 & 26.42 & 0.887 & 0.163 & 48.79 & 0.996 & 70 \\
    \bottomrule
  \end{tabular}
\end{table}

\begin{table}[t]
  \centering
  \caption{LSEG-specific hyperparameters used for joint RGB and semantic feature training.}
  \label{sup:tab:lseg_hyperparameters}
  \setlength{\tabcolsep}{4pt}
  \begin{tabular}{ll}
    \toprule
    Parameter & Value \\
    \midrule
    LSEG output feature dimension & 512 \\
    LSEG $\mathrm{L}_1$ loss weight $\lambda_{\text{LSEG}}$ & 1.0 \\
    LSEG cosine similarity weight $\lambda_{\text{cos}}$ & 0.1 \\
    Per-primitive feature dim. (shared with RGB) & 80 \\
    \bottomrule
  \end{tabular}
\end{table}

\section{Feature interpolation}
\label{sec:feature_interpolation}

For a spatial coordinate $\mathbf{p}$ lying on the surface of our topology, or inside of its volume, we can compute the feature vector by barycentrically interpolating the feature vectors at the vertices of the shape.
For example, for a 2D triangle with vertices $\mathbf{v}_1, \mathbf{v}_2, \mathbf{v}_3$, 
the feature vector $f(\mathbf{p})$ is computed as

\begin{equation}
    f(\mathbf{p}) = \sum_{i=1}^{n} \alpha_i f_i
\end{equation}
where $f_i$ are the features at the triangle vertices, and $\alpha_i$ are the barycentric coordinates at the evaluated point. In the case of 2D triangles,
barycentric weights are computed as
\begin{equation}
    \alpha_i = \frac{A_i}{A}, \quad A = \sum_{j=1}^{n} A_j
\end{equation}
where $A$ is the total area of the triangle and $A_i$ is the area of the sub-triangle formed by the evaluation point $\mathbf{p}$ and the edge opposite to vertex $i$. For a triangle with vertices $\mathbf{v}_1, \mathbf{v}_2, \mathbf{v}_3$:
\begin{align}
    A_1 &= \frac{1}{2} \left\| (\mathbf{v}_2 - \mathbf{p}) \times (\mathbf{v}_3 - \mathbf{p}) \right\|, \nonumber \\
    A_2 &= \frac{1}{2} \left\| (\mathbf{v}_3 - \mathbf{p}) \times (\mathbf{v}_1 - \mathbf{p}) \right\|, \nonumber \\
    A_3 &= \frac{1}{2} \left\| (\mathbf{v}_1 - \mathbf{p}) \times (\mathbf{v}_2 - \mathbf{p}) \right\|.
\end{align}

For 2D connected triangles, this is similar to vertex features recently introduced in the context of neural global illumination~\cite{ruisu2025}. Our formulation however is more general and
extends to higher order topologies and disconnected meshes. For tetrahedra, barycentric weights are similarly computed as

\begin{equation}
    \alpha_i = \frac{V_i}{V}, \quad V = \sum_{j=1}^{n} V_j
\end{equation}
where $V$ is the total volume of the tetrahedron and $V_i$ is the volume of the sub-tetrahedron formed by the evaluation point $\mathbf{p}$ and the face opposite to vertex $i$. For a tetrahedron with vertices $\mathbf{v}_1, \mathbf{v}_2, \mathbf{v}_3, \mathbf{v}_4$:
\begin{align}
    V_1 &= \frac{1}{6} \left\| (\mathbf{v}_2 - \mathbf{p}) \times (\mathbf{v}_3 - \mathbf{p}) \cdot (\mathbf{v}_4 - \mathbf{p}) \right\|, \nonumber \\
    V_2 &= \frac{1}{6} \left\| (\mathbf{v}_3 - \mathbf{p}) \times (\mathbf{v}_4 - \mathbf{p}) \cdot (\mathbf{v}_1 - \mathbf{p}) \right\|, \nonumber \\
    V_3 &= \frac{1}{6} \left\| (\mathbf{v}_4 - \mathbf{p}) \times (\mathbf{v}_1 - \mathbf{p}) \cdot (\mathbf{v}_2 - \mathbf{p}) \right\|, \nonumber \\
    V_4 &= \frac{1}{6} \left\| (\mathbf{v}_1 - \mathbf{p}) \times (\mathbf{v}_2 - \mathbf{p}) \cdot (\mathbf{v}_3 - \mathbf{p}) \right\|.
\end{align}

\subsection{Optimization choices and hyperparameters}
\label{sup:sec:optimization_details}

\cref{tab:ablation_opt} ablates the individual contributions of the training strategies described in the main paper 
(learning rate scheduling, EMA on the MLP weights, loss scaling, color-refinement phase, and opacity and scale regularization).
\cref{sup:tab:hyperparameters} lists the full set of hyperparameters and optimization details used for radiance field reconstruction.

\cref{tab:ablation_encoding} ablates the encoding function applied to the interpolated features before the decoding MLP. We compare no encoding (identity), a ReLU activation, a cosine encoding, and a paired cosine-and-sine encoding.

\begin{table}[t]
  \centering
  \caption{
    Ablation study on training strategies.
    The rows are \emph{not} additive, i.e.\ we only test one strategy at a time.
    All experiments on the MipNeRF360 dataset, using 64 features per primitive and a 128$\times$3 MLP.
    All MCMC baselines in \cref{tab:comparison} already use the same opacity regularizer; row E) ablates its removal from NHT.
    Note that the scale regularization (F) does not significantly affect reconstruction quality, but does improve render time.
}
  \label{tab:ablation_opt}
  \setlength{\tabcolsep}{5pt}
  \begin{tabular}{clccc}
    \toprule
    & Configuration & PSNR$\uparrow$ & SSIM$\uparrow$ & LPIPS (Alex)$\downarrow$ \\
    \midrule
    & Our Model & 28.49 & 0.828 & 0.153 \\
    \midrule
    A) & No LR Scheduling           & 28.37 & 0.829 & 0.147 \\
    B) & No EMA on MLP Weights       & 28.46 & 0.828 & 0.152 \\
    D) & No Color Refinement Phase   & 28.35 & 0.827 & 0.154 \\
    E) & No Opacity Regularization   & 27.85 & 0.806 & 0.181 \\
    F) & No Scale Regularization     & 28.50 & 0.828 & 0.153 \\
    G) & No Direction Encoding       & 28.11 & 0.816 & 0.155 \\
    I) & No Direction Scaling        & 28.35 & 0.822 & 0.158 \\
    \bottomrule
  \end{tabular}
\end{table}

\begin{table}[t]
  \centering
  \caption{Ablating the choice of feature encoding function (MipNeRF360 dataset).}
  \label{tab:ablation_encoding}
  \setlength{\tabcolsep}{5pt}
  \begin{tabular}{lcccc}
    \toprule
    Encoding & PSNR$\uparrow$ & SSIM$\uparrow$ & LPIPS$\downarrow$\\
    \midrule
    None (identity)       & 28.21 & 0.819 & 0.255 \\
    ReLU                  & 28.19 & 0.819 & 0.255 \\
    Cosine                & 28.34 & 0.826 & 0.234 \\
    Cosine \& Sine         & \cellcolor{bestcell}\textbf{28.46} & \cellcolor{bestcell}\textbf{0.830} & \cellcolor{bestcell}\textbf{0.232} \\
    \bottomrule
  \end{tabular}
\end{table}

\begin{table}[t]
  \centering
  \caption{Hyperparameters and optimization details for radiance field reconstruction (MipNeRF360, Tanks \& Temples, Deep Blending). Values follow our \emph{gsplat} implementation~\cite{ye2025gsplat}.}
  \label{sup:tab:hyperparameters}
  \setlength{\tabcolsep}{3pt}
  \small
  \begin{tabular}{ll}
    \toprule
    \textbf{Training} & \\
    \midrule
    Color refinement phase & Last 3\,000 steps (geometry frozen, no reg.) \\
    \midrule
    \textbf{Loss \& regularization} & \\
    \midrule
    SSIM weight $\lambda$ (D-SSIM vs $\mathrm{L}_1$) & 0.1 \\
    Opacity regularization $\lambda_{\alpha}$ & 0.02 \\
    Scale regularization $\lambda_s$ & 0.005 \\
    \midrule
    \textbf{Learning rates (initial)} & \\
    \midrule
    Positions (means) & $1.6\times10^{-4}$ \\
    Scales & $5\times10^{-3}$ \\
    Opacities & $5\times10^{-2}$ \\
    Quaternions & $1\times10^{-3}$ \\
    Deferred features & $1.5\times10^{-2}$ \\
    Deferred MLP & $6.8\times10^{-4}$ \\
    \midrule
    \textbf{LR schedule} & \\
    \midrule
    Positions & Exponential decay (final factor 0.01) \\
    Features \& MLP & Cosine annealing (final factor 0.1) \\
    \midrule
    \textbf{EMA (deferred MLP)} & \\
    \midrule
    Decay $\gamma$ & 0.95 \\
    Start step & 0 \\
    \midrule
    \textbf{Appearance (NHT)} & \\
    \midrule
    Feature dim.\ per primitive & 16-64 \\
    MLP hidden dim.\ $\times$ layers & $64-128\times2-3$ \\
    View encoding & Spherical harmonics (2nd degree) \\
    \bottomrule
  \end{tabular}
\end{table}

\begin{table*}[t]
  \centering
  \begin{minipage}[t]{0.48\textwidth}
    \centering
    \caption{Ablating feature count $N$. Measured on an RTXA6000 Ada.}
    \label{tab:ablation_featuredim}
    \vspace{0.5em}
    \setlength{\tabcolsep}{4pt}
    \begin{tabular}{ccccc}
      \toprule
      Dim & PSNR$\uparrow$ & SSIM$\uparrow$ & LPIPS$\downarrow$ & FPS$\uparrow$ \\
      \midrule
      4 & 27.18 & 0.809 & 0.288 & 165 \\
      8 & 27.83 & 0.818 & 0.261 & 156 \\
      16 & 28.22 & 0.826 & 0.246 & 142 \\
      24 & 28.31 & 0.828 & 0.240 & 134 \\
      32 & 28.39 & 0.829 & 0.237 & 133 \\
      48 & 28.43 & 0.829 & 0.231 & 116 \\
      64 & 28.48 & 0.828 & 0.230 & 105 \\
      80 & 28.47 & 0.828 & 0.227 & 98 \\
      \bottomrule
    \end{tabular}
  \end{minipage}
  \hfill
  \begin{minipage}[t]{0.48\textwidth}
    \centering
    \caption{Ablating MLP architecture (layer width $\times$ hidden layer count). Measured on an RTXA6000 Ada.}
    \label{tab:ablation_modelsize}
    \vspace{0.5em}
    \setlength{\tabcolsep}{4pt}
    \begin{tabular}{ccccc}
      \toprule
      Size & PSNR$\uparrow$ & SSIM$\uparrow$ & LPIPS$\downarrow$ & FPS$\uparrow$ \\
      \midrule
      $16\!\times\!2$  & 28.16 & 0.824 & 0.240 & 97 \\
      $16\!\times\!4$  & 28.03 & 0.820 & 0.244 & 98 \\
      $32\!\times\!2$  & 28.25 & 0.825 & 0.238 & 97 \\
      $32\!\times\!4$  & 28.23 & 0.824 & 0.237 & 101 \\
      $64\!\times\!2$  & 28.38 & 0.827 & 0.234 & 102 \\
      $64\!\times\!4$  & 28.36 & 0.826 & 0.233 & 102 \\
      $128\!\times\!2$ & 28.47 & 0.828 & 0.230 & 106 \\
      $128\!\times\!4$ & 28.50 & 0.828 & 0.229 & 103 \\
      \bottomrule
    \end{tabular}
  \end{minipage}
\end{table*}

\section{2D Image Reconstruction Details}
\label{sec:2d_details}

We provide additional details on the high-resolution HDR image fitting experiment summarized in the main paper.
We adapt Neural Harmonic Textures to this domain while preserving the core idea: learnable per-vertex features, interpolated per pixel and activated with periodic functions,
and decoded by a shallow MLP. 
However, there are a number of differences to the 3D and semantic reconstruction tasks, which we detail below.

\subsection{Representation}
\label{sup:sec:2d_representation}

\paragraph{Triangle mesh.}
Unlike the volumetric disconnected primitives used in the scene reconstruction tasks, the 2D image fitting experiment employs a \emph{connected Delaunay triangulation} that tiles the image plane without overlap.
Every pixel belongs to exactly one triangle; there is no alpha blending, no Gaussian kernel, and no depth ordering. 
Vertices can be freely learned and positioned, and carry individual learnable feature vectors.
Given a query pixel, its containing triangle is found via a tile-accelerated rasterizer, and the features at the triangle's three vertices are interpolated to produce a per-pixel feature vector.

\paragraph{Clough--Tocher $C^1$ interpolation.}
Standard barycentric (linear) interpolation produces a $C^0$ feature field: while continuous, the feature gradient is discontinuous across triangle edges. This
turns out to be problematic when meshes are fully opaque, as it creates high frequency artifacts.
Despite being followed by a frequency encoding and a nonlinear MLP, these gradient discontinuities manifest as visible seam artifacts along every edge.

To eliminate this, we replace linear interpolation with \emph{Clough--Tocher cubic interpolation}, 
which constructs a degree-3 B\'ezier patch on each triangle from vertex values and vertex gradients, guaranteeing $C^1$ 
continuity everywhere (and $C^\infty$ within each triangle).
The 10 B\'ezier control points are:
\begin{align}
  c_{300} &= f_0, \quad c_{030} = f_1, \quad c_{003} = f_2, \nonumber \\
  c_{210} &= f_0 + \tfrac{1}{3}\,\nabla f_0 \cdot \mathbf{e}_{01}, \quad
  c_{120} = f_1 - \tfrac{1}{3}\,\nabla f_1 \cdot \mathbf{e}_{01}, \nonumber \\
  c_{021} &= f_1 + \tfrac{1}{3}\,\nabla f_1 \cdot \mathbf{e}_{12}, \quad
  c_{012} = f_2 - \tfrac{1}{3}\,\nabla f_2 \cdot \mathbf{e}_{12}, \nonumber \\
  c_{102} &= f_2 + \tfrac{1}{3}\,\nabla f_2 \cdot \mathbf{e}_{20}, \quad
  c_{201} = f_0 - \tfrac{1}{3}\,\nabla f_0 \cdot \mathbf{e}_{20}, \nonumber \\
  c_{111} &= \tfrac{1}{6}\bigl(c_{210}{+}c_{120}{+}c_{021}{+}c_{012}{+}c_{102}{+}c_{201}\bigr) \nonumber \\
          &\quad - \tfrac{1}{6}\bigl(c_{300}{+}c_{030}{+}c_{003}\bigr),
  \label{eq:ct_control_points}
\end{align}
where $f_i$ and $\nabla f_i$ are the feature value and gradient at vertex~$i$, and $\mathbf{e}_{ij} = \mathbf{v}_j - \mathbf{v}_i$.

\paragraph{Vertex gradient estimation.}
The Clough--Tocher scheme requires per-vertex gradients. We estimate these via a per-vertex least-squares fit over the one-ring neighborhood: for each vertex~$v$, we gather all neighbor vertices $\{v_j\}_{j \in \mathcal{N}(v)}$ and solve
\begin{equation}
  \nabla f_v = \mathbf{A}_v^{-1}\,\mathbf{b}_v, \quad
  \mathbf{A}_v = \sum_{j} \Delta\mathbf{p}_j\,\Delta\mathbf{p}_j^\top, \quad
  \mathbf{b}_v = \sum_{j} \Delta f_j\,\Delta\mathbf{p}_j,
\end{equation}
where $\Delta\mathbf{p}_j = \mathbf{p}_j - \mathbf{p}_v$ and $\Delta f_j = f_j - f_v$.
The 2$\times$2 system is solved analytically per vertex in a CUDA kernel.
This operation is differentiable and computed at every forward pass, enabling gradients to flow through the vertex topology. 

\paragraph{Feature encoding and MLP.}
Feature encoding and MLP decoding are the same as in the main scene reconstruction task. 
\subsection{Training}
\label{sup:sec:2d_training}

\paragraph{Pixel sampling.}
At each iteration, we sample a batch of pixels via stratified sampling.
Ground-truth values at sub-pixel coordinates are obtained via bilinear interpolation on the target image, implemented as a CUDA texture lookup for efficiency.

\paragraph{Optimizers and schedules.}
We use Adam~\cite{kingma2017adam} with separate parameter groups for positions ($\text{lr} = 10^{-4}$), features ($\text{lr} = 5 \times 10^{-3}$), and MLP weights ($\text{lr} = 5 \times 10^{-5}$).
An exponential decay schedule is applied after 20\,000 iterations with a factor of $0.33$ every 10\,000 steps, following Instant NGP~\cite{mueller2022instant}.

\paragraph{Loss function.}
We train with a pixel-wise MSE loss in $\mu$-law encoded space. We found this to produce better results than using the relative L2 loss suggested in the Instant NGP paper.
Raw 14-bit sensor values $x$ are transformed as $f(x) = \log(1 + \mu\, x/w) / \log(1 + \mu)$ with $\mu = 5000$ and $w = 2^{14} - 1 = 16{,}383$ (the white level of 14-bit data).
This compresses the dynamic range before supervision, placing relatively more emphasis on shadow detail, which is standard in HDR imaging pipelines.

\subsection{Coarse-to-Fine Densification}
\label{sup:sec:2d_densification}

Training begins from a sparse mesh initialized via \emph{edge-aware sampling}: we compute Sobel gradient magnitudes on a downsampled version of the target image and sample initial vertex positions proportionally to edge strength, with a small uniform floor for coverage.
The result is a Delaunay triangulation of the sampled points, with boundary vertices ensuring full image coverage.

The mesh is progressively refined through densification.
Every 500 iterations (starting at iteration 1\,500 and ending at 15\,000), we score each triangle by
\begin{equation}
  \text{score}_t = \overline{\text{DSSIM}}_t \cdot |\text{pixels}_t|^{\alpha},
\end{equation}
where $\overline{\text{DSSIM}}_t$ is the mean DSSIM (per-pixel structural dissimilarity) of triangle~$t$ and $\alpha = 0.75$ is the area-weighting exponent.
The area weighting prevents repeated splitting of tiny, high-error triangles and distributes vertices more evenly in proportion to their spatial impact.
Triangles with fewer than 3 pixels are excluded from scoring.

At each densification step, we insert centroid vertices for the top-scoring triangles (up to $0.35 \times V$ new vertices, where $V$ is the current count), initialize their features by averaging the parent triangle's vertex features, and re-run Delaunay triangulation on the full vertex set.
We also detect when the mesh becomes degenerate (e.g. due to vertices becoming colinear) and re-mesh, although this rarely occurs in practice.

\subsection{CUDA Rasterizer}
\label{sup:sec:2d_rasterizer}

All rasterization and interpolation operations are implemented as custom forward and backward CUDA kernels, supporting both full-image rendering and batched pixel-query modes. Training generally takes between 1-3 minutes in an RTXA6000 Ada.

\subsection{Post-Training Compression}
\label{sup:sec:2d_compression}
We also briefly explore the use of post-training compression techniques to further reduce storage requirements.

We compress vertex positions to \texttt{UInt16}, vertex features to uniform \texttt{Int8} (with a per-channel scale and offset), and MLP weights to \texttt{FP16}. 
An entropy coder (Zstandard, level~19) is applied to the serialized payload.
Combining these techniques, the resulting images can be compressed by an additional factor of approximately $3\times$ beyond the training result, with negligible quality loss.
\cref{tab:2d_compressed} summarizes the results of the post-training compression techniques.

\begin{table}[t]
\centering
\caption{Impact of post-training compression on NHT at 100$\times$.
Averages over 15 images.
\textit{Baseline}: uncompressed fp32 features/parameters + fp16 MLP weights.
\textit{int8 features}: uniform 8-bit quantization with per-channel scale/offset.
\textit{int16 positions}: fixed-point uint16 vertex positions.
\textit{int8+int16+fp16}: combined quantization (features int8, positions int16, MLP fp16).
\textit{+ Zstandard/Brotli}: entropy coding on the serialized payload.}
\label{tab:2d_compressed}
\resizebox{\textwidth}{!}{%
\begin{tabular}{lrrrccccc}
\toprule
Scheme & Bytes & BPP & Ratio & PSNR$\uparrow$ ($\mu$) & PSNR$\uparrow$ (lin) & PSNR$\uparrow$ (tm) & SSIM$\uparrow$ ($\mu$) & SSIM$\uparrow$ (tm) \\
\midrule
Baseline         & 4,809,236 & 0.8418 &  100$\times$ & 33.89 & 38.45 & 34.78 & 0.8830 & 0.9163 \\
int8 features    & 2,418,189 & 0.4233 &  198.4$\times$ & 33.85 & 38.33 & 34.71 & 0.8827 & 0.9160 \\
int16 positions  & 4,011,138 & 0.7021 &  119.6$\times$ & 33.89 & 38.45 & 34.78 & 0.8830 & 0.9163 \\
int8+int16+fp16mlp  & 1,607,793 & 0.2814 & 298.4$\times$ & 33.85 & 38.32 & 34.70 & 0.8827 & 0.9160 \\
~~+ Zstandard    & 1,472,273 & 0.2577 & 326.0$\times$ & 33.85 & 38.32 & 34.70 & 0.8827 & 0.9160 \\
~~+ Brotli       & 1,450,542 & 0.2539 & 331.0$\times$ & 33.85 & 38.32 & 34.70 & 0.8827 & 0.9160 \\
\bottomrule
\end{tabular}}
\end{table}

\subsection{Per-Image Results}
\label{sup:sec:2d_perimage}

\begin{figure*}[t]
  \centering
  \setlength{\tabcolsep}{1pt}
  \renewcommand{\arraystretch}{0.3}
  \begin{tabular}{ccc}
  \small Instant NGP & \small NHT (Ours) & \small Reference \\[2pt]
  \includegraphics[width=0.32\textwidth]{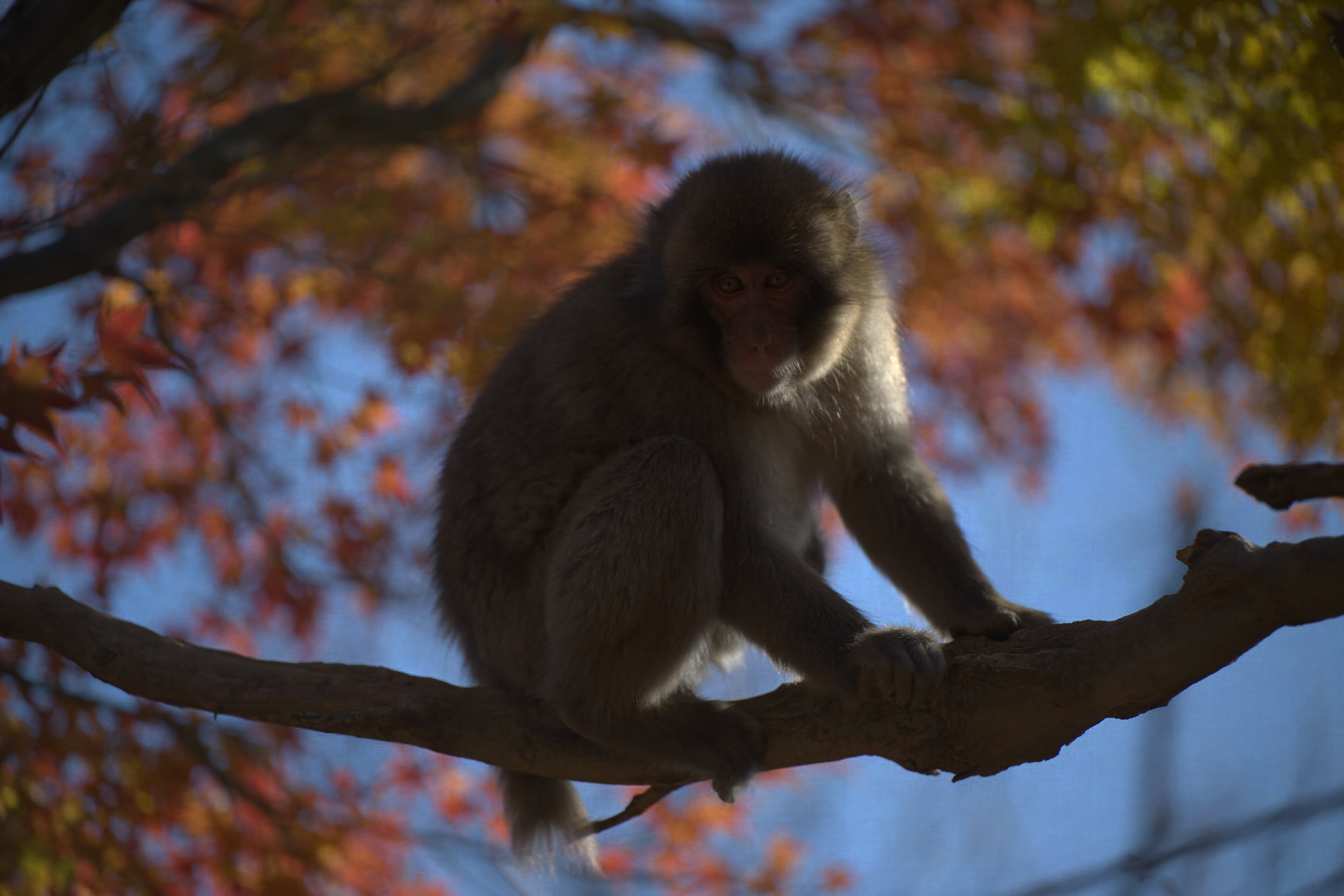} &
  \includegraphics[width=0.32\textwidth]{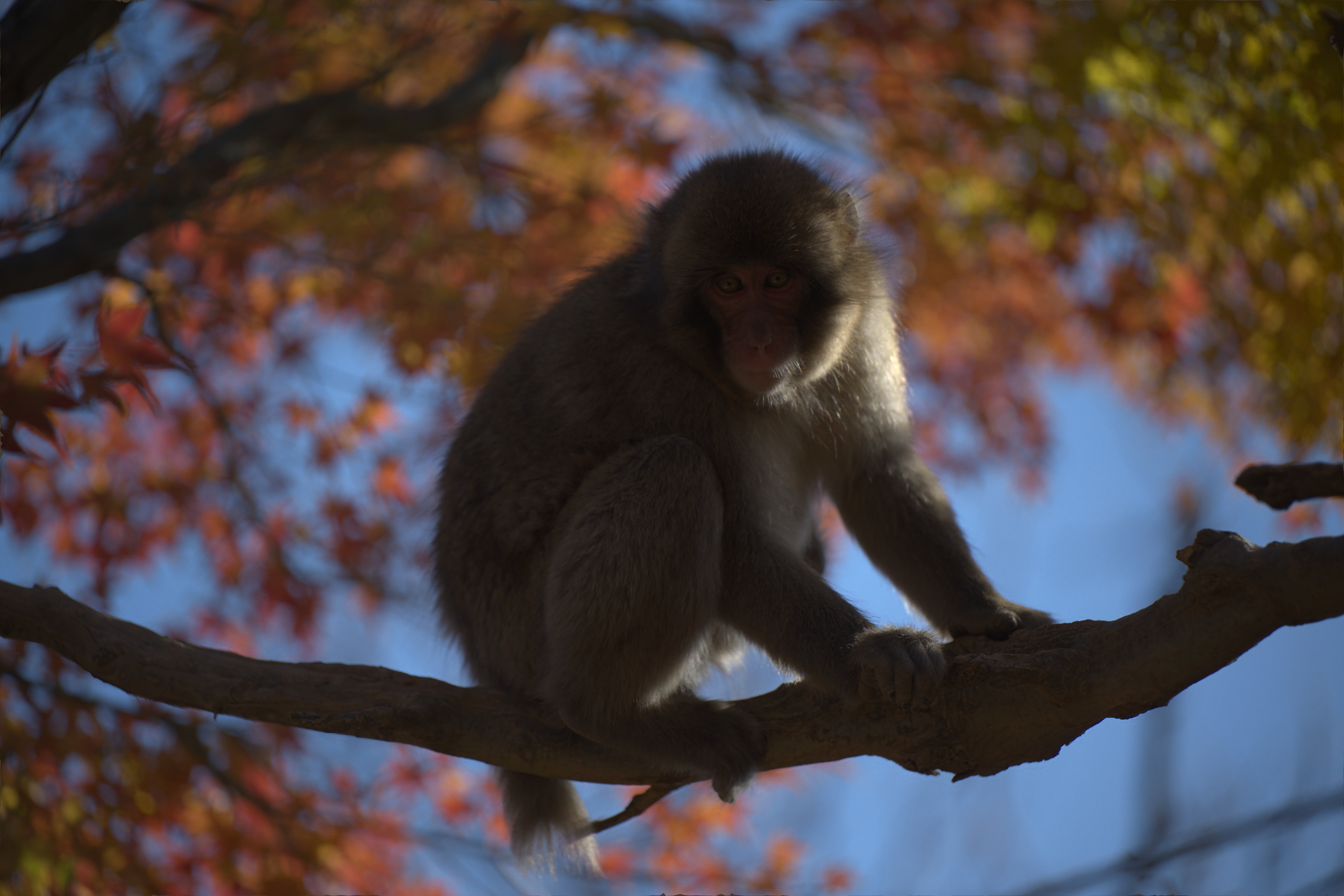} &
  \includegraphics[width=0.32\textwidth]{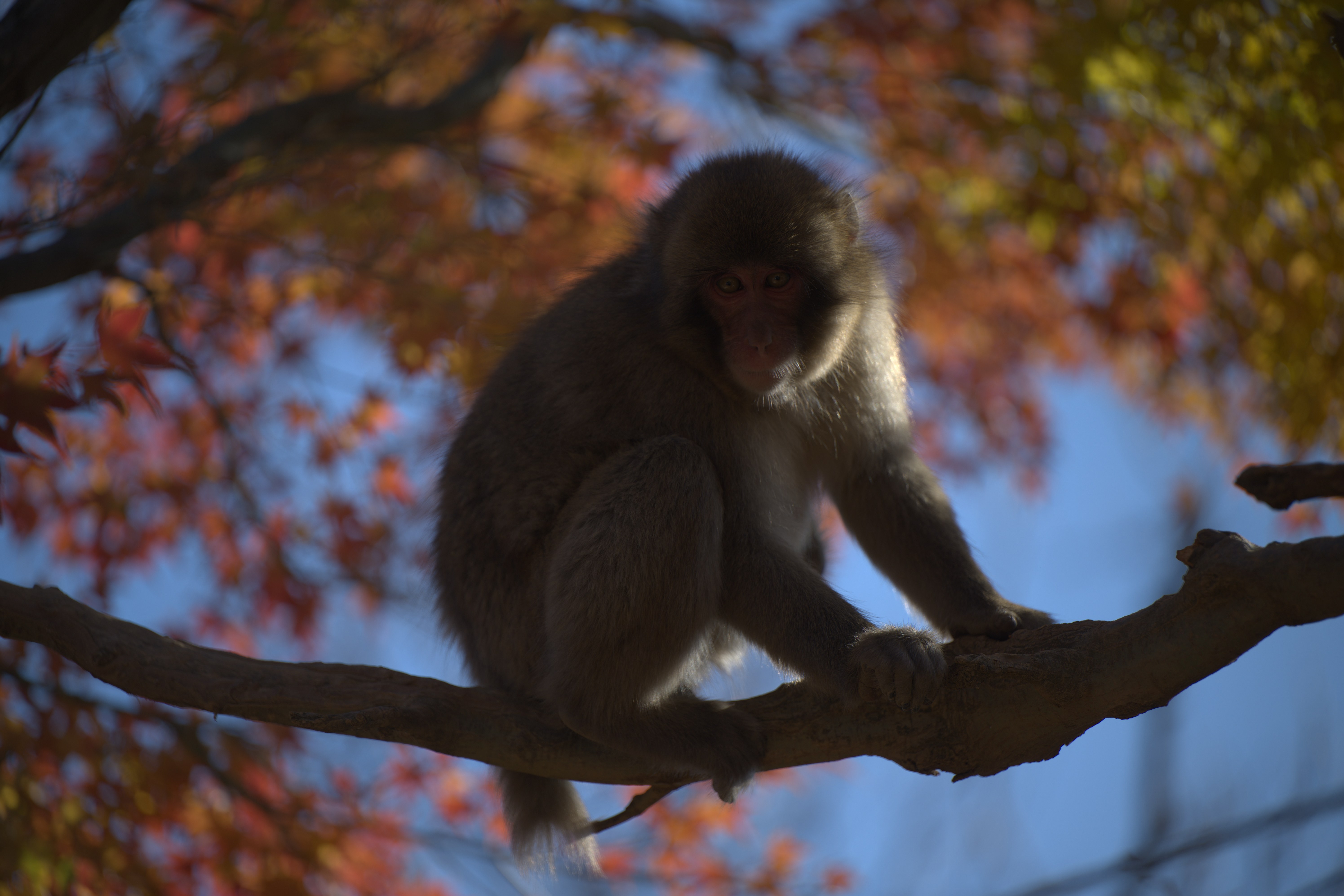} \\
  \tiny 42.5\,dB / 0.019 & \tiny 42.1\,dB / 0.010 & \tiny DSC\_3351 \\[3pt]
  \includegraphics[width=0.32\textwidth]{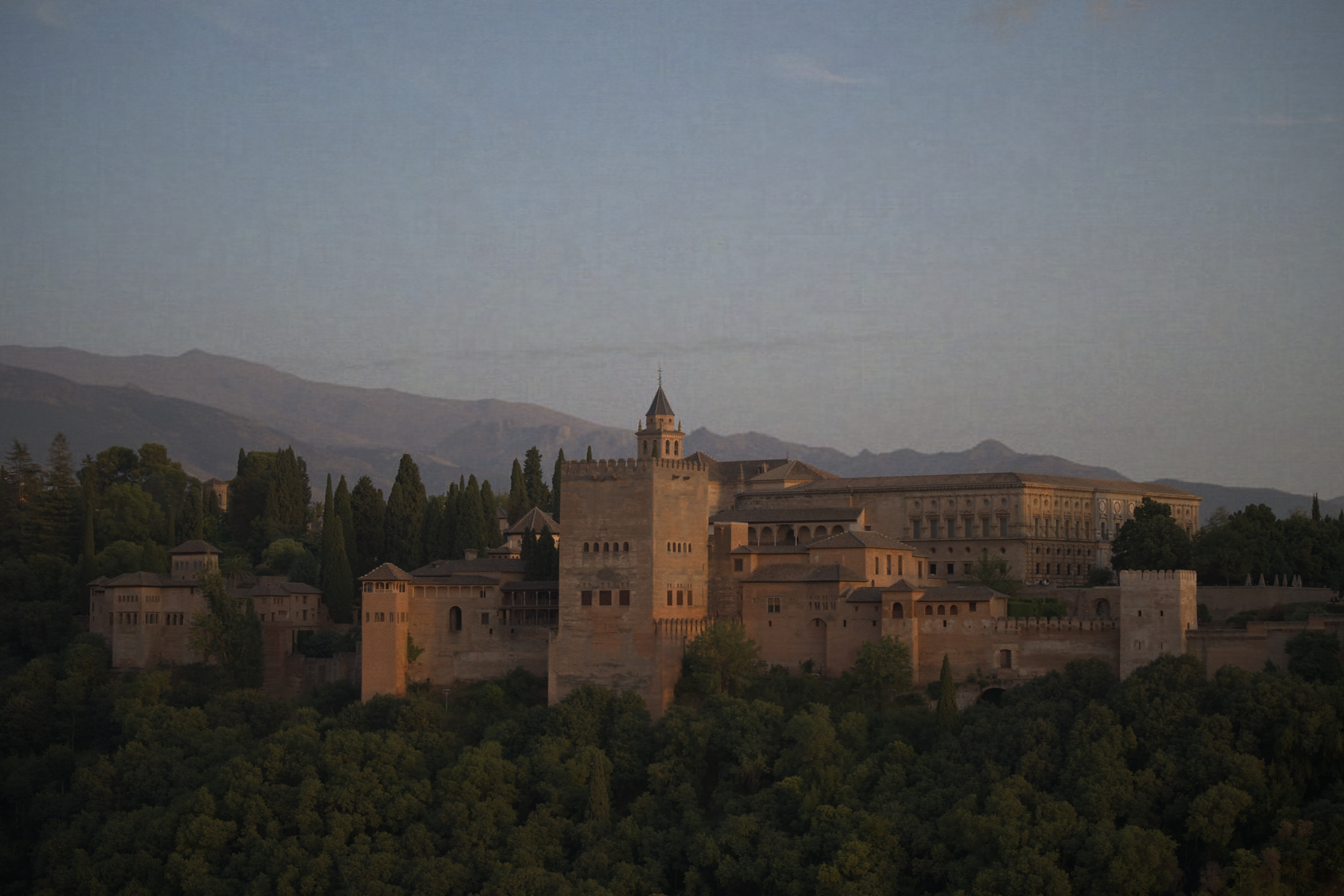} &
  \includegraphics[width=0.32\textwidth]{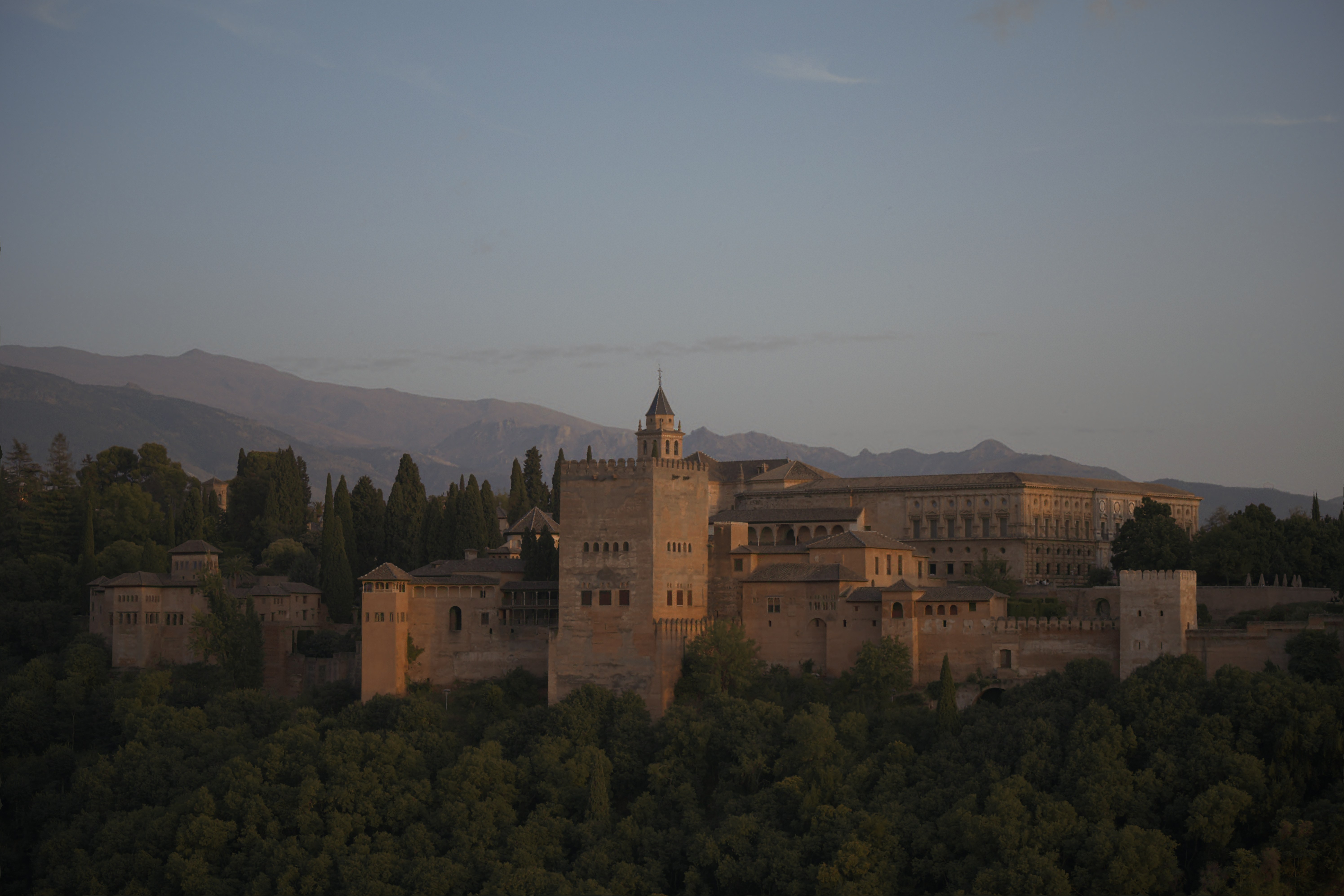} &
  \includegraphics[width=0.32\textwidth]{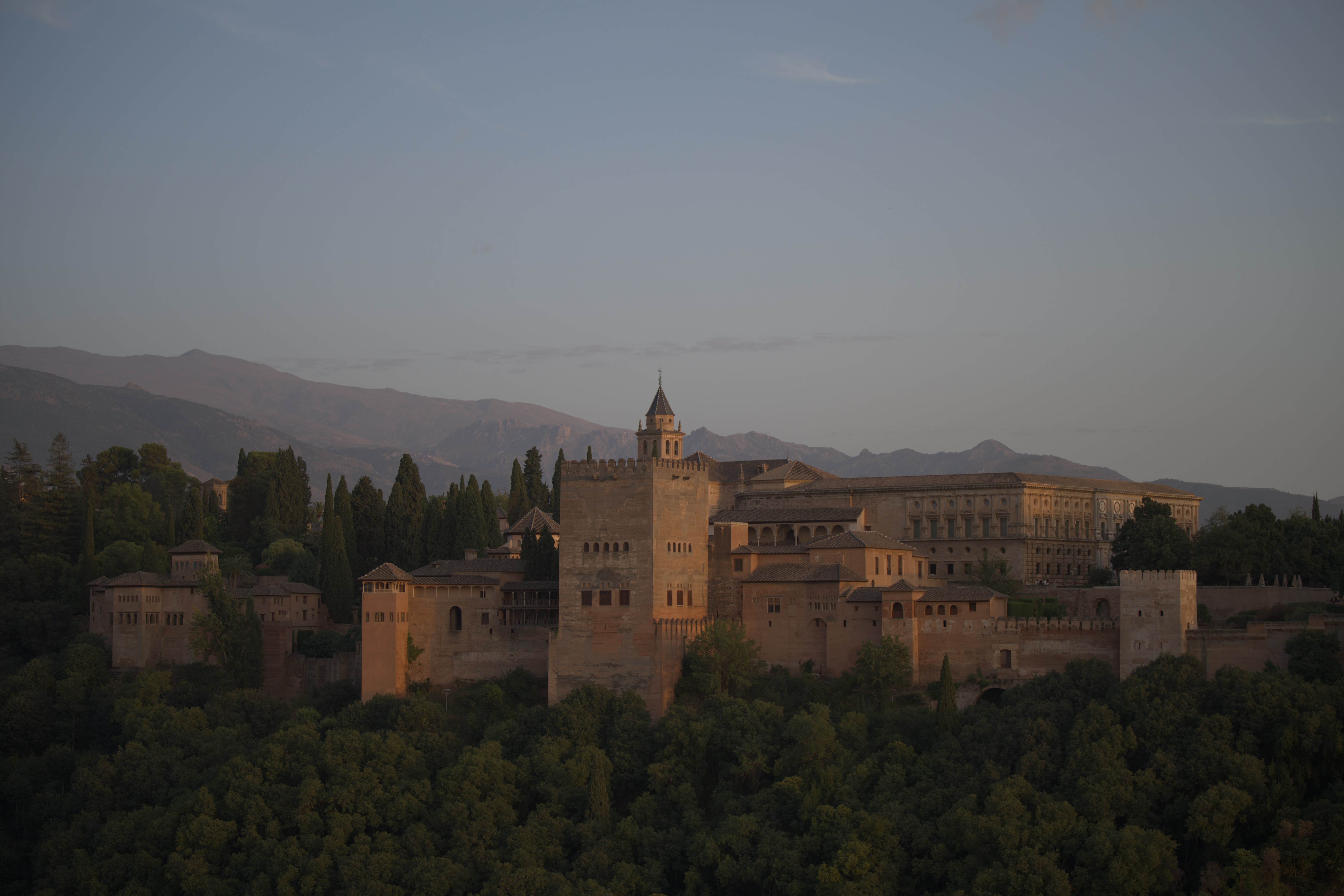} \\
  \tiny 40.2\,dB / 0.157 & \tiny 40.7\,dB / 0.085 & \tiny DSC\_6718 \\[3pt]
  \includegraphics[width=0.32\textwidth]{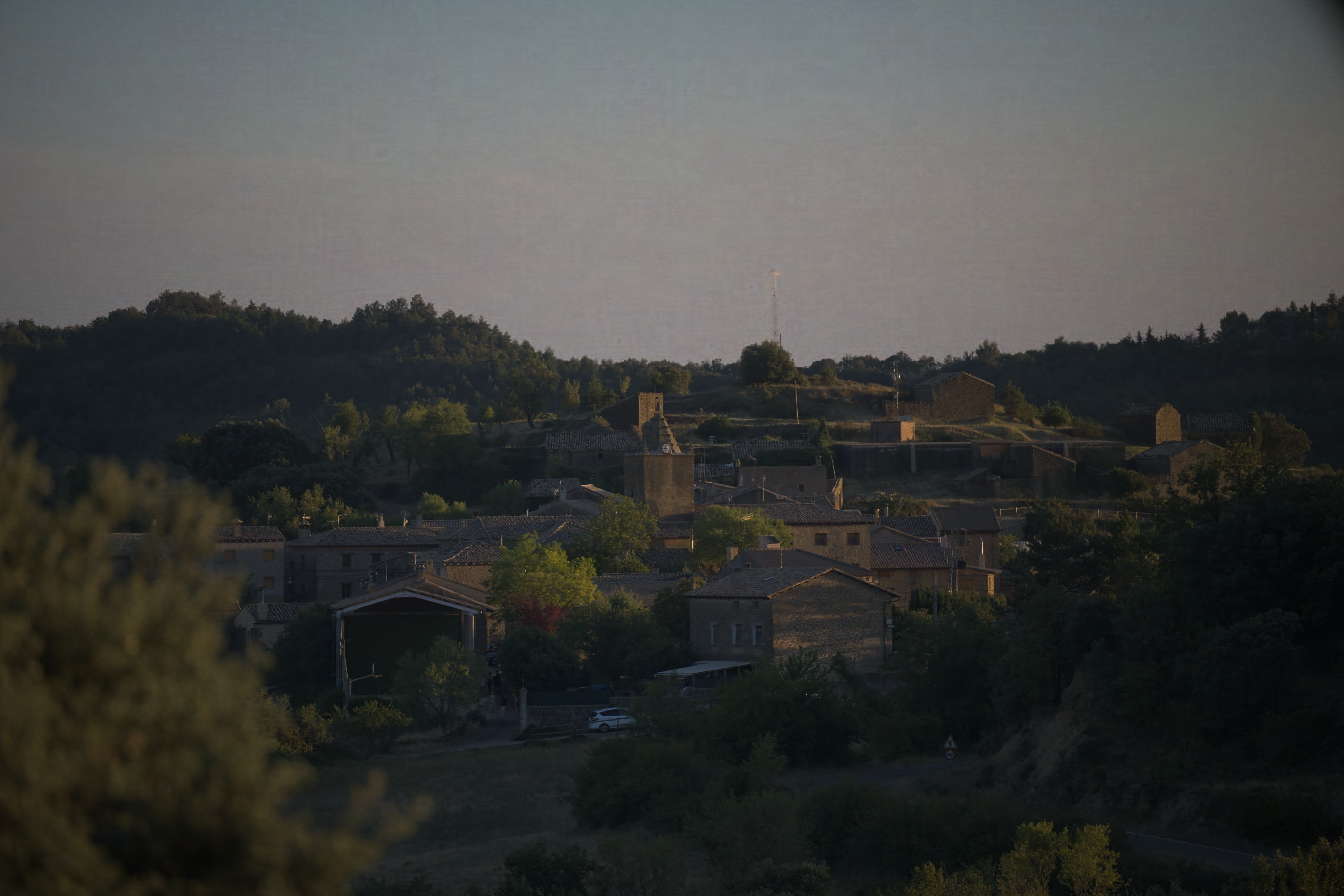} &
  \includegraphics[width=0.32\textwidth]{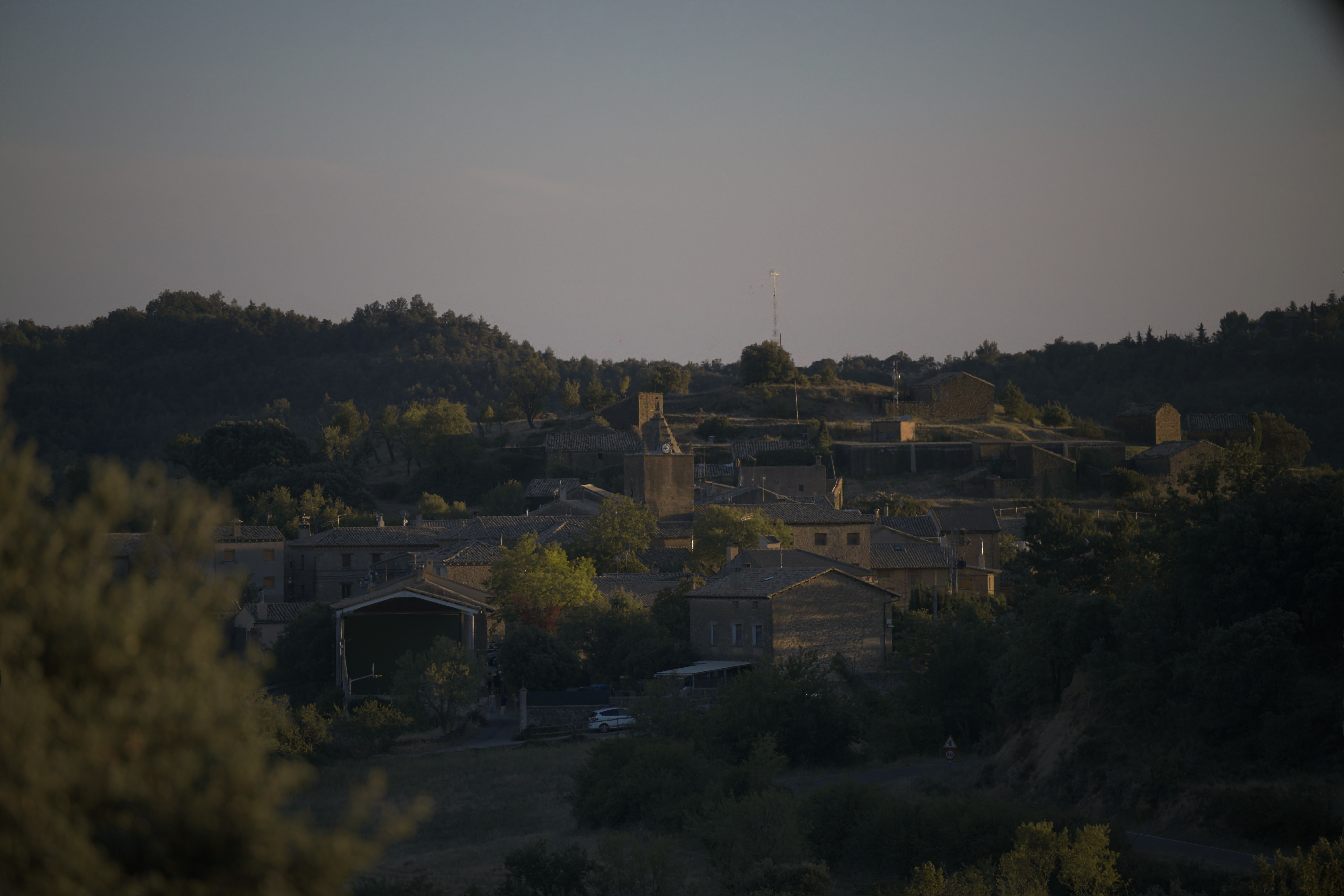} &
  \includegraphics[width=0.32\textwidth]{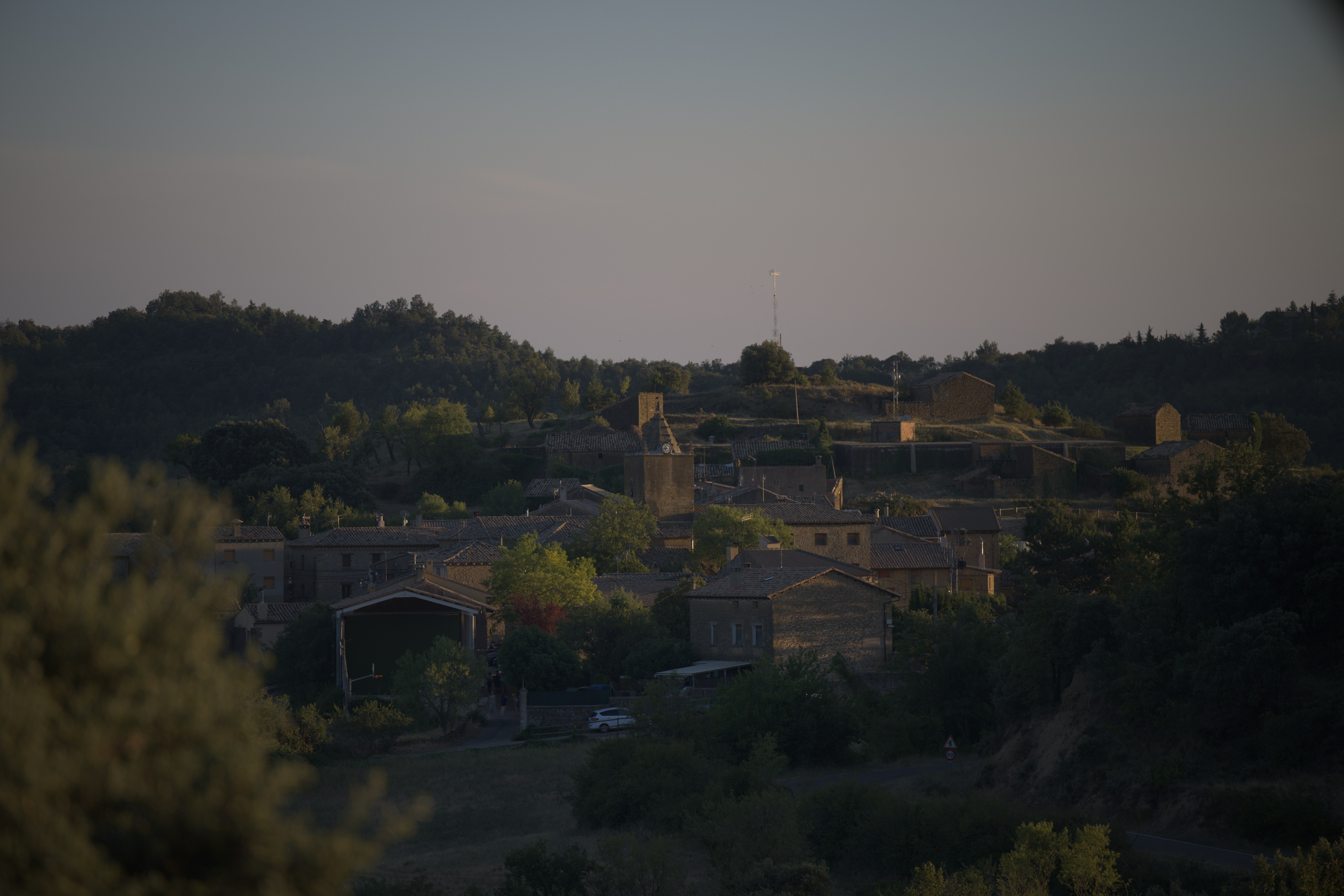} \\
  \tiny 39.7\,dB / 0.134 & \tiny 39.9\,dB / 0.055 & \tiny DSC\_1764 \\[3pt]
  \includegraphics[width=0.32\textwidth]{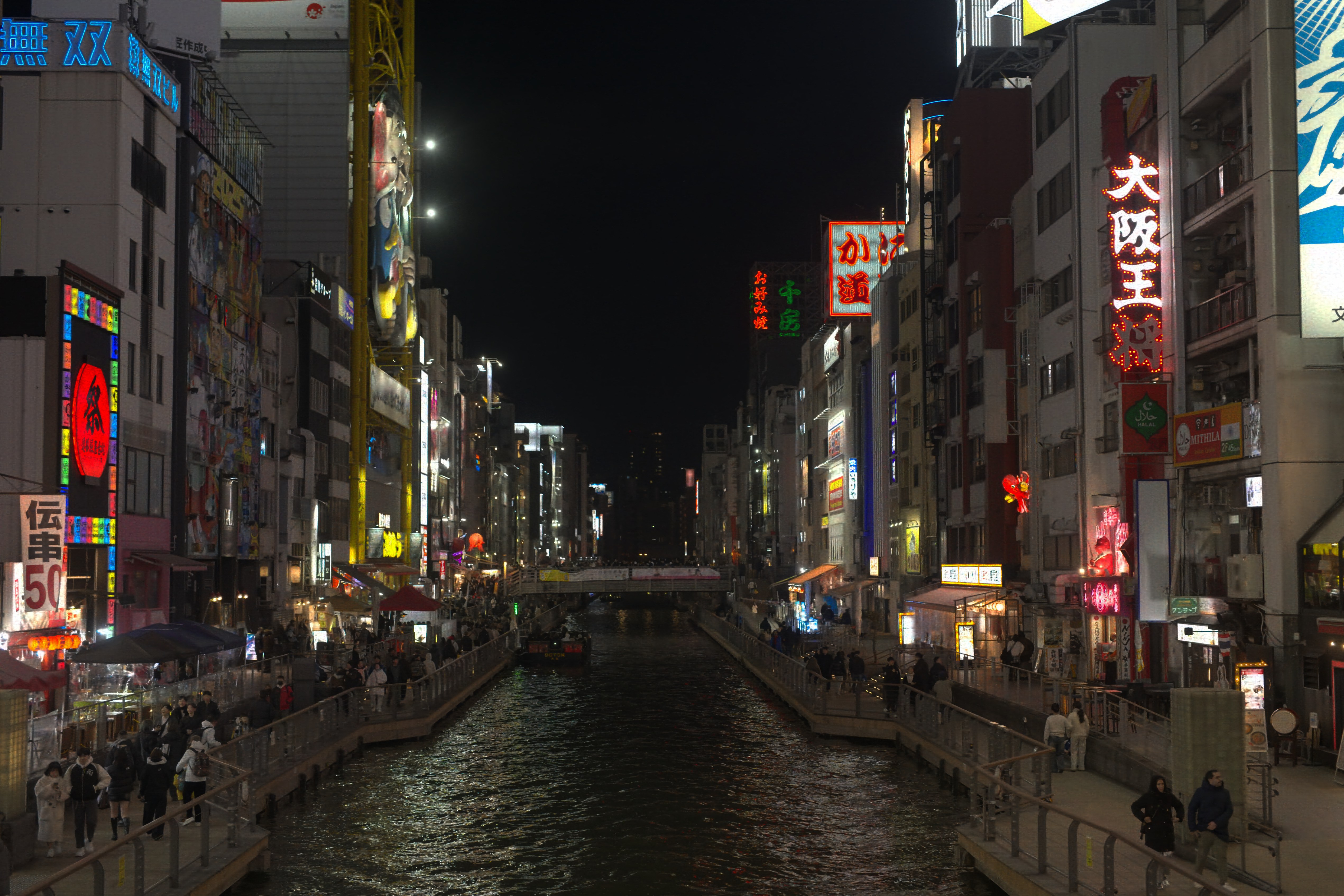} &
  \includegraphics[width=0.32\textwidth]{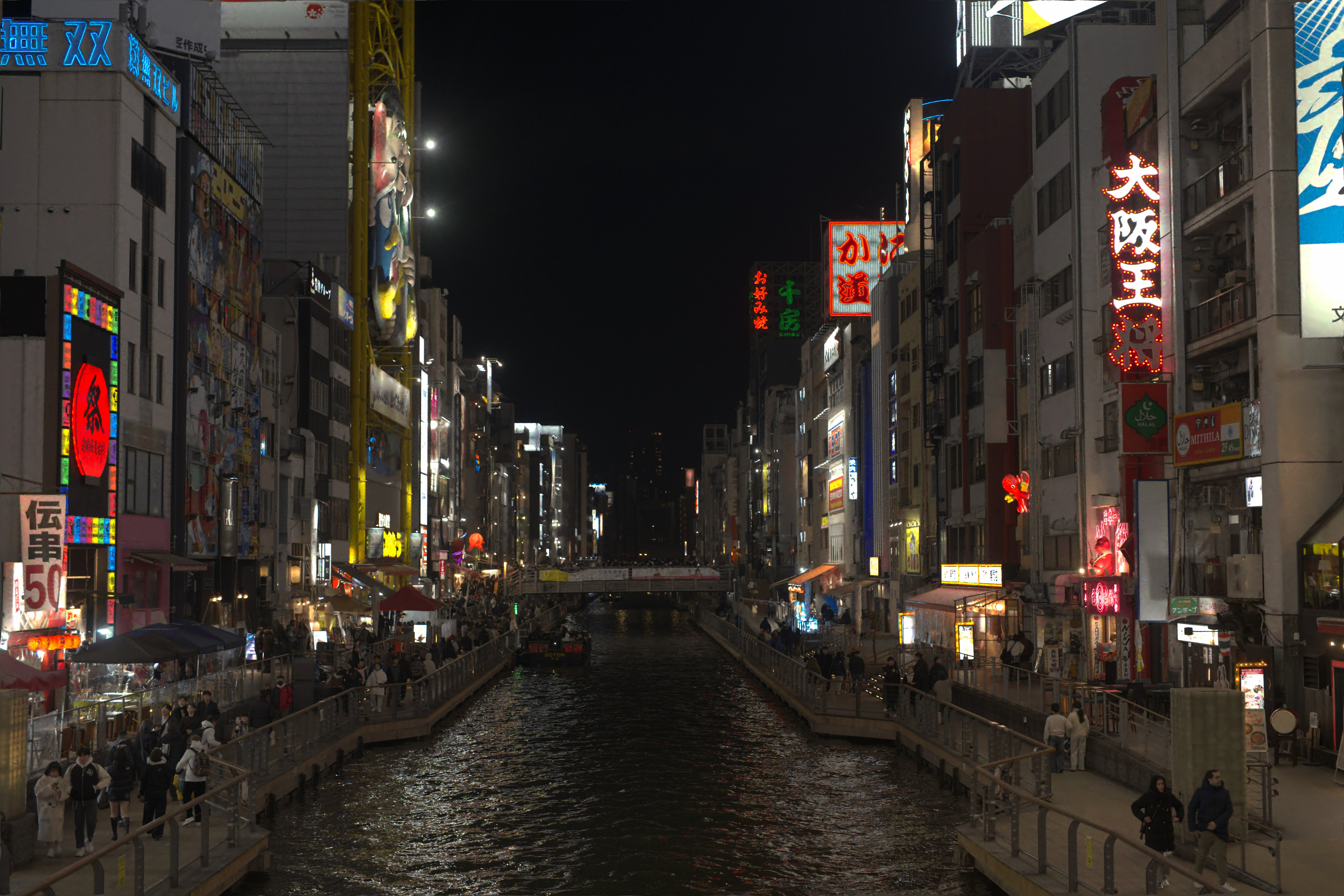} &
  \includegraphics[width=0.32\textwidth]{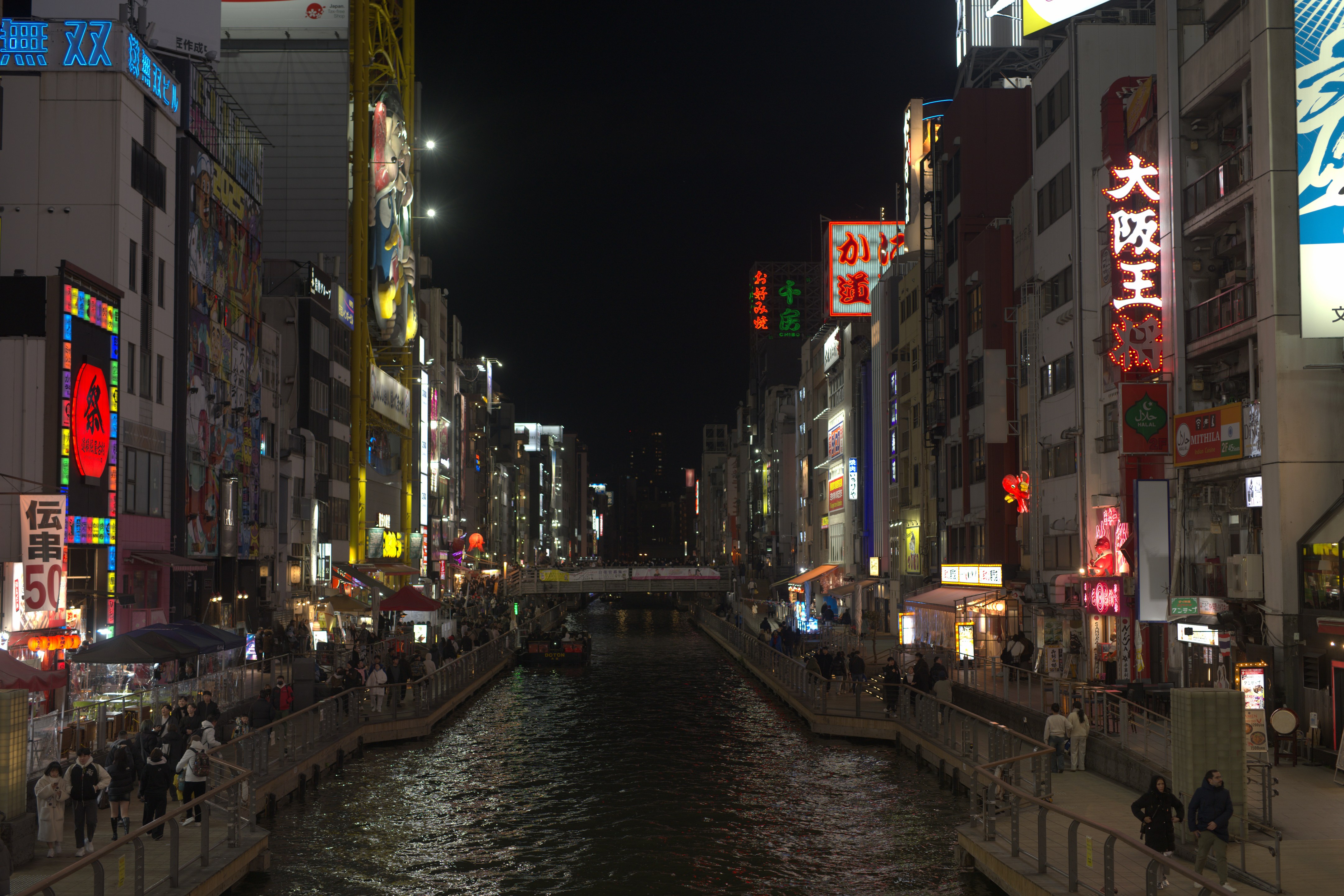} \\
  \tiny 30.5\,dB / 0.063 & \tiny 29.9\,dB / 0.059 & \tiny DSC\_6926 \\[3pt]
  \includegraphics[width=0.32\textwidth]{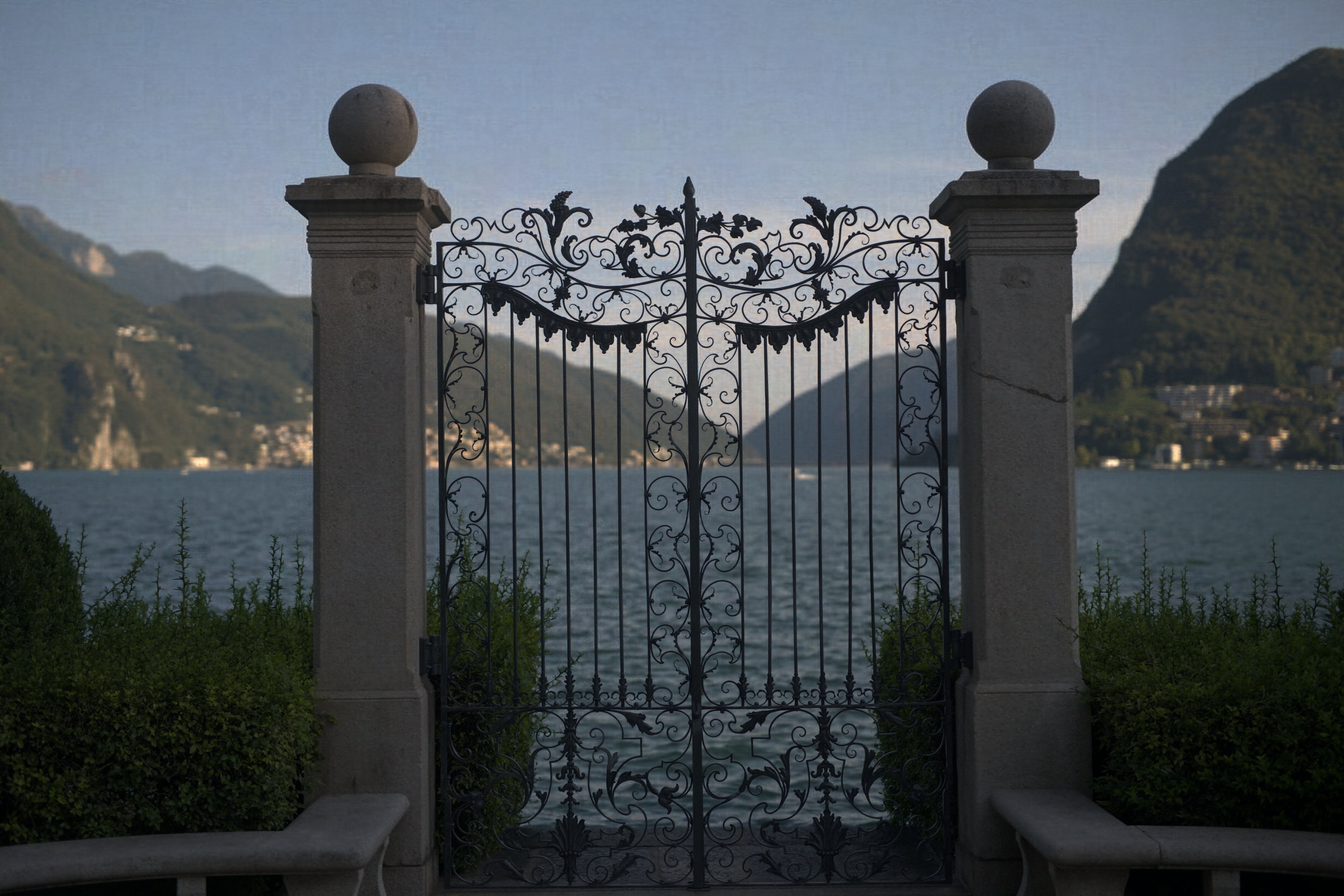} &
  \includegraphics[width=0.32\textwidth]{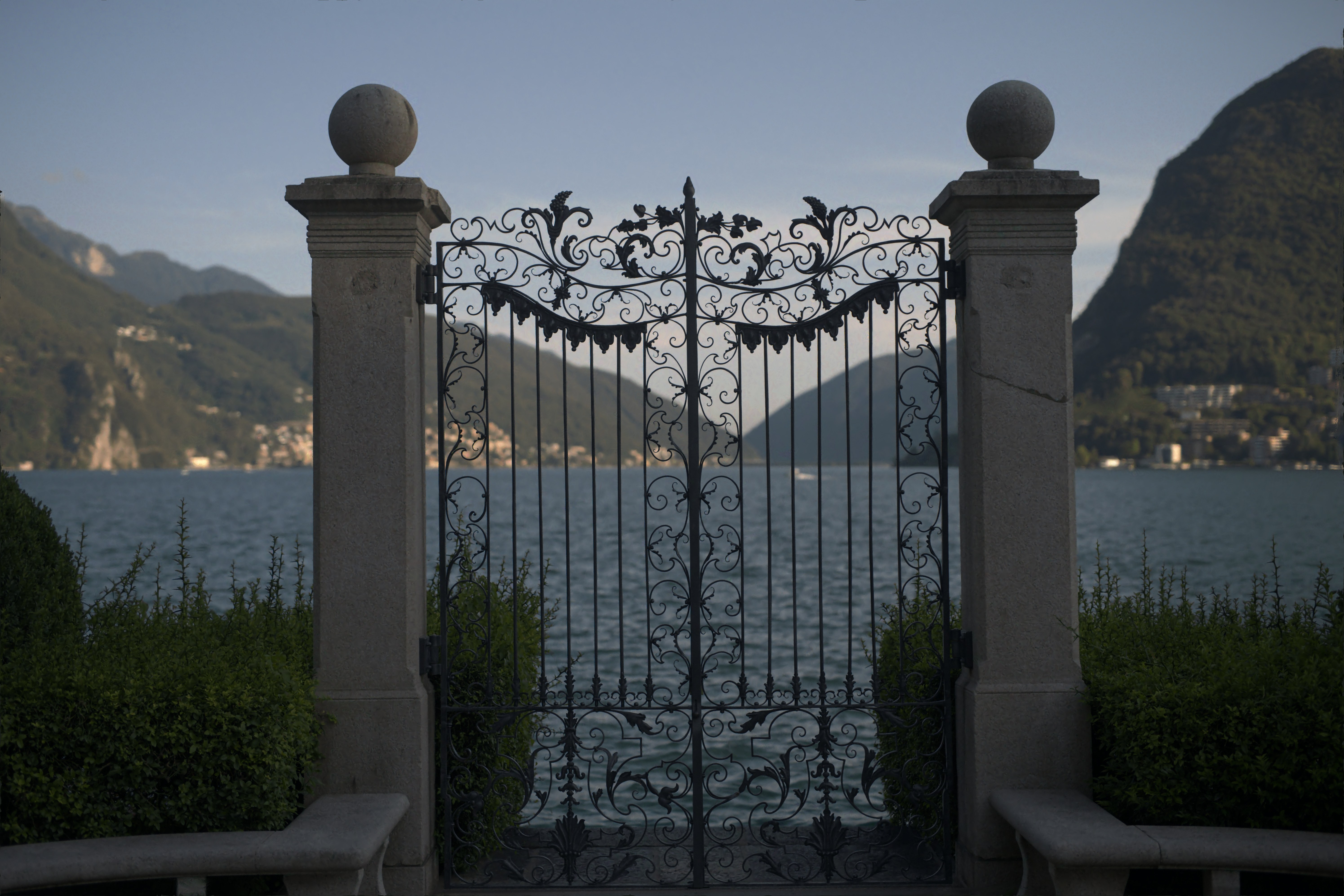} &
  \includegraphics[width=0.32\textwidth]{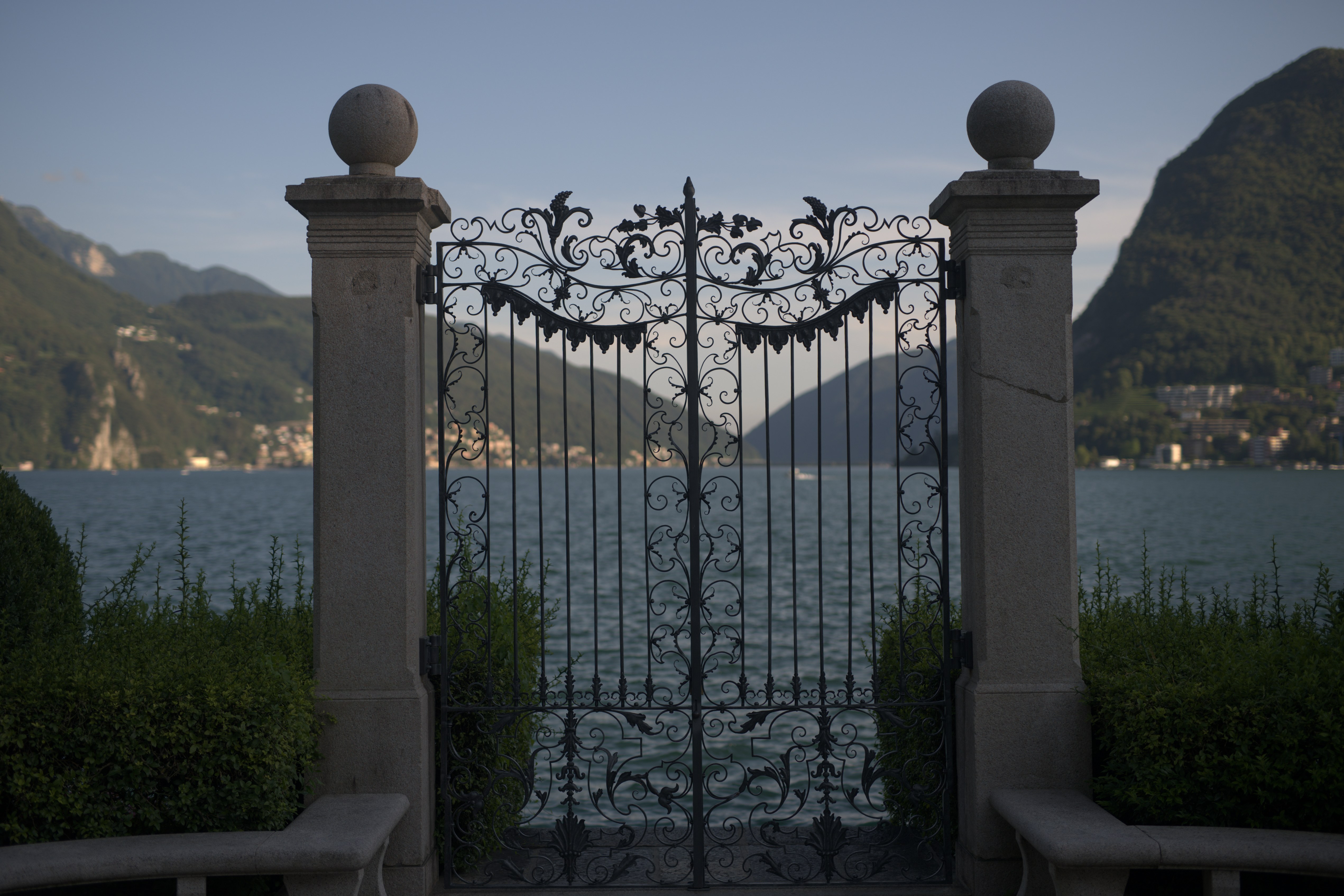} \\
  \tiny 36.9\,dB / 0.068 & \tiny 36.6\,dB / 0.035 & \tiny DSC\_7685 \\[3pt]
  \end{tabular}
  \caption{Visual comparison at 100$\times$ compression. Each cell shows tonemapped PSNR\,(dB) and LPIPS. Our method (NHT) consistently achieves lower LPIPS (better perceptual quality) than Instant NGP at comparable or higher PSNR.
  The images are taken from the dataset we curated.}
  \label{fig:2d_extended_comparison}
  \label{fig:visual_100x}
  \end{figure*}

\cref{tab:2d_perimage,tab:2d_perimage_100x} provide per-image results for the 2D image fitting experiment at $10\times$ and $100\times$ compression ratios, respectively. We provide visual comparisons in \cref{fig:image_compression} and \cref{fig:2d_extended_comparison}.
Our method achieves competitive PSNR in both $\mu$-law and tonemapped spaces while substantially outperforming Instant NGP in perceptual quality (LPIPS).
The advantage is most pronounced at high compression ratios, where the non-overlapping mesh topology and $C^1$ feature interpolation avoid the ringing and block artifacts that plague hash-grid methods.

\begin{table}[t]
\centering
\caption{Per-image comparison at 10$\times$ compression on our 14-bit HDR RAW dataset (45.7\,MP). We compare NHT (Ours), Instant NGP~\cite{mueller2022instant}, and JPEG-XL~\cite{jpegxl}. All neural methods use the same parameter budget. Metrics on tonemapped sRGB images unless noted.}
\label{tab:2d_perimage}
\resizebox{\textwidth}{!}{%
\begin{tabular}{l ccc ccc ccc ccc}
\toprule
& \multicolumn{3}{c}{PSNR$\uparrow$ ($\mu$-law)} & \multicolumn{3}{c}{PSNR$\uparrow$ (tonemapped)} & \multicolumn{3}{c}{SSIM$\uparrow$ (tonemapped)} & \multicolumn{3}{c}{LPIPS$\downarrow$ (tonemapped)} \\
\cmidrule(lr){2-4} \cmidrule(lr){5-7} \cmidrule(lr){8-10} \cmidrule(lr){11-13}
Image & NHT & NGP & JXL & NHT & NGP & JXL & NHT & NGP & JXL & NHT & NGP & JXL \\
\midrule
\texttt{DSC\_1012} & 33.98 & \underline{34.04} & \textbf{37.70} & \underline{36.92} & \textbf{37.59} & 30.38 & \underline{0.97} & 0.96 & \textbf{0.99} & \underline{0.0090} & 0.0137 & \textbf{0.0004} \\
\texttt{DSC\_1764} & \underline{40.06} & 39.04 & \textbf{49.03} & \textbf{43.61} & \underline{41.73} & 32.02 & \underline{0.98} & 0.97 & \textbf{0.99} & \underline{0.0401} & 0.1149 & \textbf{0.0012} \\
\texttt{DSC\_1967} & 33.60 & \underline{35.01} & \textbf{41.48} & \textbf{36.50} & \underline{36.19} & 27.92 & \underline{0.95} & 0.93 & \textbf{0.99} & \underline{0.0108} & 0.0178 & \textbf{0.0004} \\
\texttt{DSC\_2172} & 35.83 & \underline{36.59} & \textbf{41.74} & \underline{40.23} & 38.95 & \textbf{40.31} & \underline{0.97} & 0.95 & \textbf{0.99} & \underline{0.0182} & 0.0317 & \textbf{0.0008} \\
\texttt{DSC\_2655} & \underline{41.80} & 40.91 & \textbf{50.26} & \underline{44.38} & 42.73 & \textbf{55.95} & \underline{0.98} & \underline{0.98} & \textbf{1.00} & \underline{0.0292} & 0.0850 & \textbf{0.0008} \\
\texttt{DSC\_3095} & 38.07 & \underline{38.78} & \textbf{46.05} & \underline{39.83} & 38.43 & \textbf{52.47} & \underline{0.96} & 0.94 & \textbf{1.00} & \underline{0.0184} & 0.0450 & \textbf{0.0009} \\
\texttt{DSC\_3351} & 40.00 & \underline{42.20} & \textbf{48.40} & \underline{44.53} & 42.97 & \textbf{55.37} & \underline{0.98} & \underline{0.98} & \textbf{1.00} & \underline{0.0046} & 0.0179 & \textbf{0.0002} \\
\texttt{DSC\_4748} & 33.39 & \underline{34.68} & \textbf{37.97} & \textbf{37.53} & \underline{37.08} & 31.97 & \underline{0.95} & 0.94 & \textbf{0.99} & \underline{0.0218} & 0.0544 & \textbf{0.0010} \\
\texttt{DSC\_5007} & 36.09 & \underline{36.80} & \textbf{47.75} & \underline{37.46} & 36.54 & \textbf{49.40} & \underline{0.93} & 0.90 & \textbf{0.99} & \underline{0.0250} & 0.0430 & \textbf{0.0009} \\
\texttt{DSC\_6718} & \underline{41.30} & 40.15 & \textbf{51.27} & \underline{44.57} & 41.64 & \textbf{55.28} & \underline{0.98} & 0.97 & \textbf{1.00} & \underline{0.0464} & 0.1345 & \textbf{0.0010} \\
\texttt{DSC\_6926} & 31.24 & \underline{33.15} & \textbf{34.93} & \underline{33.27} & \textbf{34.28} & 27.42 & \underline{0.93} & 0.91 & \textbf{0.98} & \underline{0.0246} & 0.0449 & \textbf{0.0022} \\
\texttt{DSC\_9008} & \underline{40.78} & 40.10 & \textbf{49.73} & \underline{41.93} & 39.98 & \textbf{53.74} & \underline{0.98} & 0.96 & \textbf{1.00} & \underline{0.0111} & 0.0493 & \textbf{0.0006} \\
\texttt{DSC\_1658} & \underline{39.05} & 38.51 & \textbf{50.39} & \underline{41.68} & 39.57 & \textbf{53.53} & \underline{0.97} & 0.95 & \textbf{1.00} & \underline{0.0471} & 0.1445 & \textbf{0.0018} \\
\texttt{DSC\_4176} & 35.17 & \underline{36.89} & \textbf{48.17} & 36.50 & \underline{36.64} & \textbf{50.77} & \underline{0.95} & 0.93 & \textbf{1.00} & \underline{0.0215} & 0.0597 & \textbf{0.0006} \\
\texttt{DSC\_7685} & 37.11 & \underline{37.54} & \textbf{44.83} & 39.08 & \underline{39.33} & \textbf{53.30} & \underline{0.97} & 0.96 & \textbf{1.00} & \underline{0.0184} & 0.0550 & \textbf{0.0004} \\
\midrule
\textit{Average} & 37.16 & \underline{37.63} & \textbf{45.31} & \underline{39.87} & 38.91 & \textbf{44.66} & \underline{0.96} & 0.95 & \textbf{0.99} & \underline{0.0231} & 0.0608 & \textbf{0.0009} \\
\bottomrule
\end{tabular}}
\end{table}

\begin{table}[t]
\centering
\caption{Per-image comparison at 100$\times$ compression (same dataset and setup as \cref{tab:2d_perimage}).}
\label{tab:2d_perimage_100x}
\resizebox{\textwidth}{!}{%
\begin{tabular}{l ccc ccc ccc ccc}
\toprule
& \multicolumn{3}{c}{PSNR$\uparrow$ ($\mu$-law)} & \multicolumn{3}{c}{PSNR$\uparrow$ (tonemapped)} & \multicolumn{3}{c}{SSIM$\uparrow$ (tonemapped)} & \multicolumn{3}{c}{LPIPS$\downarrow$ (tonemapped)} \\
\cmidrule(lr){2-4} \cmidrule(lr){5-7} \cmidrule(lr){8-10} \cmidrule(lr){11-13}
Image & NHT & NGP & JXL & NHT & NGP & JXL & NHT & NGP & JXL & NHT & NGP & JXL \\
\midrule
\texttt{DSC\_1012} & 31.49 & \underline{31.65} & \textbf{32.58} & \underline{33.34} & \textbf{33.51} & 30.18 & \underline{0.93} & \underline{0.93} & \textbf{0.98} & \underline{0.0243} & 0.0255 & \textbf{0.0065} \\
\texttt{DSC\_1764} & \underline{37.41} & 37.22 & \textbf{41.89} & \textbf{39.95} & \underline{39.70} & 31.89 & \underline{0.96} & \underline{0.96} & \textbf{0.98} & \underline{0.0552} & 0.1341 & \textbf{0.0344} \\
\texttt{DSC\_1967} & 30.65 & \underline{31.05} & \textbf{34.25} & \underline{32.41} & \textbf{32.61} & 27.66 & \underline{0.89} & 0.88 & \textbf{0.96} & \underline{0.0328} & 0.0341 & \textbf{0.0057} \\
\texttt{DSC\_2172} & 32.96 & \underline{33.31} & \textbf{36.09} & 35.59 & \underline{35.90} & \textbf{38.59} & \underline{0.92} & \underline{0.92} & \textbf{0.97} & \underline{0.0419} & 0.0436 & \textbf{0.0182} \\
\texttt{DSC\_2655} & \underline{40.30} & 40.16 & \textbf{43.56} & \textbf{41.74} & \underline{41.36} & 26.05 & \underline{0.98} & \underline{0.98} & \textbf{0.99} & \underline{0.0396} & 0.0942 & \textbf{0.0121} \\
\texttt{DSC\_3095} & \underline{36.52} & 35.93 & \textbf{39.03} & \underline{36.74} & 35.82 & \textbf{43.61} & \underline{0.93} & 0.92 & \textbf{0.98} & \underline{0.0459} & 0.0663 & \textbf{0.0184} \\
\texttt{DSC\_3351} & 40.79 & \underline{41.23} & \textbf{43.53} & \underline{42.12} & \textbf{42.51} & 29.01 & \underline{0.98} & \underline{0.98} & \textbf{0.99} & \underline{0.0099} & 0.0193 & \textbf{0.0020} \\
\texttt{DSC\_4748} & 31.05 & \underline{31.26} & \textbf{32.83} & \underline{34.23} & \textbf{34.38} & 31.47 & \underline{0.91} & \underline{0.91} & \textbf{0.96} & \underline{0.0516} & 0.0724 & \textbf{0.0142} \\
\texttt{DSC\_5007} & 33.88 & \underline{34.00} & \textbf{38.84} & 33.65 & \underline{33.68} & \textbf{36.35} & \underline{0.85} & 0.84 & \textbf{0.95} & \underline{0.0692} & 0.0729 & \textbf{0.0143} \\
\texttt{DSC\_6718} & 38.55 & \underline{38.68} & \textbf{43.99} & \underline{40.74} & 40.19 & \textbf{47.59} & \underline{0.96} & \underline{0.96} & \textbf{0.99} & \underline{0.0847} & 0.1571 & \textbf{0.0362} \\
\texttt{DSC\_6926} & 29.48 & \underline{29.72} & \textbf{30.01} & \underline{29.91} & \textbf{30.54} & 27.02 & \underline{0.87} & 0.86 & \textbf{0.93} & \underline{0.0587} & 0.0629 & \textbf{0.0331} \\
\texttt{DSC\_9008} & 38.38 & \underline{38.39} & \textbf{43.15} & \underline{37.90} & \textbf{37.92} & 34.54 & \underline{0.95} & \underline{0.95} & \textbf{0.98} & \underline{0.0275} & 0.0514 & \textbf{0.0096} \\
\texttt{DSC\_1658} & \underline{36.56} & 36.45 & \textbf{41.84} & \underline{37.87} & 37.29 & \textbf{44.60} & \underline{0.93} & \underline{0.93} & \textbf{0.98} & \underline{0.0890} & 0.1768 & \textbf{0.0489} \\
\texttt{DSC\_4176} & 33.49 & \underline{33.56} & \textbf{38.28} & 33.60 & \underline{33.63} & \textbf{40.03} & \underline{0.90} & 0.89 & \textbf{0.97} & \underline{0.0521} & 0.0880 & \textbf{0.0123} \\
\texttt{DSC\_7685} & 34.42 & \underline{34.91} & \textbf{38.10} & 36.63 & \underline{36.86} & \textbf{44.49} & \underline{0.94} & \underline{0.94} & \textbf{0.98} & \underline{0.0345} & 0.0681 & \textbf{0.0081} \\
\midrule
\textit{Average} & 35.06 & \underline{35.17} & \textbf{38.53} & \textbf{36.43} & \underline{36.39} & 35.54 & \underline{0.93} & 0.92 & \textbf{0.97} & \underline{0.0478} & 0.0778 & \textbf{0.0183} \\
\bottomrule
\end{tabular}}
\end{table}

\cref{tab:2d_hyperparameters} provides the full set of hyperparameters for the 2D image fitting experiment.

\begin{table}[t]
  \centering
  \caption{Hyperparameters and optimization details for the 2D image fitting experiment. The parameter budget (maximum vertices, MLP size) is derived from the target compression ratio and image dimensions and bitrate.}
  \label{tab:2d_hyperparameters}
  \setlength{\tabcolsep}{3pt}
  \small
  \begin{tabular}{ll}
    \toprule    
    \textbf{Representation} & \\
    \midrule
    Mesh topology & Delaunay triangulation (shared vertices) \\
    Interpolation & Clough--Tocher $C^1$ cubic\\
    Feature activation & SinCos: $[\sin(f), \cos(f)]$\\
    Feature dim.\ per vertex & 4-16 \\
    MLP hidden dim.\ $64-128$ $\times$ $2-4$ layers & \\
    \midrule
    \textbf{Training} & \\
    \midrule
    Total iterations & 25\,000 \\
    Pixel sampling & Stratified, 160\,000 pixels/batch \\
    Loss function & MSE \\
    \midrule
    \textbf{Learning rates (initial)} & \\
    \midrule
    Vertex positions & $1 \times 10^{-4}$ \\
    Vertex features & $5 \times 10^{-3}$ \\
    MLP weights & $5 \times 10^{-5}$ \\
    \midrule
    \textbf{LR schedule} & \\
    \midrule
    All parameters & Exponential decay: \\
                    & $\eta(t) = \eta_0 \cdot 0.33^{\max(0,\,(t - 20\text{k})\,/\,10\text{k})}$ \\
    \midrule
    \textbf{Densification} & \\
    \midrule
    Strategy & Coarse-to-fine Delaunay re-meshing \\
    Schedule & Steps 1\,500--15\,000, every 500 steps \\
    Growth rate & 0.35$\times$ current vertex count per step \\
    Scoring & DSSIM-based, area-weighted: \\
            & $\text{score}_t = \overline{\text{DSSIM}}_t \cdot |\text{pixels}_t|^{0.75}$ \\
    Min.\ triangle size & 3 pixels (skip smaller triangles) \\
    Initialization & Edge-aware sampling (Sobel) \\
    \midrule
    \textbf{HDR pipeline} & \\
    \midrule
    Input & 14-bit RAW (Nikon NEF), 45.7\,MP \\
    Encoding space & $\mu$-law: $f(x) = \log(1 + \mu\, x/w) / \log(1+\mu)$ \\
    $\mu$ parameter & 5\,000 \\
    White level $w$ & $2^{14} - 1 = 16\,383$ \\
    GT pixel lookup & Bilinear interpolation \\
    \bottomrule
  \end{tabular}
\end{table}


\end{document}